\documentclass{article}

\usepackage[preprint]{neurips_2026}

\usepackage[utf8]{inputenc}
\usepackage[T1]{fontenc}

\usepackage{graphicx}
\usepackage{comment}
\usepackage{subcaption}
\usepackage{wrapfig}
\usepackage{float}
\usepackage{tabularx}
\usepackage{xcolor}

\usepackage{placeins}

\usepackage{booktabs}
\usepackage{makecell}
\usepackage{amsmath}
\usepackage{amsfonts}
\usepackage{nicefrac}
\usepackage{microtype}
\usepackage{url}

\usepackage{fvextra}
\usepackage{fancyvrb}
\usepackage{inconsolata}
\usepackage[most]{tcolorbox}

\definecolor{graybox}{gray}{0.95}

\newtcolorbox{user_prompt}{
  enhanced,
  breakable,
  colback=graybox,
  colframe=gray!50,
  boxrule=0.3pt,
  arc=1mm,
  left=2mm,
  right=2mm,
  top=2mm,
  bottom=2mm,
  fontupper=\ttfamily\scriptsize
}

\usepackage[
  colorlinks=true,
  linkcolor={red!50!black},
  citecolor={blue!50!black},
  urlcolor={blue!80!black},
  hyperfootnotes=false
]{hyperref}

\usepackage[capitalize]{cleveref}

\newcommand{\appitem}[2]{%
  #1 \dotfill \hyperref[#2]{\pageref*{#2}}\\
}

\newcommand{\appsubitem}[2]{%
  \hspace*{1.75em} #1 \dotfill \hyperref[#2]{\pageref*{#2}}\\
}

\makeatletter
\newcommand{\blankthanks}[1]{%
  \g@addto@macro\@thanks{%
    \footnotetext[0]{#1}%
  }%
}
\makeatother

\title{Neuron Populations Exhibit\\
Divergent Selectivity with Scale
}

\author{%
\centerline{%
\textbf{Amil Dravid}\textsuperscript{1}
\quad
\textbf{Yasaman Bahri}\textsuperscript{1}
\quad
\textbf{Alexei A. Efros}\textsuperscript{1}
\quad
\textbf{Yossi Gandelsman}\textsuperscript{2}
}%
\\[0.75em]
\centerline{%
\hspace*{-2.7em}%
{\small\textsuperscript{1}UC Berkeley
\qquad
\textsuperscript{2}TTIC}%
}%
\\[-0.1em]
\centerline{%
\hspace*{-1.5em}%
}%
}

\begin{document}
\maketitle

\vspace{-2em}
\begin{abstract}

We investigate whether neuron populations within neural networks evolve predictably with scale, extending scaling laws beyond macroscopic observables such as loss. To probe this question, we study \textit{Rosetta Neurons}, a previously characterized class of neurons whose activation patterns are similar across independently trained models~\citep{dravid2023rosetta}. In separate analyses of language models up to 30B parameters and vision models up to 5B parameters, we observe that the population of Rosetta Neurons follows a sublinear power law in model size, growing in absolute number but occupying a shrinking fraction of the total neuron count. We further observe a \textit{Neuron Polarization Effect}: Rosetta Neurons become more selective and increasingly monosemantic with scale, separating from a growing non-Rosetta population that remains less selective. An analytical model balancing feature utility against limited neuron capacity explains the sublinear power-law scaling and this polarization effect. Finally, we find that Rosetta Neurons become more domain-specialized with scale and illustrate their selectivity through a targeted data-filtering case study for continued pretraining. Our results point to a scaling law for interpretable, shared neuron-level structure, linking model size to systematic changes in neuron universality, selectivity, and specialization.\footnote{Project Page:
\mbox{\href{https://avdravid.github.io/rosetta-neuron-scaling}{\nolinkurl{https://avdravid.github.io/rosetta-neuron-scaling}}}\\[-0.1em] \hspace*{1.55em}
YB, AE, and YG jointly advised this work.}
\end{abstract}

\section{Introduction}
\begin{figure}[t]
    \centering
    \vspace{-0.1cm}
    \includegraphics[width=1.0\linewidth]{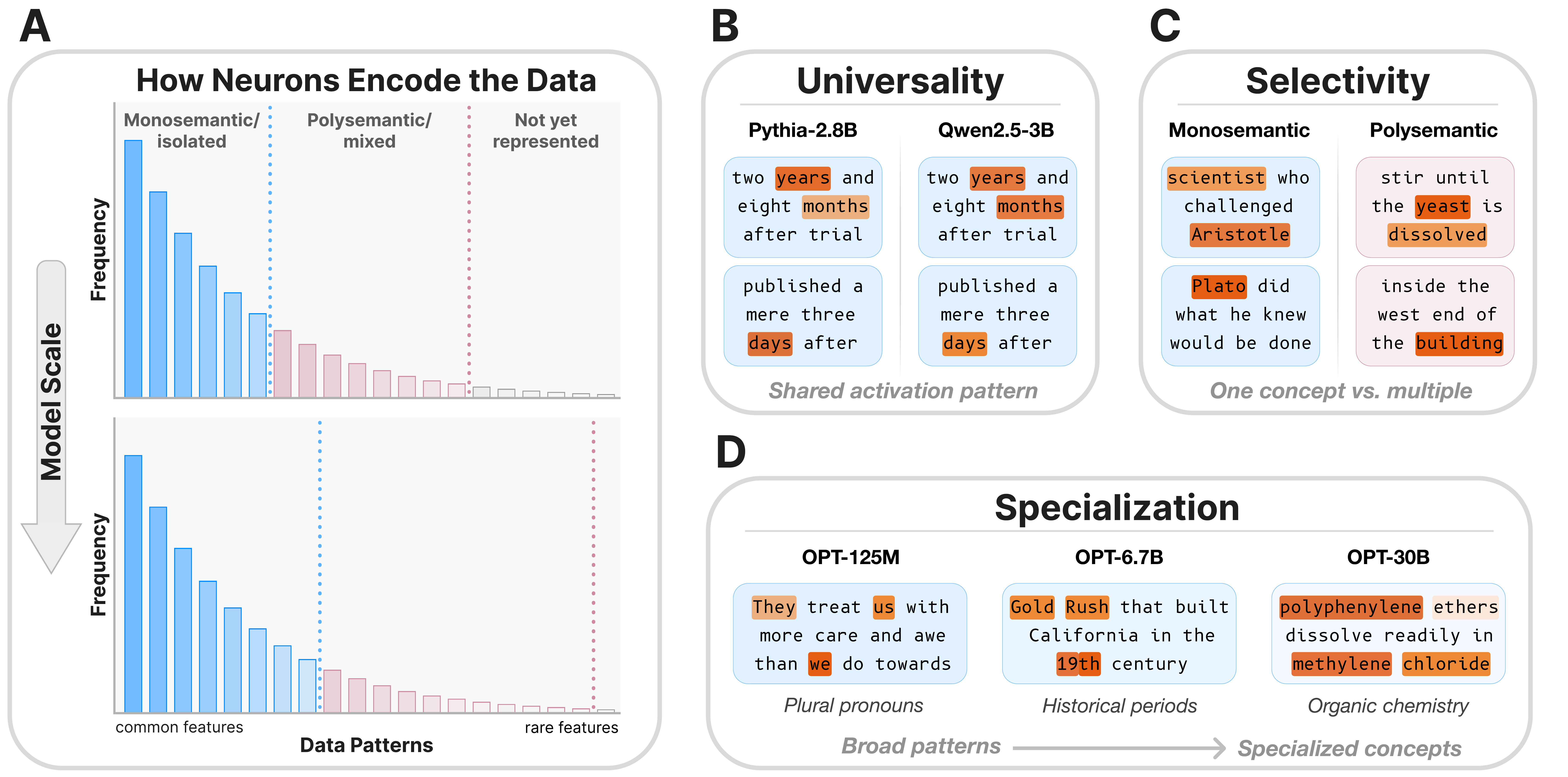}
 \caption{\textbf{Neuron populations across scale.} To study how neuron populations scale, we use Rosetta Neurons: units that recur across different models. (A) Features compete for representation in a finite set of neurons, leaving them isolated, mixed, or unrepresented at a given scale. This picture guides our analysis of universality, selectivity, and specialization. In panels B–D, each column shows top-activating contexts from a single neuron. (B) Universality: how does the recurring Rosetta Neuron population scale? (C) Selectivity: do recurring neurons become increasingly monosemantic relative to polysemantic neurons? (D) Specialization: which features of the data distribution are encoded by Rosetta Neurons at different scales? Across language and vision models, Rosetta Neurons grow in number but shrink as a fraction of all neurons, while becoming more selective and specialized.}
   \label{fig:teaser_figure}
    \vspace{-0.4cm}
\end{figure}

A central question in both deep learning and neuroscience is how neurons encode structure in the world. In biological systems, this question has motivated longstanding debates about whether representations are localized in single units~\citep{hubel1962receptive} or distributed across populations~\citep{haxby2001distributed}, as well as studies of how such structure might recur across subjects~\citep{hasson2004intersubject}. An analogous question arises for artificial neural networks. Despite a growing body of work probing neurons, such as those encoding sentiment in language models~\citep{radford2017learning} or object segments in vision models~\citep{bau2017network}, most neurons in today's large-scale models are not easily interpretable~\citep{bills2023language}.

One reason for this difficulty is that neuron-level representations may vary in how cleanly they isolate individual features. Some neurons appear relatively monosemantic, responding selectively to coherent semantic concepts, while \textit{superposition} predicts that others may be polysemantic, encoding multiple unrelated features when models represent more features than available dimensions~\citep{elhage2022toy}. This makes monosemanticity and polysemanticity natural questions at the neural-population level: as models scale, which neurons become selective, which remain in superposition, and how do these properties relate to shared internal function across independently trained models?

These questions become especially salient in the scaling regime, where models change not only in performance but also in representational capacity. Neural scaling laws~\citep{kaplan2020scaling, hoffmann2022training, hestness2017deep} have revealed striking regularities in external model behavior, with quantities such as loss following precise power-law relationships with compute, data, and model size. Yet these laws say little about how representations are organized inside the model. We therefore apply a similar scaling analysis at the neuron level, asking whether cross-model recurring neuron populations evolve predictably with scale: how their size changes, whether they become more selective and monosemantic, and how their features specialize with respect to the data distribution.

To make these questions measurable, we adopt the notion of \textit{Rosetta Neurons} from \cite{dravid2023rosetta}: neurons whose activation patterns recur across independently trained models. We use this recurring population to study three neuron-level properties across scale: universality, selectivity, and specialization (\Cref{fig:teaser_figure}). Prior work suggests that cross-model recurrence can reflect stable features of the data distribution rather than idiosyncrasies of any one model, but leaves open whether such recurring neurons follow systematic laws across scale and model families in both language and vision.

In this paper, we treat Rosetta Neurons as a scaling observable, identifying recurring cross-model units in language and vision models spanning 80M to 30B parameters (\Cref{sec:rosetta-methods}). In~\Cref{scaling_law_section}, we apply a scaling-law analysis to this population, showing that its size grows predictably with model size according to a sublinear power law in neuron count: larger models contain more shared units, yet these units occupy a shrinking fraction of the total neuron population.

We then develop a phenomenological model linking this sublinear scaling to limited neuron capacity (\Cref{subsec:analytical_model}). As networks scale, more features become monosemantically represented in individual neurons and thus more likely to recur across models. However, this set grows more slowly than the broader pool of new features represented in network-specific patterns of superposition (\Cref{fig:teaser_figure}A). This theory predicts a \textit{Neuron Polarization Effect}: Rosetta Neurons become increasingly selective with scale, separating from the polysemantic non-Rosetta population. We validate this prediction across language and vision models (\Cref{sec:properties}) and further show that Rosetta Neurons become increasingly specialized, shifting toward domains such as code and mathematics.

Finally, we demonstrate this specialization functionally in~\Cref{sec:data_filtering}: Rosetta Neuron activations can filter data from a specific code domain with near-oracle accuracy, yielding continued-pretraining performance that matches training on ground-truth domain data. Together, our analysis provides a scalable way to identify shared, interpretable, and predictable structure within large models, revealing a neuron-level population whose size, selectivity, and specialization evolve systematically with scale.


\section{Related Work}

\textbf{Neural scaling laws.}
A large body of work has shown that neural network performance follows predictable scaling behavior with respect to model scale, dataset size, and training compute, often well described by power laws of the form \(\mathcal{L}(x)=\mathcal{L}_{\infty}+Ax^{-\alpha}\) \citep{hestness2017deep,kaplan2020scaling}. Subsequent work refined these observations by identifying compute-optimal training regimes \citep{hoffmann2022training} and demonstrating similar scaling behavior across modalities and architectures \citep{zhai2022scaling}. A number of works have sought to explain these laws from various perspectives, including (but not limited to) the alignment of tasks to models, the structure of the data distribution, the intrinsic dimensionality of the data manifold, the geometry of learned features, and the learning dynamics of rare tasks under capacity constraints~\citep{bordelon2020spectrum, michaud2023quantization, bahri2024explaining, liu2025superposition, cagnetta2026deriving, huang2026larger}. We extend this perspective to internal representations, using Rosetta Neurons to study whether a form of shared neuron-level structure also scales predictably.

\textbf{Structured representations and superposition.}
Understanding how neural networks develop structured representations has been a central question in deep learning. Interpretability work suggests that learned representations can exhibit approximately linear organization \citep{mikolov2013distributed,arora2018linear,park2024linear}. Mechanistic studies show that limited-capacity models can represent features in superposition, with toy-model analyses framing this as a capacity-allocation tradeoff between ignoring, isolating, and superposing features~\citep{elhage2022toy,scherlis2022polysemanticity}. Motivated by this structured but superposed organization, sparse decomposition approaches seek to recover interpretable features that are more selective and monosemantic~\citep{bricken2023monosemanticity}. We build on these ideas at the neuron level, showing that scaling polarizes neurons into a more selective, interpretable Rosetta population against a polysemantic background.

\textbf{Representational similarity and universality.}
A complementary line of work studies the extent to which learned representations are shared across models and training runs. Various similarity measures have been proposed to quantify alignment between neural representations \citep{raghu2017svcca,morcos2018insights,kornblith2019similarity}, while alternative approaches such as model stitching probe functional equivalence \citep{bansal2021revisiting,lenc2015understanding}. These ideas build on earlier notions of representational similarity from neuroscience \citep{kriegeskorte2008representational,edelman1998representation}, where shared geometric structure is used to compare representations across systems \citep{kriegeskorte2008matching,haxby2001distributed}. Together, these perspectives have supported growing evidence for convergent structure in learned representations \citep{sorscher2022neural,huh2024position,liconvergence}. At the neuron level, \citet{dravid2023rosetta} identified shared ``Rosetta Neurons'' across diverse vision models, while \citet{gurneeuniversal} studied shared neurons in small-scale GPT-2 models trained from different random seeds. Beyond these settings, we study Rosetta Neurons within both language and vision models spanning heterogeneous architectures and datasets, and characterize their scaling laws.

\section{Identifying Rosetta Neurons}
\label{sec:rosetta-methods}

We now describe how we identify Rosetta Neurons -- common neurons with similar responses across models. We define MLP neurons and the token-wise activations used to compare them, then describe how to measure pairwise similarity and filter these similarities into reliable shared neuron pairs.
\subsection{MLP Neurons in Transformer Models}
We study both language and vision models built on the Transformer architecture~\citep{vaswani2017attention}. Given an input sequence of text tokens or image patches, the model maps each token to an embedding, producing a sequence \(e_1,\dots,e_T \in \mathbb{R}^d\). These are then processed through a series of blocks that alternate between multi-head self-attention and multilayer perceptron (MLP) layers. We focus exclusively on the MLP layers. Let \(h_t^{(\ell)}\) denote the hidden state at token position $t$ in layer $\ell$. The MLP first applies an affine transformation followed by an element-wise nonlinearity $\phi$:
\begin{equation}
m_t^{(\ell)} = \phi\!\left(W_{\mathrm{in}}^{(\ell)} h_t^{(\ell)} + b_{\mathrm{in}}^{(\ell)}\right) \in \mathbb{R}^{d_{\mathrm{mlp}}},
\end{equation}
It then projects this activation back to the residual stream as \(\tilde h_t^{(\ell)}=W_{\mathrm{out}}^{(\ell)}m_t^{(\ell)}+b_{\mathrm{out}}^{(\ell)}\). Each coordinate \(m_t^{(\ell,c)}\) of the intermediate activation vector is a \emph{neuron}, indexed by layer \(\ell\) and channel \(c\).

\subsection{Quantifying Pairwise Neuron Similarity}

We now describe how to compare MLP neurons between two models. Let model \(A\) and model \(B\) be two Transformers of the same modality. We run the same dataset of inputs \(\mathcal X=\{x_i\}_{i=1}^n\) through both models, where each \(x_i\) is either an image or a sequence of text. The Transformer processes each input jointly to produce token-wise activations, which are then aligned across models. In vision models, activations are aligned to a common spatial grid, while in language models they are aligned to a shared sequence of text positions. Details of this alignment procedure are given in~\Cref{sec:aligning_tokens}.

Consider a neuron \(u=(\ell_A,c_A)\) in model \(A\) and a neuron \(v=(\ell_B,c_B)\) in model \(B\). Let \(m_t^u(x_i)\) and \(m_t^v(x_i)\) denote their activations at aligned token \(t\) for input \(x_i\). We compare neurons \(u\) and \(v\) by computing the Pearson correlation between these activations on all aligned tokens across the dataset: 
\begin{equation}
\mathrm{sim}(u,v)
=
\frac{1}{N}\sum_{i,t}
\frac{m_t^u(x_i)-\mu_u}{\sigma_u}
\cdot
\frac{m_t^v(x_i)-\mu_v}{\sigma_v},
\label{eq:pearson_correlation}
\end{equation}

where \(\mu_u\) and \(\sigma_u^2\) are the empirical mean and variance for neuron $u$ over the dataset of \(N\) total aligned tokens, and \(\mu_v\) and \(\sigma_v^2\) are defined analogously for neuron $v$. We compute this similarity for all neuron pairs between models $A$ and $B$, producing a large table of pairwise similarities.

\begin{figure}[t!]
    \centering
    \includegraphics[width=1.0\linewidth]{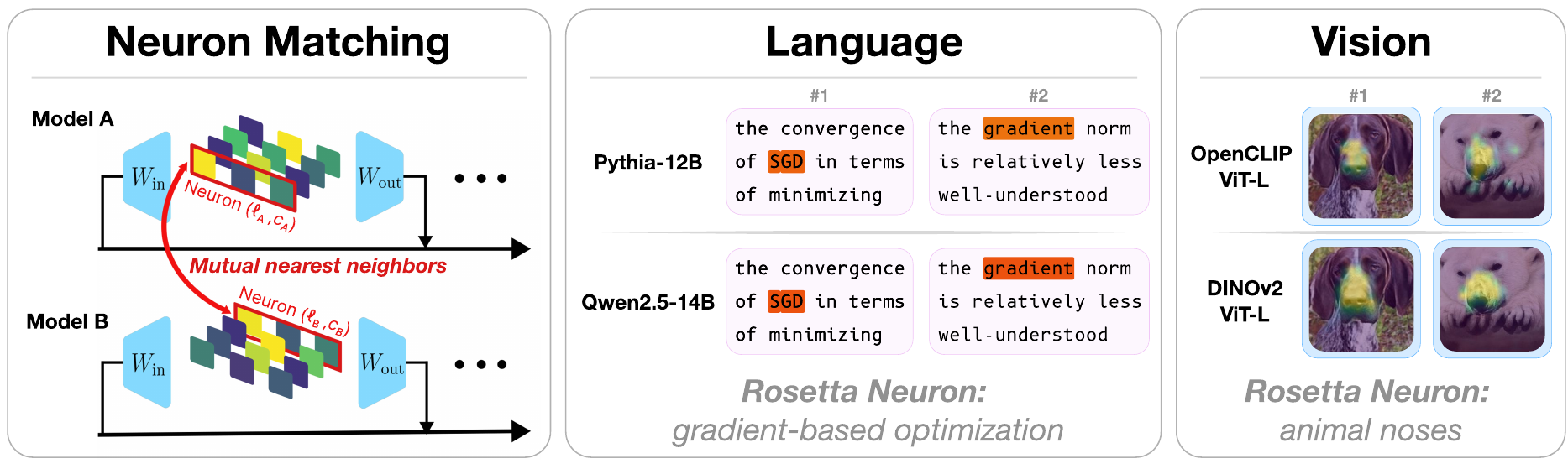}
    \caption{\textbf{Identifying Rosetta Neurons.} We compare MLP neuron activations across independently trained models on the same inputs and identify mutual nearest-neighbor pairs under Pearson correlation. The language and vision examples show individual matched neuron pairs firing on the same high-activating inputs, revealing similar activation patterns and coherent shared concepts.}

   \label{fig:method_figure}
  \vspace{-10pt}
\end{figure}

\subsection{Filtering Pairwise Neuron Correspondences}
To detect reliable matches between neurons, we retain only those pairs that are nearest neighbors of one another under our similarity metric. Specifically, for neuron \(u\in\mathcal{N}(A)\) and \(v\in\mathcal{N}(B)\), we include the match \((u,v)\) in \(\mathcal{R}(A,B)\) iff \(v\in\mathrm{NN}_k(u;B)\) and \(u\in\mathrm{NN}_k(v;A)\), where \(\mathrm{NN}_k(u;B)\) denotes the top-\(k\) neurons in model \(B\) most similar to \(u\), and analogously for \(\mathrm{NN}_k(v;A)\). Unless otherwise stated, all experiments use the default setting $k=1$; we ablate this choice in~\Cref{subsec:topk_ablation}. Mutual nearest-neighbor matching provides a simple way to retain robust correspondences while filtering asymmetric or noisy nearest-neighbor matches~\citep{dekel2015best}.  We therefore interpret the pairs of neurons in \(\mathcal{R}(A,B)\) as common units across models, and refer to them as \emph{Rosetta Neurons}. \Cref{fig:method_figure} visualizes how these neurons are identified and what the resulting matches look like.


\section{Scaling Laws for Rosetta Neurons}
\label{scaling_law_section}
In this section, we study how the number of Rosetta Neurons scales with model size in both language and vision models. We first describe the experimental setting, including the data and model families. We then analyze the resulting scaling trends in both domains and compare them against a null baseline. We conclude with a phenomenological model to explain the observed scaling behavior.

\subsection{Experimental Setup}

\textbf{Data.} In the language setting, each pair of models used for matching is evaluated on a shared set of approximately 10 million tokens formed by sampling sequences i.i.d. from The Pile \citep{gao2020pile}. For vision models, we follow \citet{dravid2023rosetta} and match a generative model with a discriminative model. We sample 50,000 class-balanced images from a diffusion model by conditioning on ImageNet-1k labels \citep{deng2009imagenet}, and then pass these images through the discriminative model. Ablations on the data are provided in \Cref{sec:dataset_ablations}. 

\textbf{Model Families.} In the language domain, we consider the Pythia, GPT-2, OPT, and Qwen-2.5 model families, spanning roughly 100 million to 30 billion parameters \citep{gpt2,pythia,opt,qwen2.5}. For vision, we use discriminative models from the OpenCLIP, DINOv2, and Pixio families, spanning scales from approximately 80 million to 5 billion parameters~\citep{openclip, radford2021learning, dinov2, pixio}. For the generative model, we leverage one-step diffusion models built on the Diffusion Transformer architecture~\citep{peebles2023scalable}. We provide further details on the model families in~\Cref{sec:rosetta_neuron_scaling_appendix}.

\textbf{Forming the scaling curves.} Given two model families, we form a point on the scaling curve by selecting one model from each family at approximately matched scale. We apply the matching procedure from Section~\ref{sec:rosetta-methods} and record the number of discovered Rosetta Neurons. Repeating this across increasingly larger models yields a scaling curve for that family pair. When two or more curves share a common model, we collapse them into a single trajectory by intersecting the Rosetta Neurons identified within the shared model across the comparisons. This produces a stricter notion of universality by keeping neurons shared across more than two model families, allowing us to test whether the same scaling trend holds under a more selective definition of Rosetta Neurons.

\textbf{Power-law functional form.} We model the number of discovered Rosetta Neurons as a power law in model size, $|\mathcal{R}| = c x^{\alpha}$, where \(|\mathcal{R}|\) denotes the number of discovered neuron correspondences. Because the matched models are selected to be at approximately the same scale, we use their average total neuron count \(x\) to represent the scale of the matched models. In this parameterization, the exponent \(\alpha\) is the primary quantity of interest, since it governs how rapidly the number of shared neurons grows with scale, while the constant \(c\) depends more strongly on the particular model families and experimental setting. We estimate the fitted curve using ordinary least squares in log-log space.

\begin{figure}[t]
    \centering

    \begin{subfigure}[t]{0.49\textwidth}
        \centering
        \includegraphics[
            width=\linewidth,
            trim={0bp 0bp 785.585bp 0bp},
            clip
        ]{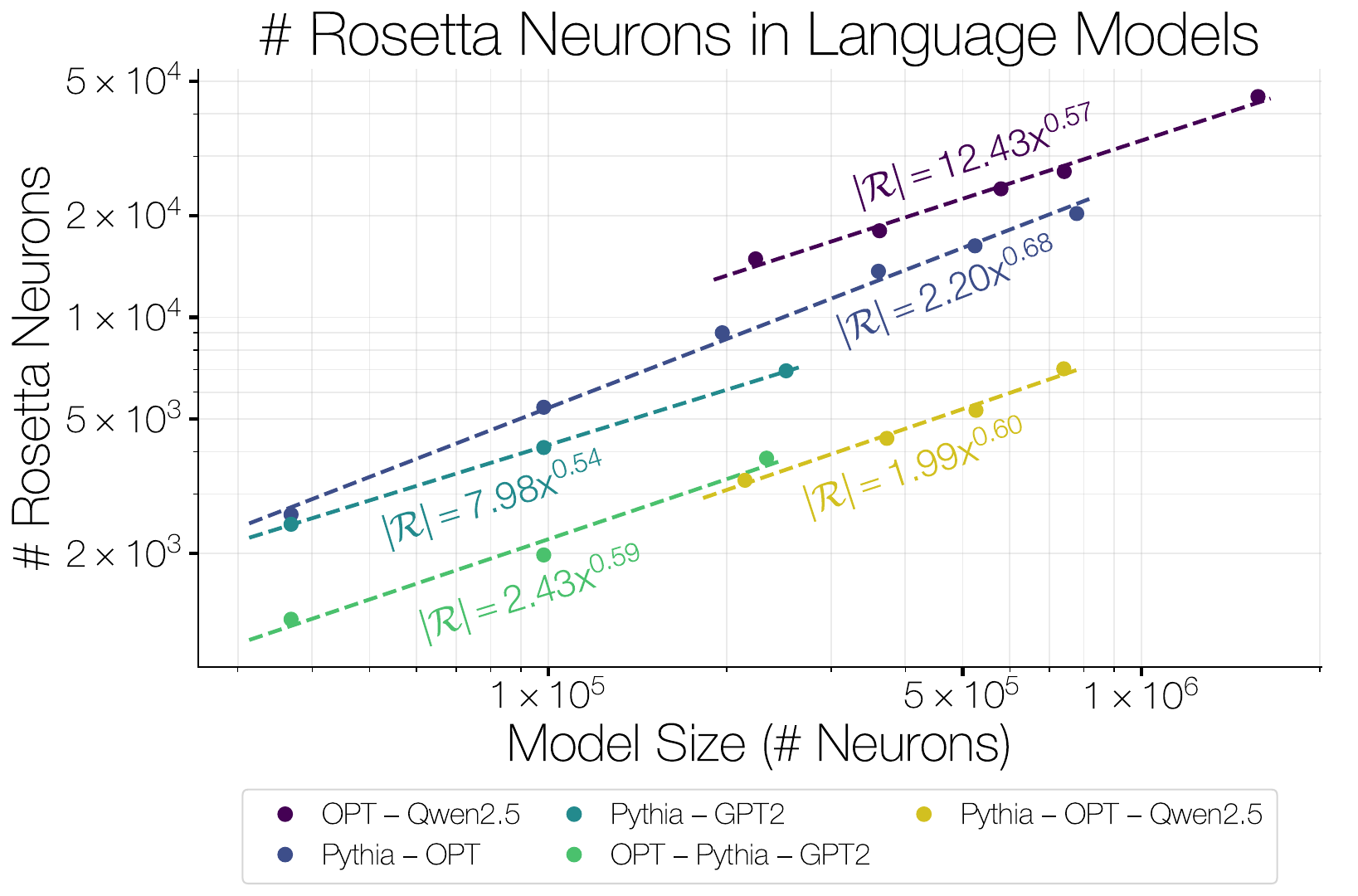}
        \caption{Rosetta Neuron scaling laws in language models.}
        \label{fig:language_scaling}
    \end{subfigure}
    \hfill
    \begin{subfigure}[t]{0.49\textwidth}
        \centering
        \includegraphics[
            width=\linewidth,
            trim={785.585bp 0bp 0bp 0bp},
            clip
        ]{figures/language_vision_scaling_side_by_side.pdf}
        \caption{Rosetta Neuron scaling laws in vision models.}
        \label{fig:vision_scaling}
    \end{subfigure}

    \caption{\textbf{Scaling laws for Rosetta Neurons in language and vision models.} We plot the number of discovered Rosetta Neurons for various model families at different scales. Dashed lines show power-law fits in log-log space. Across all family comparisons, the fitted exponents are sublinear, and the corresponding fits achieve $R^2$ values around 0.99. Further details are provided in~\Cref{sec:rosetta_neuron_scaling_appendix}.}
    \label{fig:main_scaling_figure}
    \vspace{-1em}
\end{figure}

\subsection{Rosetta Neurons Follow Power-Law Scaling}
\label{subsec:rosetta_powerlaw}
We establish that the Rosetta Neuron population follows a sublinear power law in both language and vision models. We then show that this trend is absent in untrained networks, indicating that the scaling law is a property of learned representations rather than an artifact of the matching procedure.

\textbf{Power-law scaling in language models.} 
Across the language model families we study, the number of Rosetta Neurons increases predictably with model size as shown in~\Cref{fig:language_scaling}. The resulting curves are well fit by power laws, with \(R^2\) values around 0.99. Notably, the fitted exponents lie in a narrow and consistently sublinear range, between approximately $0.5-{0.7}$. This indicates that the number of shared neurons grows predictably with model size, but slower than the rate at which models are scaling. These scaling laws span model sizes from roughly 100 million to 30 billion parameters, corresponding to about 40 thousand to 2 million neurons. The same qualitative behavior holds for the collapsed curves (Pythia-OPT-Qwen2.5, OPT-Pythia-GPT2), suggesting that the scaling trends remain even under a stricter notion of universality.

\textbf{Power-law scaling in vision models.}
We observe that a similar scaling trend for Rosetta Neurons emerges in vision models trained with distinct objectives. The scaling curves are well described by power laws, as shown in \Cref{fig:vision_scaling}, with $R^2$ values around 0.99. The fitted exponents fall in a sublinear range between approximately $0.5-{0.7}$, indicating a stable Rosetta Neuron scaling regime across model families. These scaling laws span model sizes from roughly 80 million to 5 billion parameters, or 40 thousand to 600 thousand neurons. Importantly, this trend holds across models trained with contrastive image-text supervision, self-distillation, masked autoencoding, and diffusion or flow-based generative modeling. Collapsing curves across shared models reveals a similar scaling pattern, suggesting that the observed scaling law for Rosetta Neurons is robust across multiple models.

\begin{wrapfigure}{r}{0.45\textwidth}
    \vspace{-15pt}
    \centering
    \includegraphics[width=0.45\textwidth]
    {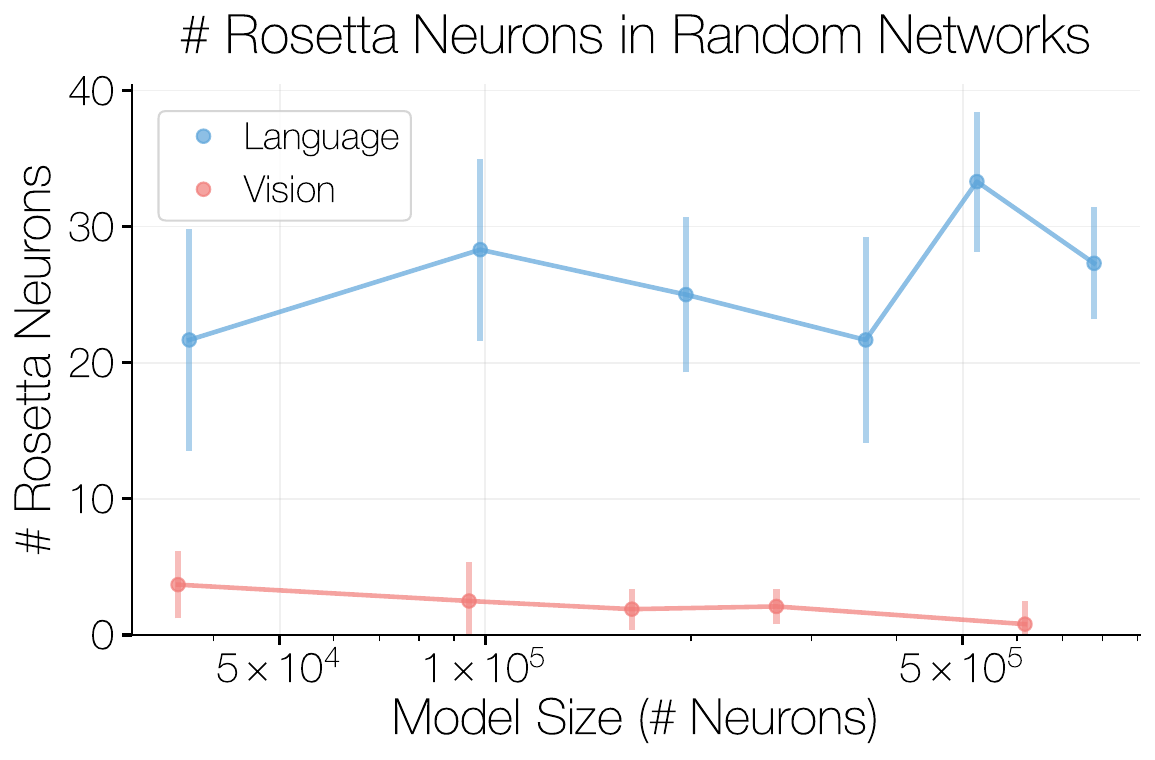}
    \vspace{-16pt}
    \caption{\textbf{Rosetta Neuron counts in untrained networks lack systematic scaling.}}
    \label{fig:rosetta-random}
    \vspace{-0.5cm}
\end{wrapfigure}

\textbf{Power-law scaling is absent in untrained networks.}
To test whether our previously observed scaling laws could be induced by the matching procedure itself, we apply the same pipeline to untrained networks initialized according to their architecture-specific random initialization schemes. We report the results across three random seeds in~\Cref{fig:rosetta-random}. In both language and vision, this yields a marginal number of Rosetta Neuron matches, with no systematic trend as model size increases. Thus, the increasing count of Rosetta Neurons cannot be explained trivially by a larger selection pool. This suggests that our discovery algorithm does not by itself induce the observed scaling behavior. A complementary trained-network null in~\Cref{supp:permutation_null} shows that the scaling trend is also absent when input alignment is corrupted. 

\subsection{An Analytical Model of Rosetta Neuron Scaling}
\label{subsec:analytical_model}

The empirical scaling law in~\Cref{fig:main_scaling_figure} raises a question:
why do Rosetta Neurons grow as a sublinear power law, occupying a shrinking
fraction of the total neuron population? We adopt a capacity-allocation view of
superposition~\citep{elhage2022toy,scherlis2022polysemanticity}: many
features compete for representation in a limited number of neuron coordinates.
High-importance features, which contribute more to reducing downstream loss,
are worth representing more cleanly, while lower-importance features remain
mixed in superposition. This gives a natural model of Rosetta Neurons: neurons
mostly explained by a single shared feature are reproducible across independently
trained networks, whereas neurons that mix many features can do so in
network-specific ways and are harder to match. The analytical model below formalizes this intuition; derivations and simulations are in~\Cref{sec:theory_supp}.

\textbf{Feature-isolation setup.} We consider the setting where a network with \(N\) neuron coordinates remains in a superposition regime as it scales~\citep{liu2025superposition}: it represents \(A(N)>N\) latent features with \(A(N)=\Omega(N)\), so not every feature can receive a dedicated coordinate. This setting could arise through several non-exclusive mechanisms, which we do not attempt to distinguish here. It could come from effectively infinite data relative to finite model size~\citep{elhage2022toy}, or from regularization pressures~\citep{liu2025superposition, bricken2023emergence}. We assume a power-law feature-importance spectrum, motivated by prior scaling-law theories in which learnable structure is organized into ranked modes~\citep{michaud2023quantization, bordelon2020spectrum, bahri2024explaining}. Specifically, features have importance $w_r \propto r^{-\beta}$ with $\beta > 1$, where smaller $r$ denotes a more important feature and $\beta > 1$ ensures finite total importance. Intuitively, importance measures a feature's value for reducing loss, reflecting factors such as frequency, predictiveness, and task relevance.

In a superposition regime, a single neuron may contain signal from many features, and a single feature may be distributed across many neurons. We summarize how cleanly feature \(r\) is captured by any individual neuron with an isolation score \(s_r \ge 0\). Informally, \(s_r\) compares how much of the activation variance in the neuron most aligned with feature \(r\) comes from
feature \(r\), versus from other features mixed into that neuron. Thus, \(s_r\) is large when one neuron is dominated by feature \(r\), and small when every neuron containing feature \(r\) is strongly mixed with other features. We formalize this quantity in~\Cref{sec:theory_supp}, where \(s_r\) is derived
as a signal-to-interference ratio.

We now use this isolation score to relate superposition to Rosetta Neuron matching across models. Assuming matched networks trained on the same data distribution share a common feature spectrum, Rosetta Neurons can be analyzed through whether a feature is sufficiently isolated in a representative single model. Intuitively, high-isolation features produce reproducible matches because the same shared feature dominates the neuron, whereas low-isolation features are obscured by model-specific mixtures of other tail features. We formalize this intuition in~\Cref{supp:empirical-matching} by relating \(s_r\) to expected cross-model activation correlation under model-specific interference. We call a feature Rosetta-detectable if \(s_r \ge \tau\): its shared signal is strong enough relative to model-specific interference to produce a reproducible single-neuron match. We take \(\tau>1\), which ensures in the signal-to-interference model that each detectable feature has a distinct dedicated neuron.


\begin{figure}[t]
    \centering
    \includegraphics[width=0.95\linewidth]{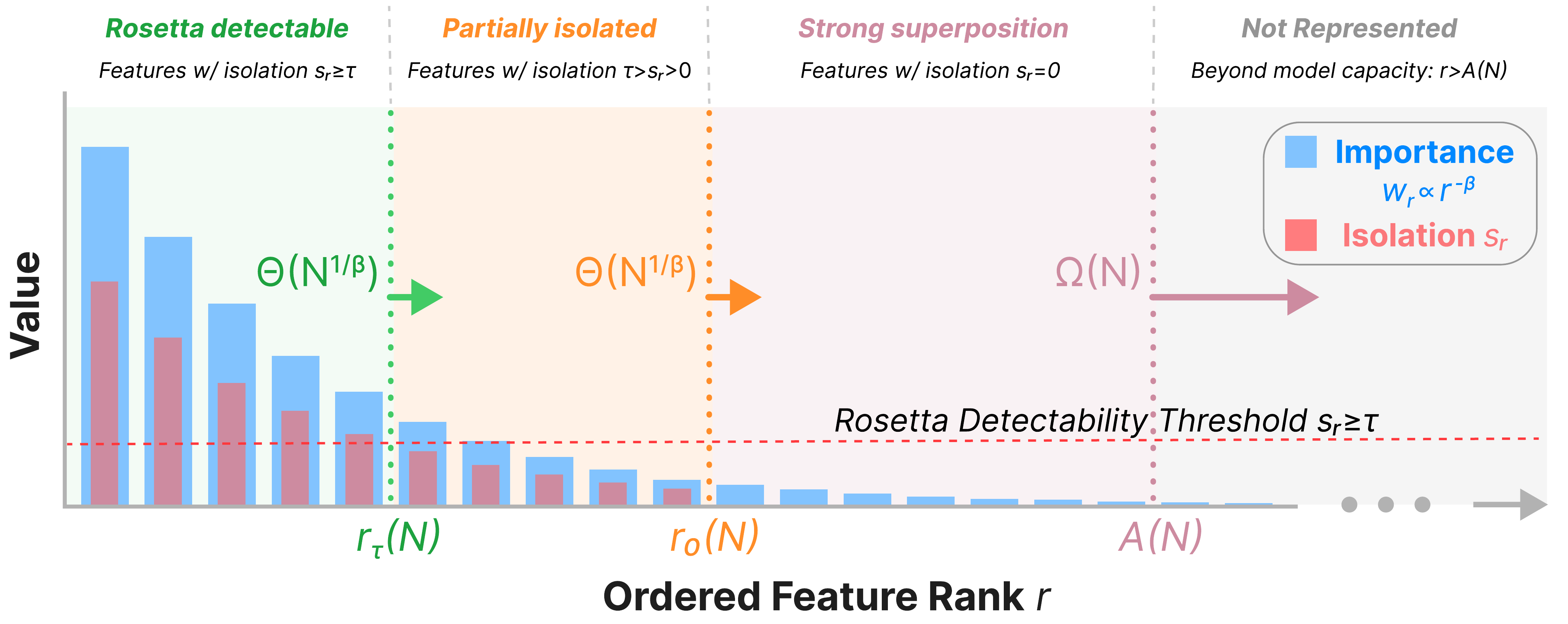}
    \vspace{-0.2cm}
    \caption{\textbf{Feature-isolation frontiers.}
    Features are ordered by decreasing importance $w_r \propto r^{-\beta}$.
    The optimal allocation partitions the spectrum into Rosetta-detectable features
    with $s_r \ge \tau$, partially isolated features with $0 < s_r < \tau$,
    strongly superposed features with $s_r = 0$, and features beyond the represented
    set $A(N)$. The frontiers $r_\tau(N)$ and $r_0(N)$ scale as
    $\Theta(N^{1/\beta})$, yielding the sublinear Rosetta Neuron count
    $R_\tau(N)=\Theta(N^{1/\beta})$.}
    \label{fig:feature-isolation-frontier}
    \vspace{-1em}
\end{figure}

\textbf{Capacity-allocation objective.} Given this setup, we model the network
as solving an allocation problem with two ingredients. First, detecting a feature
from a noisy single-neuron activation has diminishing returns: making a poorly
represented feature cleaner is valuable, but further purifying an already-clean
feature helps less. Second, the total budget for isolating features in single
neurons scales linearly with the number of coordinates \(N\). This yields:
\begin{equation}
    \max_{s_r\ge 0}\sum_r w_r\log(1+s_r)
    \qquad \text{s.t.}\qquad
    \sum_r s_r\le \kappa N .
\end{equation}
In~\Cref{sec:theory_supp}, we derive the logarithmic utility from a Gaussian channel model in which the neuron most aligned with feature \(r\) provides a noisy estimate of that feature's activation. Increasing \(s_r\) reduces the optimal negative log-likelihood by an amount proportional to \(\log(1+s_r)\).
The linear budget follows from a bounded-activation-energy condition in the underlying linear superposition model. Solving this allocation problem in the continuum limit yields a simple frontier structure:
\begin{equation}
    s^\star(r;N)
    =
    \left[
        \left(\frac{r_0(N)}{r}\right)^\beta - 1
    \right]_+,
    \qquad [u]_+ = \max(u,0).
\end{equation}
so features below a cutoff rank \(r_0(N)\) are more cleanly isolated in individual neurons, while features beyond it remain in superposition with $s_r=0$. This cutoff is set by the total budget:
\begin{equation}    
    \kappa N
    =
    \int_1^{r_0}
    \left[
        \left(\frac{r_0}{r}\right)^\beta - 1
    \right]dr
    =
    \Theta(r_0^\beta),
\end{equation}

and hence \(r_0(N)=\Theta(N^{1/\beta})\). Let \(R_\tau(N)\) count threshold-crossing features, which correspond to Rosetta
Neurons in this idealized model. Solving \(s^\star(r_\tau;N)=\tau\) for $r_\tau$ yields
\begin{equation}    
    r_\tau(N)=r_0(N)(1+\tau)^{-1/\beta},
    \qquad
    R_\tau(N)=\Theta(r_\tau(N))=\Theta(N^{1/\beta}).
    \label{eq:rosetta_scaling_main}
\end{equation}

Since \(\beta>1\), the Rosetta count grows as a sublinear power law. Figure~\ref{fig:feature-isolation-frontier} summarizes this frontier structure: as model size \(N\) increases, the Rosetta-detectability frontier moves outward, while a long tail of lower-importance features remains either partially isolated or strongly superposed with \(s_r=0\).

\textbf{Prediction: Neuron Polarization.} As network scale
increases, the Rosetta frontier moves outward in rank, but the features already inside
the frontier become cleaner. Averaging the optimal allocation over the Rosetta
set and the non-Rosetta tail, respectively, gives
\begin{equation}    
    \bar{s}_{\mathrm{Rosetta}}(N)
    =
    \Theta\!\left(N^{(\beta-1)/\beta}\right),
    \qquad
    \bar{s}_{\mathrm{non\text{-}Rosetta}}(N)
    =
    O\!\left(\frac{N^{1/\beta}}{A(N)}\right)
    \to 0 .
\end{equation}

Thus, Rosetta Neurons become increasingly clean and selective, while the unresolved feature
tail remains packed in a polysemantic background. We validate this prediction in the next section.

\section{Properties of Rosetta Neurons}
\label{sec:properties}

Having established Rosetta Neuron scaling laws, we next investigate how their properties change with scale. We test the predicted Neuron Polarization Effect across modalities, examine semantic specialization in language models, and conclude with a case study demonstrating that an individual Rosetta Neuron can be both selective and specialized enough to support domain-specific data filtering. Qualitative examples in~\Cref{subsec:non-rosetta-vs-rosetta} show the same pattern, with Rosetta Neurons firing on more interpretable and coherent concepts compared to non-Rosetta neurons.


\begin{figure}[t]
    \centering
    \begin{subfigure}[b]{0.49\textwidth}
        \centering
        \includegraphics[width=\textwidth]{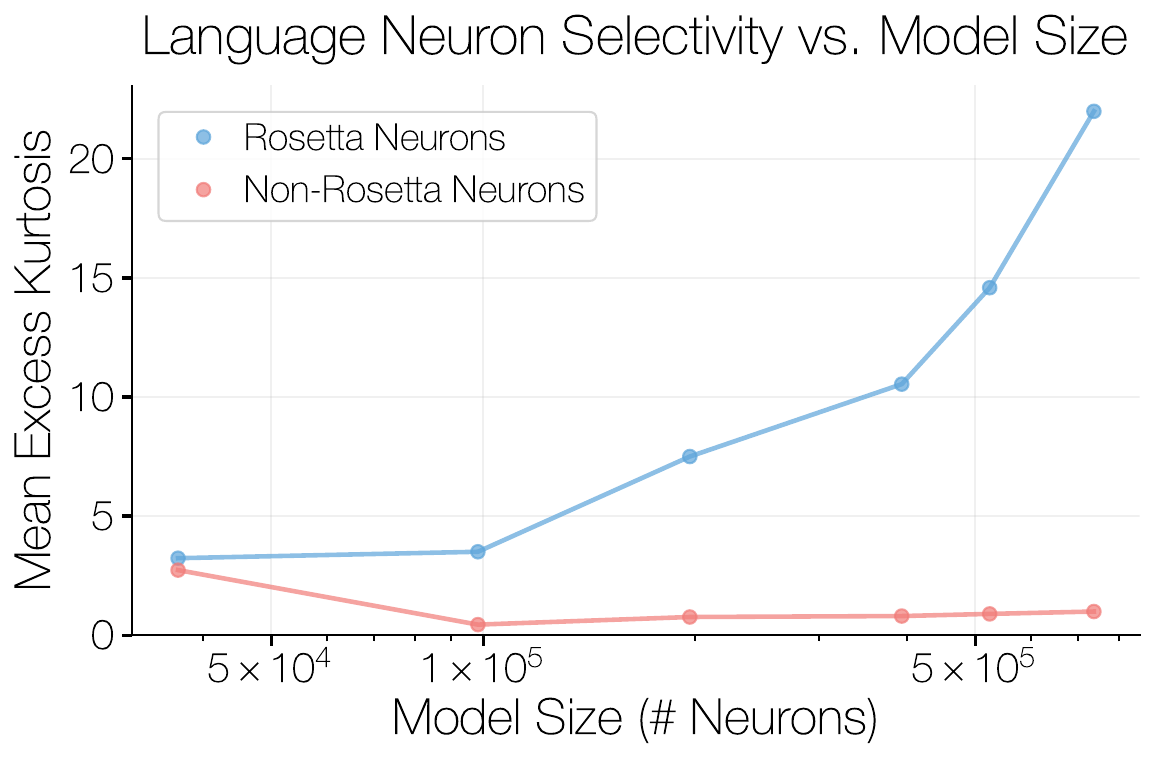}
        \caption{Vocabulary-space neuron selectivity in Pythia.}
        \label{fig:language_kurtosis}
    \end{subfigure}
    \hfill 
    \begin{subfigure}[b]{0.49\textwidth}
        \centering
        \includegraphics[width=\textwidth]{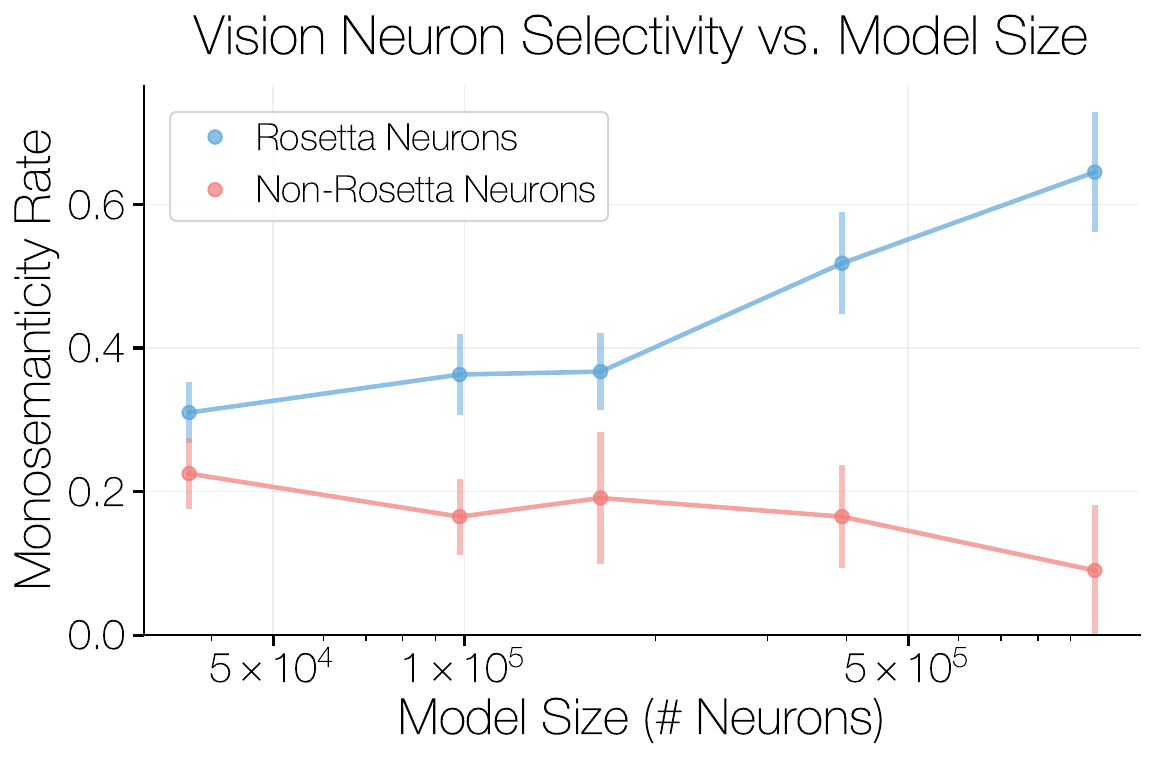}
        \caption{VLM-judged monosemanticity rate in OpenCLIP.}
        \label{fig:vision_monosemanticity}
    \end{subfigure}
    
    \caption{\textbf{The Neuron Polarization Effect in language and vision models.} (a) In language models, Rosetta Neurons show increasing mean excess kurtosis of vocabulary-space projections with scale. Non-Rosetta neurons remain near zero, indicating weak selectivity. (b) In vision models, VLM-judged monosemanticity increases with scale for Rosetta Neurons and decreases for non-Rosetta neurons.}
    \label{fig:neuron_polarization_effect_figure}
\end{figure}
\subsection{The Neuron Polarization Effect in Large-Scale Models}
\label{polsemanticity_section}

We next test the predicted Neuron Polarization Effect: scaling separates increasingly monosemantic Rosetta Neurons from a comparatively polysemantic non-Rosetta background. We observe this trend using heuristic neuron selectivity measures in both language and vision models.

\textbf{Language models.} We measure language-model neuron selectivity through vocabulary-space projections~\citep{gurneeuniversal}. For each neuron \(u=(\ell,c)\), we compare its output weight \(W_{\mathrm{out}}^{u}\) to each token's unembedding vector \(W_U[v]\) via cosine similarity:
$s_v^u=\cos\!\left(W_{\mathrm{out}}^{u},\,W_U[v]\right).$ We use the excess kurtosis of these alignments as the neuron’s output-side selectivity metric, where higher values indicate concentration on a small set of tokens. In Pythia, we compute the mean of this metric separately over Rosetta and non-Rosetta neurons, defining the Rosetta set at each scale using the OPT--Pythia matches. We plot this metric as a function of model size in~\Cref{fig:language_kurtosis}, and find that the Rosetta mean increases with scale, indicating greater concentration on a small set of tokens and more coherent, monosemantic functionality~\citep{avrahamy2026disentangling}. In contrast, the non-Rosetta mean decreases and remains near zero, consistent with weak selectivity and a more polysemantic population. We observe the same qualitative trend in other language model families (\Cref{sec:rosetta_properties_supp}).

\textbf{Vision models.}  Since vision models do not admit the same natural vocabulary-space selectivity measure as language models, we use a VLM-as-a-judge proxy for monosemanticity, building on the established practice of using multimodal models to describe neurons~\citep{oikarinenclip, shaham2024multimodal}. At each scale, we evaluate the OpenCLIP neurons identified in the Diffusion--OpenCLIP Rosetta matches. For each neuron, we present GPT-5.4 with its top 20 activating images, activation maps, and overlays, and ask whether the responses reflect a single coherent visual feature. We evaluate five disjoint subsets of 100 Rosetta Neurons and five matched subsets of non-Rosetta neurons. In~\Cref{fig:vision_monosemanticity}, we plot the fraction of neurons judged monosemantic as OpenCLIP model size scales. This fraction increases with scale for Rosetta Neurons and decreases for non-Rosetta neurons, consistent with the Neuron Polarization Effect observed in language models. Additional results on other models, methodological details, and a metric reliability study are provided in~\Cref{sec:vlm_appendix}.

\subsection{Rosetta Neurons Become More Specialized with Scale in Language Models}
\label{specialization_section}
We now investigate whether Rosetta Neurons become more specialized with scale by measuring their top-activating document types across Pythia models of different sizes. 
Our analytical model predicts that as the Rosetta frontier expands, lower-ranked features can become sufficiently clean and reproducible to be shared across independently trained models. As rarer data patterns lie deeper in the feature importance spectrum, larger models should increasingly contain Rosetta Neurons selective for specialized domains, appearing as a scale-dependent shift in document-type firing.

\textbf{Measuring Rosetta Neuron firing by document type.} We measure how Rosetta Neuron firing patterns vary across text categories with model size. At each Pythia scale, we use Rosetta Neurons from the OPT–Pythia scaling runs and collect each neuron’s top-20 activating contexts from the Pile validation set. We assign each context to one of five categories: code, math, formal/scientific text, general prose, or conversational text. For each category, we compare its share of Rosetta Neuron top activations with its token share in the validation cache. A value of 1 corresponds to the corpus baseline, where top Rosetta activations fall in a category at the same fraction as that category’s token share in the dataset. Additional details on the category construction are in~\Cref{sec:rosetta_properties_supp}.

\begin{figure}[t]
    \centering
    \includegraphics[width=0.99\linewidth]{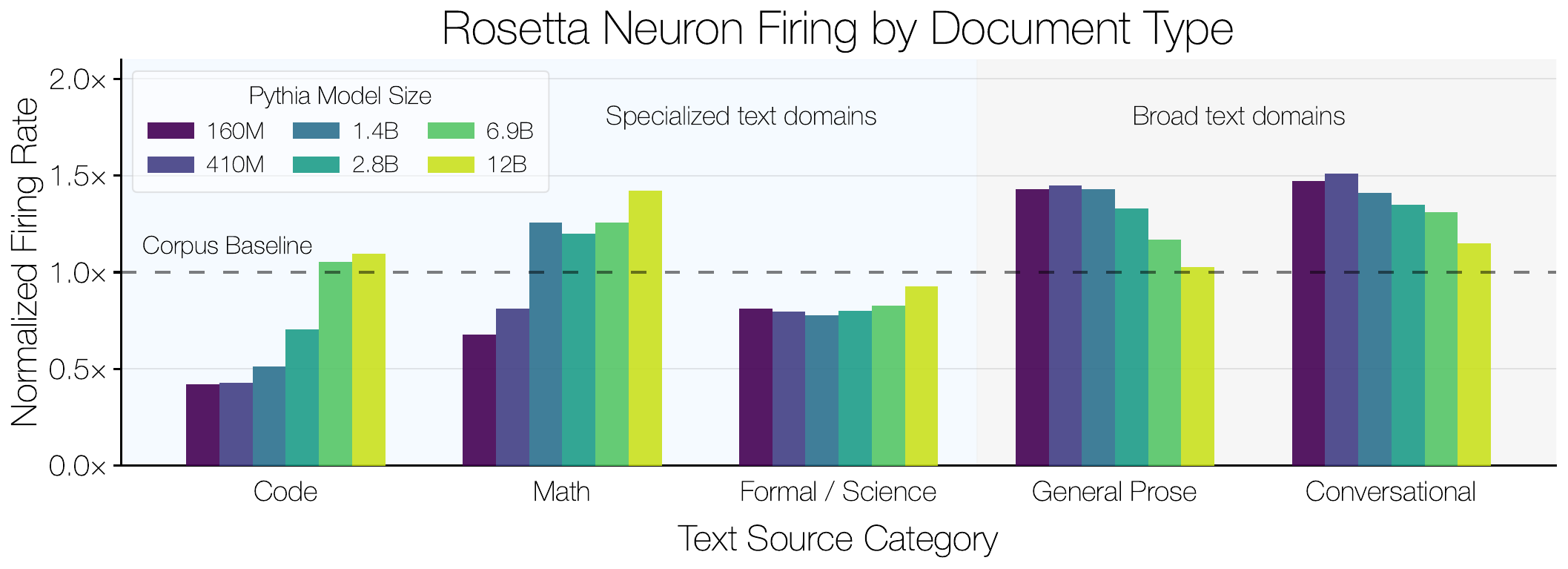}
    \vspace{-0.5em}
   \caption{\textbf{Rosetta Neuron document-type firing in Pythia.} For each Pythia model size, we plot how often top-activating Rosetta Neuron contexts fall into a document category, normalized by that category's token frequency in the validation set. The dashed line marks the corpus baseline. With scale, Rosetta Neuron firing shifts toward specialized categories such as code and math.}
    \label{fig:rosetta_firing_docs}
\end{figure}

\textbf{Rosetta Neurons shift toward specialized domains with scale.}~\Cref{fig:rosetta_firing_docs} shows that Rosetta Neuron firing becomes increasingly concentrated on specialized text domains with model scale. Code and math become increasingly represented among top activations relative to their dataset frequency, while broader categories such as general prose and conversational text decline. Since all Pythia models were trained on the same data mixture and token budget, this shift cannot be explained by greater exposure to specialized data. Instead, scale appears to change how models organize internal representations: larger models do not just add shared neurons for basic features, but expand the Rosetta population toward increasingly specialized domains. Although our results are aggregated over many Rosetta Neurons, it is consistent with qualitative examples in~\Cref{sec:qualitative_supp}, where individual Rosetta Neurons become increasingly domain-selective rather than firing broadly across text types. In contrast, a non-Rosetta neuron baseline in~\Cref{sec:rosetta_properties_supp} shows fairly stable document-type preferences across scale, with no corresponding shift toward specialized domains. In the next section, we move from population-level specialization to a single-neuron case study, testing whether an individual Rosetta Neuron is both selective and specialized enough to support domain-specific data filtering.

\subsection{Testing Rosetta Neuron Selectivity with Data Filtering}
\label{sec:data_filtering}
 We next study specialization at the single-neuron level, testing whether an individual Rosetta Neuron can identify a coherent, specialized domain well enough to support data filtering. In a controlled code-domain setting, we use ground-truth labels to evaluate domain recovery and continued pretraining to test downstream utility.

\textbf{Data-filtering setup.} We use CodeSearchNet~\citep{husain2019codesearchnet}, a multilingual code corpus with source-language labels. We take JavaScript as a representative target domain, whose training split contains roughly 58K parsed code functions totaling 16M tokens. Each filtering method selects functions from the multilingual pool under a 16M token budget, and we evaluate recovery by F1 score against the ground-truth JavaScript subset. We then test how well the recovered data supports learning the target domain. We continue pretraining GPT2-1.5B, which has limited code exposure, on the selected data and report JavaScript test-set perplexity. 

Under this matched token-budget setup, we compare four filtering methods. The Rosetta Neuron filter uses a single JavaScript-selective Rosetta Neuron from Pythia-6.9B, discovered in the Pythia–OPT matching runs on the Pile. We score CodeSearchNet training functions by this neuron’s activations and select the highest-scoring functions up to the 16M-token budget.  We apply the same procedure to a non-Rosetta neuron that activates on JavaScript text, identified on the same Pile cache. We compare these neuron-based filters against two baselines: a uniform random sample from the multilingual pool and the oracle JavaScript subset. Additional experimental details are in~\Cref{sec:data_filter_supp}.

\begin{wraptable}{r}{0.35\textwidth}
\centering
\resizebox{\linewidth}{!}{%
\begin{tabular}{lcc}
\toprule
\textbf{Method} & \textbf{F1} & \textbf{Test PPL} \\
\midrule
Base Model & -- & 6.73 \\
\cmidrule(lr){1-3}
Random & 0.06 & $3.59\pm{0.07}$ \\
Non-Rosetta & 0.09 & $3.23\pm{0.07}$ \\
Rosetta & $0.98$ & $3.02\pm{0.05}$ \\
Oracle & $1.00$ & $3.01\pm{0.04}$ \\
\bottomrule
\end{tabular}
}
\vspace{0.2em}
\caption{\textbf{JavaScript filtering.} Matched-budget filters on CodeSearchNet; PPL reports mean $\pm$ 95\% CI over three runs.}
\label{tab:js_filtering}
\end{wraptable}

\textbf{Rosetta Neuron selectivity predicts useful target-domain data.} \Cref{tab:js_filtering} reports each filter’s F1 recovery of the ground-truth JavaScript subset and test-set perplexity over three training runs. All filters improve over the base model, consistent with GPT2-1.5B's limited code exposure. Notably, the Rosetta filter recovers nearly the entire JavaScript subset, achieving 0.98 F1. This selectivity translates into downstream utility: continued pretraining on Rosetta-filtered data reduces test perplexity from 3.59 for random filtering to 3.02, nearly matching the oracle at 3.01. In contrast, the non-Rosetta JavaScript neuron improves over random filtering but is less selective and yields weaker downstream gains. Qualitative examples in~\Cref{sec:data_filter_supp} suggest that the non-Rosetta neuron is less specialized, firing on broader web-programming syntax and JavaScript-adjacent languages rather than specifically JavaScript. While this single-neuron experiment serves as a controlled case study of downstream data filtering, qualitative examples in~\Cref{sec:qualitative_supp} suggest that other Rosetta Neurons exhibit similarly interpretable, selective firing.


\section{Discussion, Limitations, and Future Work}
\label{sec:discussion}

We studied whether the internal organization of neural networks evolves predictably with scale. Using Rosetta Neurons as a probe, we found sublinear power-law growth of shared single-neuron structure in language and vision models. Together, our analytical model and empirical measurements suggest that
scaling polarizes neurons: a shared Rosetta population grows more selective, while the remaining neurons form a larger, superposed background. We further showed that Rosetta Neurons become more specialized with scale, demonstrating this specialization in a controlled data-filtering study for continued pretraining. We close by discussing the limitations and scope of the analysis, along with future directions; additional discussion and failure cases are in~\Cref{sec:further_limitations}.

Our analysis targets a specific form of neuron-level structure: Rosetta Neurons. This is only a subset of the structure inside neural networks. Some computations may be carried by circuits~\citep{elhage2021mathematical}, attention heads~\citep{olsson2022context}, or subspaces that don't appear as neuron correspondences~\citep{wang2018towards,bricken2023emergence}. Thus, the Rosetta population should not be viewed as a map of all shared computation. We instead view Rosetta Neurons as a tractable observable: a shared neuron-level population for studying structure that is not specific to any single training run, architecture, or model family. Combining universality with scale makes Rosetta Neurons a powerful discovery tool: cross-model recurrence identifies candidate interpretable units without target concepts or auxiliary training, while comparisons across model size expose predictable changes in this structure. However, the resulting trends should be interpreted within the scope of our measurements. For instance, our monosemanticity measurements rely on modality-specific proxies, and should be viewed as relative population-level trends. 

More broadly, our results point toward a bridge between macroscopic scaling behavior and the microscopic organization of learned representations. Standard scaling laws describe external quantities such as loss as a function of scale, while Rosetta Neurons provide a complementary lens on how shared structure can evolve predictably at the level of individual units. This opens several directions: discovering other internal observables whose structure changes predictably with scale, studying how universal neurons emerge during training, how they are transformed by post-training, and how they relate to distributed forms of shared computation. Our analytical model offers a simple account of Rosetta Neuron scaling and polarization, but deriving these phenomena from gradient-based training dynamics remains an important next step. Such an account could explain how universal structure emerges from optimization, and may inform objectives or regularizers that shape not only loss scaling but also the internal organization and adaptability of learned representations.

\textbf{Acknowledgments.} We thank members of Berkeley AI Research and the Redwood Center for Theoretical Neuroscience for helpful discussions. We are particularly grateful to Mason Kamb, Phillip Isola, David Bau, Yizhou Liu, Sophie Wang, Grace Luo, Stephanie Fu, Jasmine Shone, Tamar Rott Shaham, and Tyler Bonnen for their thoughtful feedback.  YB is a visiting scholar at UC Berkeley and member of the Simons Collaboration on the Physics of Learning \& Neural Computation. AD is supported by the US Department of Energy Computational Science Graduate Fellowship.  Additional support came from ONR MURI, NSF IIS-2403305, and the Google-BAIR Commons Program.

\bibliographystyle{neurips_2026}
\bibliography{neurips_2026}

\newpage
\appendix
\clearpage
\appendix

\section*{Appendix}

\appitem{A \quad Rosetta Neuron Visualizations}{sec:qualitative_supp}
\appsubitem{A.1 \quad Rosetta Neurons across Model Scales}{sec:qualitative_supp}
\appsubitem{A.2 \quad Qualitative Comparison of Rosetta and Non-Rosetta Neurons}{subsec:non-rosetta-vs-rosetta}

\appitem{B \quad Aligning Token-Wise Activations Across Models}{sec:aligning_tokens}

\appitem{C \quad Further Details on Rosetta-Neuron Scaling}{sec:rosetta_neuron_scaling_appendix}
\appsubitem{C.1 \quad Model Families}{app:model_families}
\appsubitem{C.2 \quad Robustness to the Mutual Top-\(k\) Criterion}{subsec:topk_ablation}
\appsubitem{C.3 \quad Input-Permutation Null}{supp:permutation_null}

\appitem{D \quad Detailed Derivation of the Rosetta Neuron Scaling Model}{sec:theory_supp}
\appsubitem{D.1 \quad A Minimal Capacity-Allocation Model}{app:minimal-allocation-model}
\appsubitem{D.2 \quad Solving the Allocation Problem}{app:solving-allocation}
\appsubitem{D.3 \quad Prediction: Neuron Polarization}{app:polarization-prediction}
\appsubitem{D.4 \quad Deriving the Isolation Score from Superposition}{app:carrier-sir-model}
\appsubitem{D.5 \quad Deriving the Logarithmic Utility and Linear Isolation Budget}{app:objective-and-budget-microfoundation}
\appsubitem{D.6 \quad Connection to Empirical Rosetta Matching}{supp:empirical-matching}
\appsubitem{D.7 \quad Synthetic Validation of the Analytical Model}{supp:simulation}

\appitem{E \quad Additional Results on Rosetta Neuron Properties}{sec:rosetta_properties_supp}
\appsubitem{E.1 \quad Additional Vocabulary-Space Selectivity Results}{app:kurtosis_results}
\appsubitem{E.2 \quad Document-Type Firing Analysis}{app:document_firing_details}
\appsubitem{E.3 \quad Depth-Wise Distribution of Rosetta Neurons}{app:depthwise_rosetta}

\appitem{F \quad Data Filtering Experimental Details}{sec:data_filter_supp}

\appitem{G \quad Dataset Ablations}{sec:dataset_ablations}
\appsubitem{G.1 \quad Ablation on the Number of Tokens Used for Language Model Matching}{sec:dataset_ablations}
\appsubitem{G.2 \quad Ablation on the Number of Images Used for Vision Model Matching}{sec:dataset_ablations}
\appsubitem{G.3 \quad Ablation on the Image Distribution Used for Vision Model Matching}{sec:dataset_ablations}

\appitem{H \quad Additional Details on VLM-as-a-Judge}{sec:vlm_appendix}
\appsubitem{H.1 \quad Detailed Experimental Setup}{sec:detailed_vlm_setup}
\appsubitem{H.2 \quad Sensitivity to the Number of Top-\(k\) Activating Images}{sec:topk_sensitivity}
\appsubitem{H.3 \quad Validation of VLM-as-a-Judge as a Predictive Metric}{sec:vlm_validation}
\appsubitem{H.4 \quad Results for Neuron Selectivity in Other Vision Models}{sec:more_vision_selectivity}

\appitem{I \quad Further Discussion and Limitations}{sec:further_limitations}
\appsubitem{I.1 \quad DINOv3 as an Informative Failure Case}{sec:dinov3_failure}
\appsubitem{I.2 \quad Operationalizing Monosemanticity and Polysemanticity}{sec:mono_vs_poly}

\appitem{J \quad Compute Resources}{sec:compute}

\clearpage

\section{Rosetta Neuron Visualizations}
\label{sec:qualitative_supp}

\subsection{Rosetta Neurons across Model Scales}
\label{subsec:qualitative_scale}
\begin{figure}[h]
    \centering
    \vspace{1em}
    
    \begin{subfigure}[b]{0.49\textwidth}
        \centering
        \includegraphics[width=\textwidth]{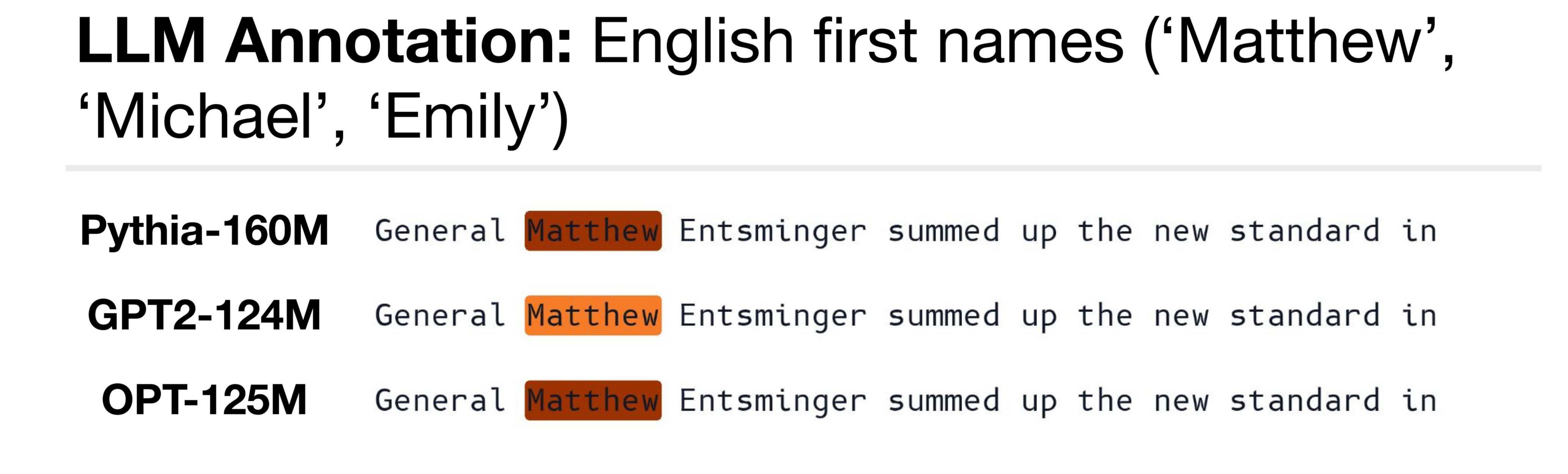}
        \caption{Pythia-160M: L5/U1767. GPT2-124M: L7/U1625. OPT-125M: L1/U2325.}
        \label{fig:llm_qualitative0_a}
    \end{subfigure}
    \hfill
    \begin{subfigure}[b]{0.49\textwidth}
        \centering
        \includegraphics[width=\textwidth]{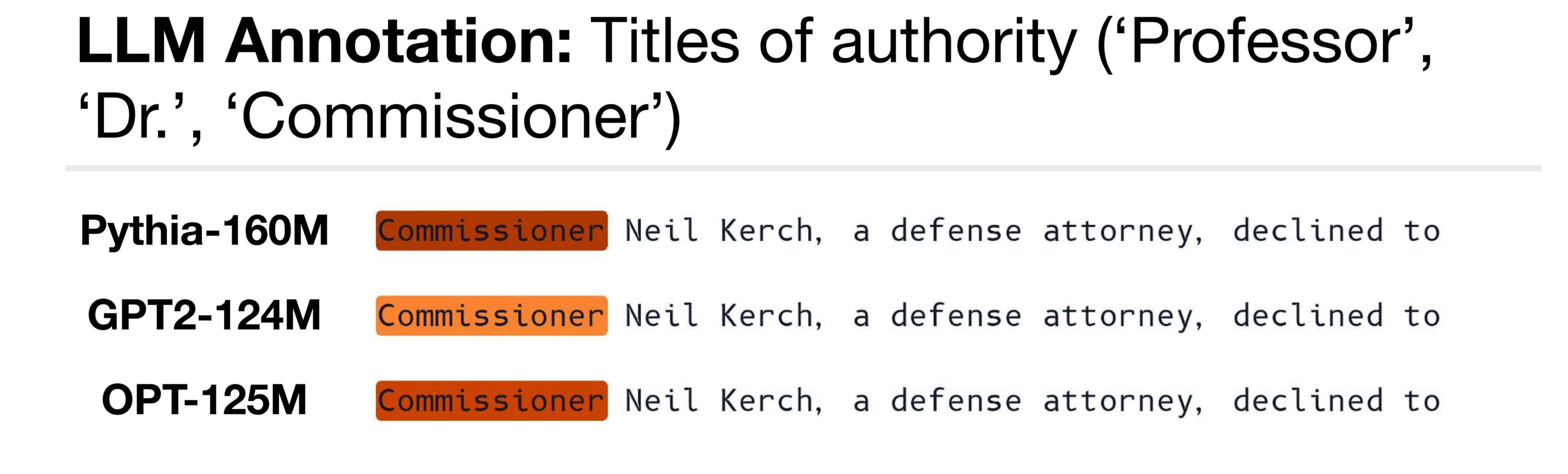}
        \caption{Pythia-160M: L5/U1898. GPT2-124M: L8/U557. OPT-125M: L8/U2115.}
        \label{fig:llm_qualitative0_b}
    \end{subfigure}
    
    \vspace{1em}
    
    \begin{subfigure}[b]{0.49\textwidth}
        \centering
        \includegraphics[width=\textwidth]{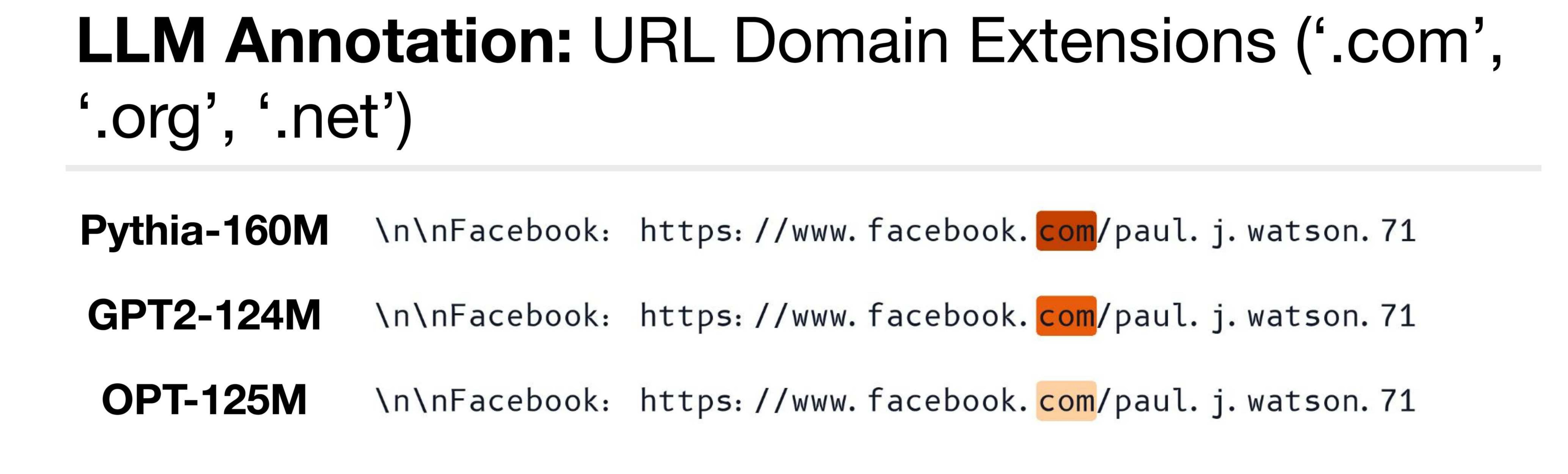}
        \caption{Pythia-160M: L10/U2431. GPT2-124M: L11/U424. OPT-125M: L8/U2837.}
        \label{fig:llm_qualitative0_c}
    \end{subfigure}
    \hfill
    \begin{subfigure}[b]{0.49\textwidth}
        \centering
        \includegraphics[width=\textwidth]{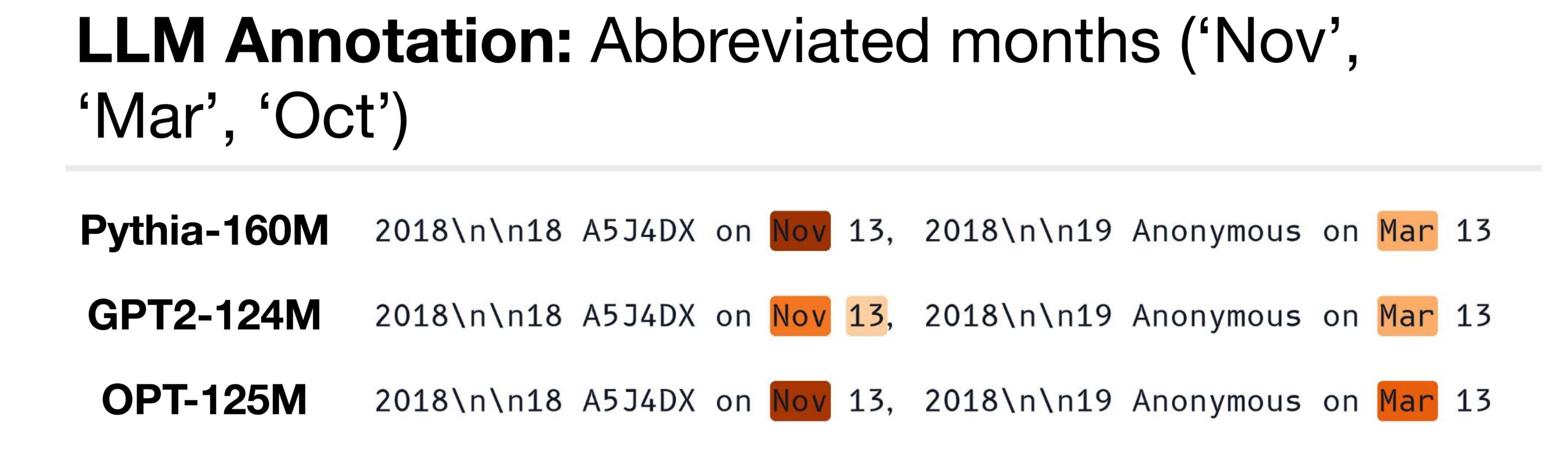}
        \caption{Pythia-160M: L0/U2626. GPT2-124M: L9/U649. OPT-125M: L0/U888.}
        \label{fig:llm_qualitative0_d}
    \end{subfigure}
    
    \caption{\textbf{LLM annotations for Rosetta Neurons.} Comparison between Pythia-160M, GPT2-124M, OPT-125M.}
    \label{fig:llm_qualitative0}
    \vspace{0.5em}
\end{figure}

\begin{figure}[h]
    \centering
    
    \begin{subfigure}[b]{0.49\textwidth}
        \centering
        \includegraphics[width=\textwidth]{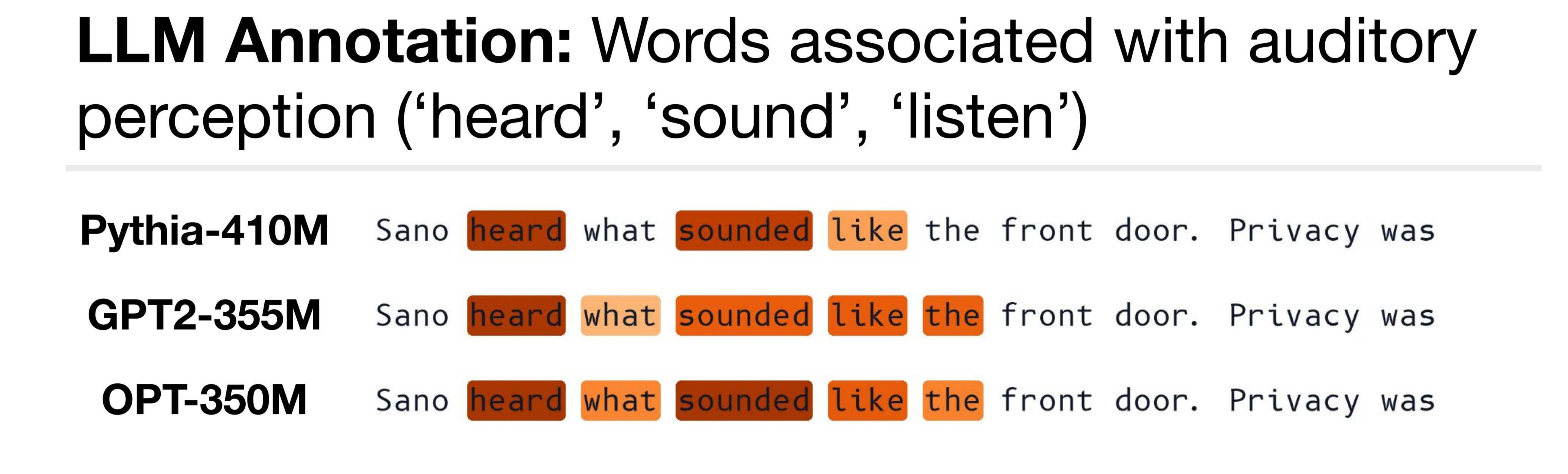}
        \caption{Pythia-410M: L10/U3047. GPT2-355M: L19/U1698. OPT-350M: L12/U927.}
        \label{fig:llm_qualitative1_a}
    \end{subfigure}
    \hfill
    \begin{subfigure}[b]{0.49\textwidth}
        \centering
        \includegraphics[width=\textwidth]{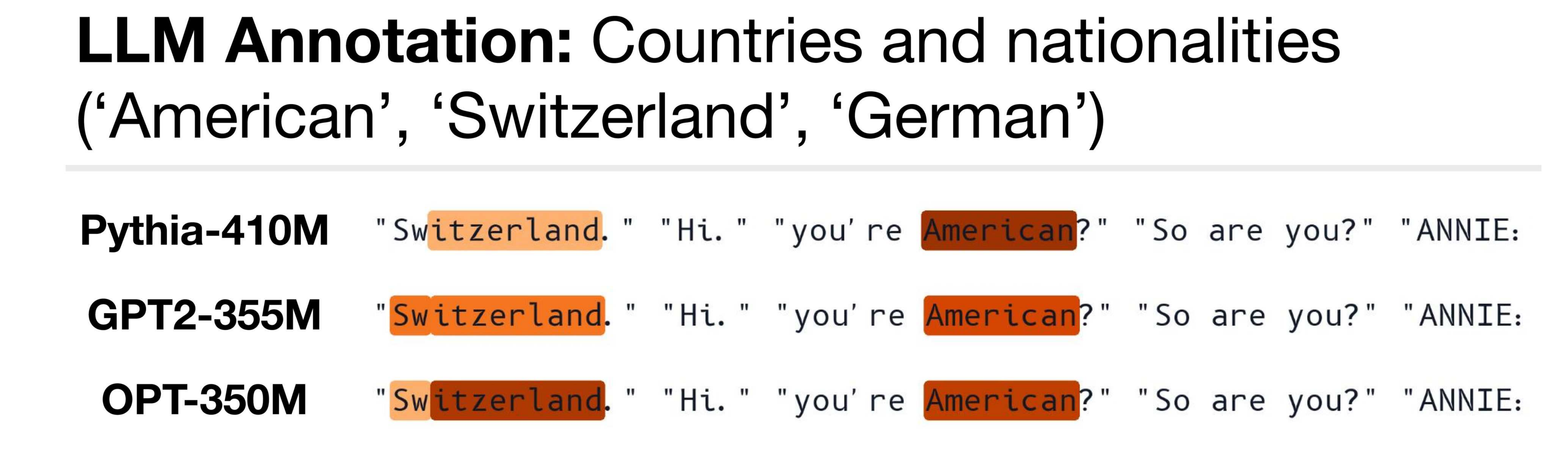}
        \caption{Pythia-410M: L9/U1693. GPT2-355M: L15/U1724. OPT-350M: L13/U846.}
        \label{fig:llm_qualitative1_b}
    \end{subfigure}
    
    \vspace{1em}
    
    \begin{subfigure}[b]{0.49\textwidth}
        \centering
        \includegraphics[width=\textwidth]{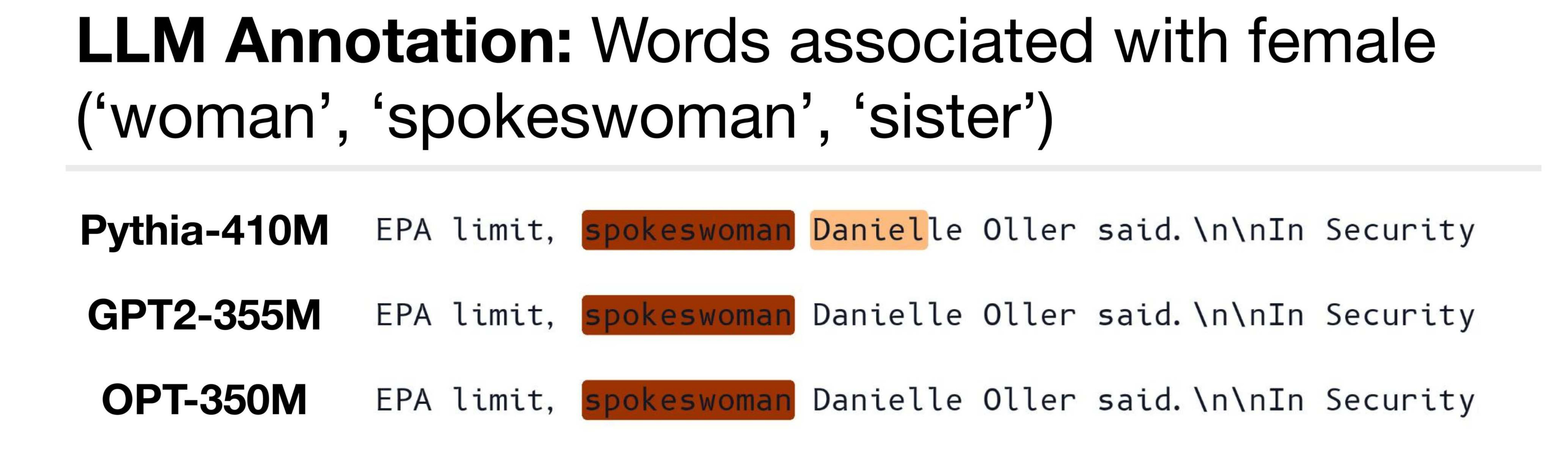}
        \caption{Pythia-410M: L23/U39. GPT2-355M: L23/U149. OPT-350M: L23/U2921.}
        \label{fig:llm_qualitative1_c}
    \end{subfigure}
    \hfill
    \begin{subfigure}[b]{0.49\textwidth}
        \centering
        \includegraphics[width=\textwidth]{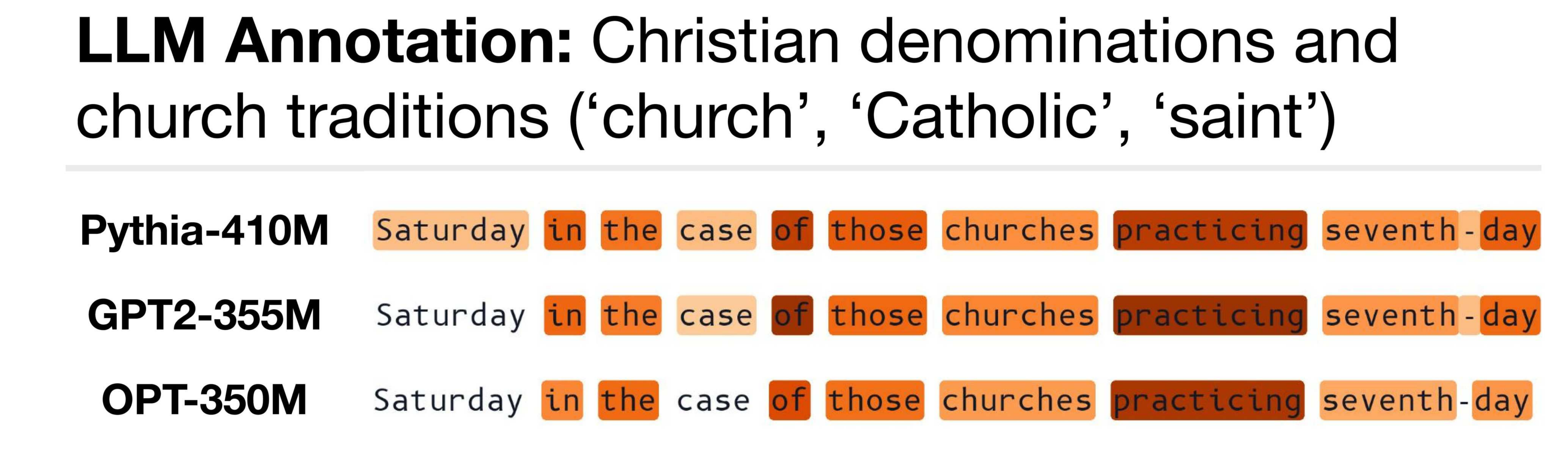}
        \caption{Pythia-410M: L23/U3863. GPT2-355M: L23/U1026. OPT-350M: L22/U1841.}
        \label{fig:llm_qualitative1_d}
    \end{subfigure}
    
    \caption{\textbf{LLM annotations for Rosetta Neurons.} Comparison between Pythia-410M, GPT2-355M, OPT-350M.}
    \label{fig:llm_qualitative1}
\end{figure}

\begin{figure}[p]
    \centering
    
    \begin{subfigure}[b]{0.49\textwidth}
        \centering
        \includegraphics[width=\textwidth]{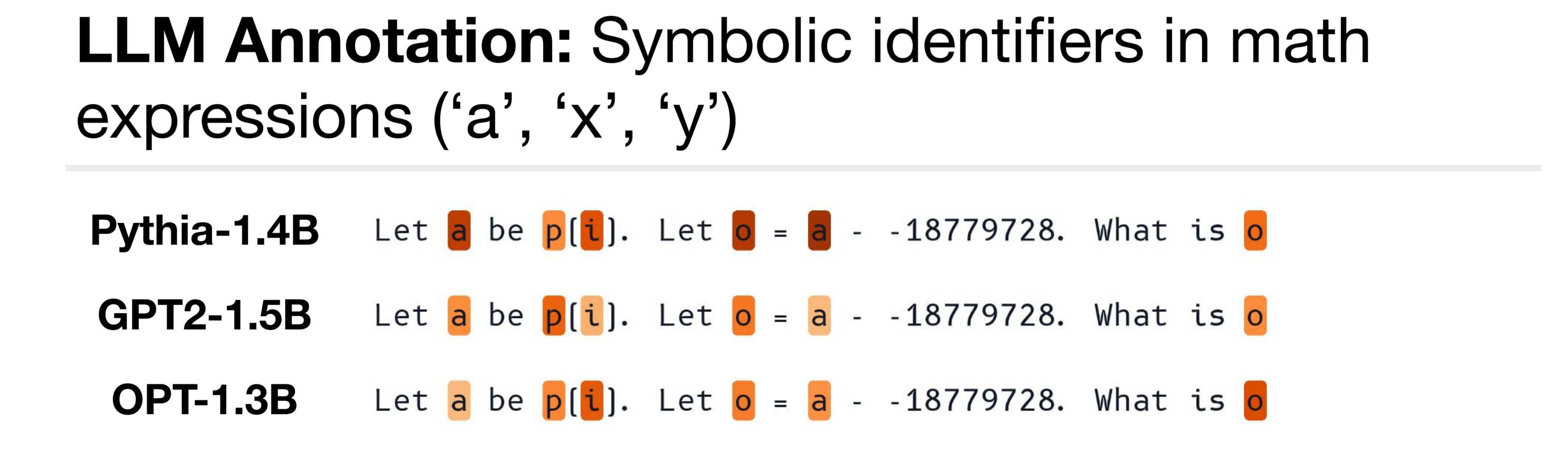}
        \caption{Pythia-1.4B: L2/U3564. GPT2-1.5B: L12/U1978. OPT-1.3B: L3/U652.}
        \label{fig:llm_qualitative2_a}
    \end{subfigure}
    \hfill
    \begin{subfigure}[b]{0.49\textwidth}
        \centering
        \includegraphics[width=\textwidth]{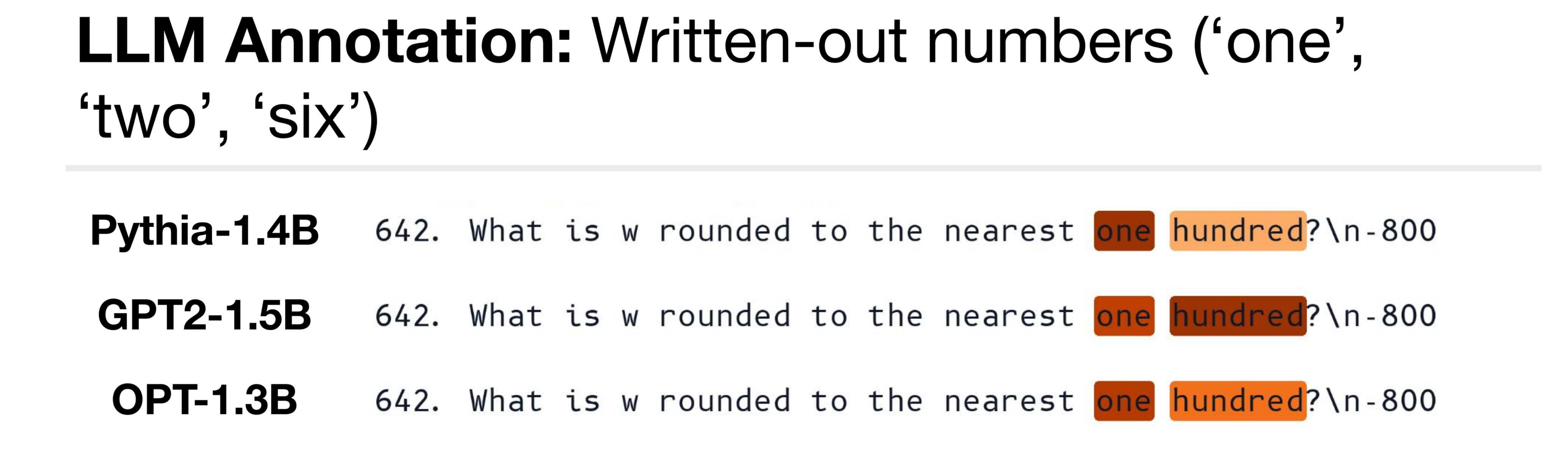}
        \caption{Pythia-1.4B: L9/U835. GPT2-1.5B: L19/U2508. OPT-1.3B: L17/U7443.}
        \label{fig:llm_qualitative2_b}
    \end{subfigure}
    
    \vspace{1em}
    
    \begin{subfigure}[b]{0.49\textwidth}
        \centering
        \includegraphics[width=\textwidth]{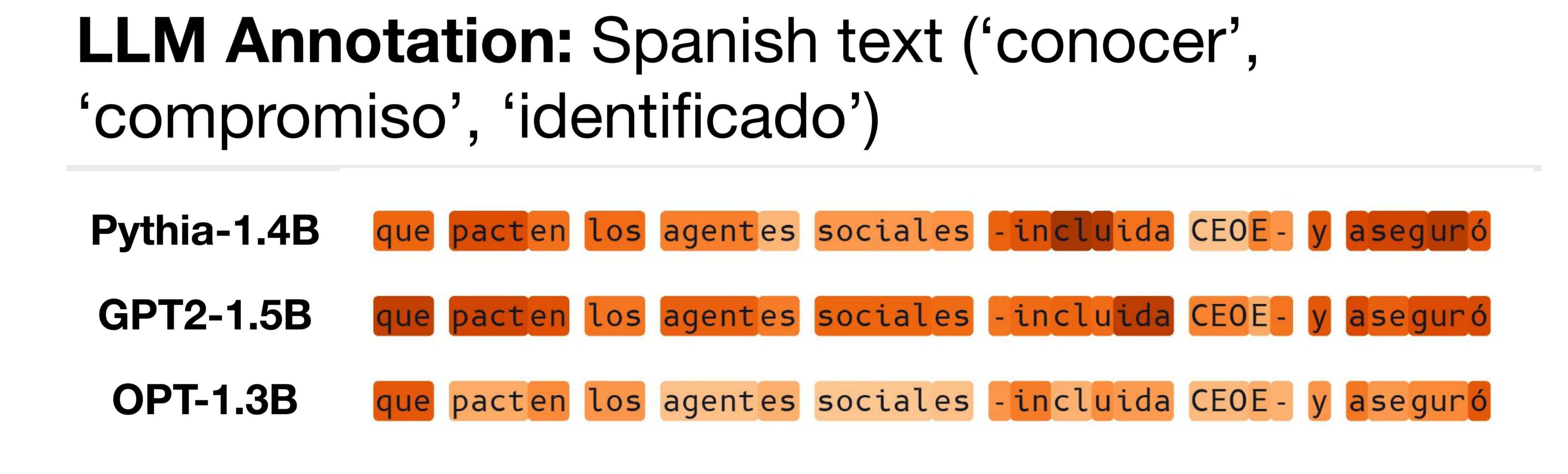}
        \caption{Pythia-1.4B: L15/U2969. GPT2-1.5B: L34/U491. OPT-1.3B: L17/U3726.}
        \label{fig:llm_qualitative2_c}
    \end{subfigure}
    \hfill
    \begin{subfigure}[b]{0.49\textwidth}
        \centering
        \includegraphics[width=\textwidth]{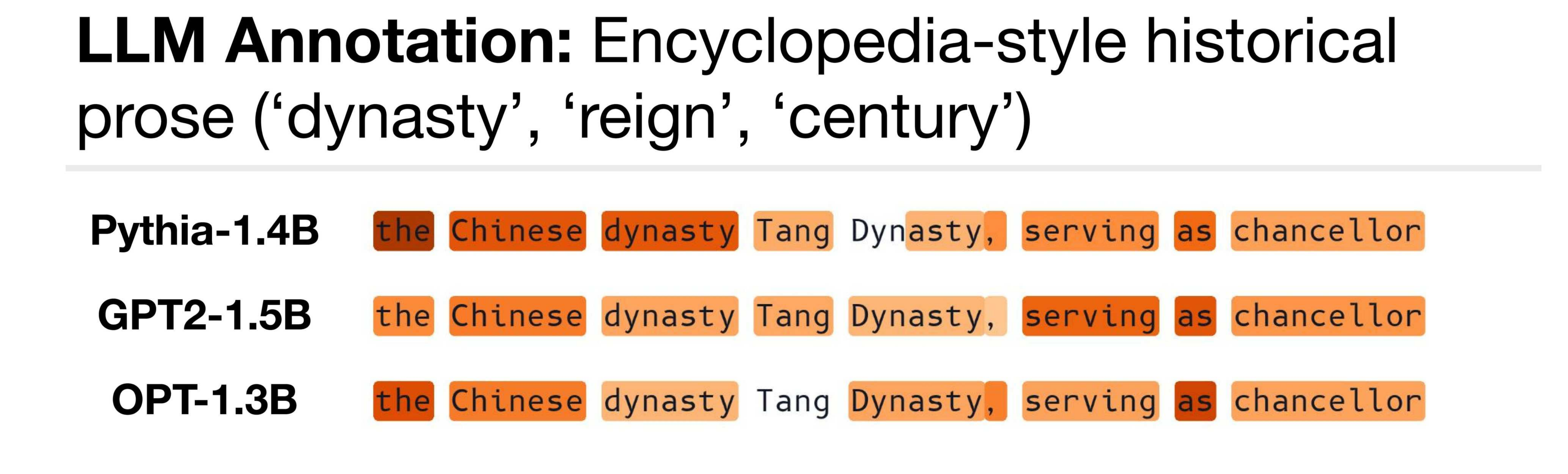}
        \caption{Pythia-1.4B: L15/U189. GPT2-1.5B: L35/U1327. OPT-1.3B: L18/U4373.}
        \label{fig:llm_qualitative2_d}
    \end{subfigure}
    
    \caption{\textbf{LLM annotations for Rosetta Neurons.} Comparison between Pythia-1.4B, GPT2-1.5B, OPT-1.3B.}
    \label{fig:llm_qualitative2}
\end{figure}

\begin{figure}[p]
    \centering
    
    \begin{subfigure}[b]{0.49\textwidth}
        \centering
        \includegraphics[width=\textwidth]{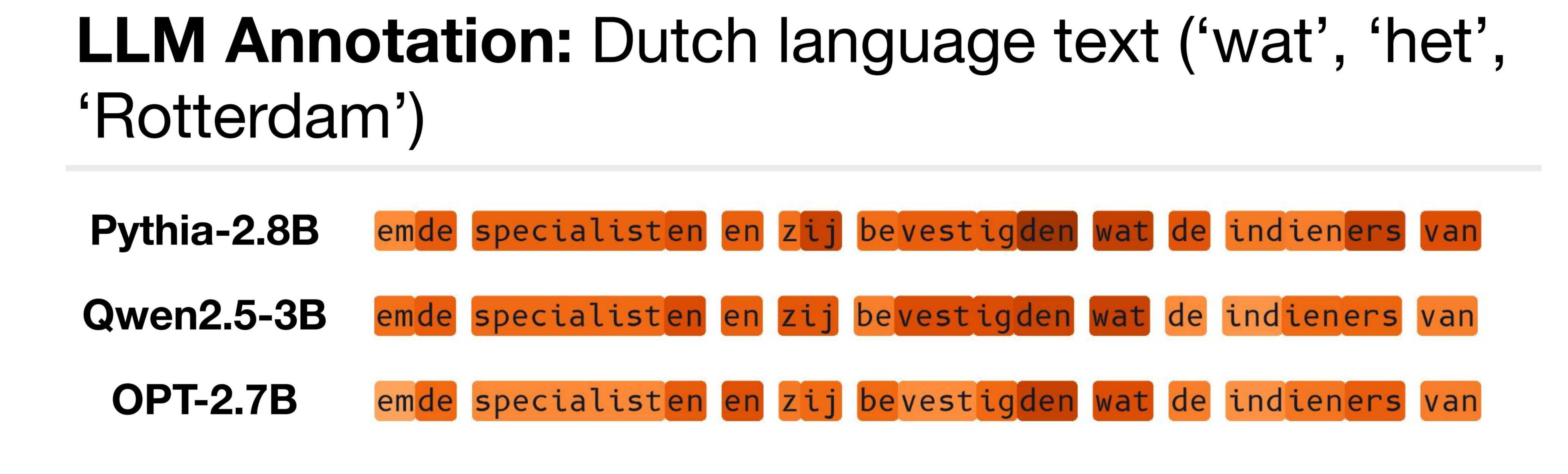}
        \caption{Pythia-2.8B: L20/U9199. Qwen2.5-3B: L31/U5812. OPT-2.7B: L25/U7525.}
        \label{fig:llm_qualitative3_a}
    \end{subfigure}
    \hfill
    \begin{subfigure}[b]{0.49\textwidth}
        \centering
        \includegraphics[width=\textwidth]{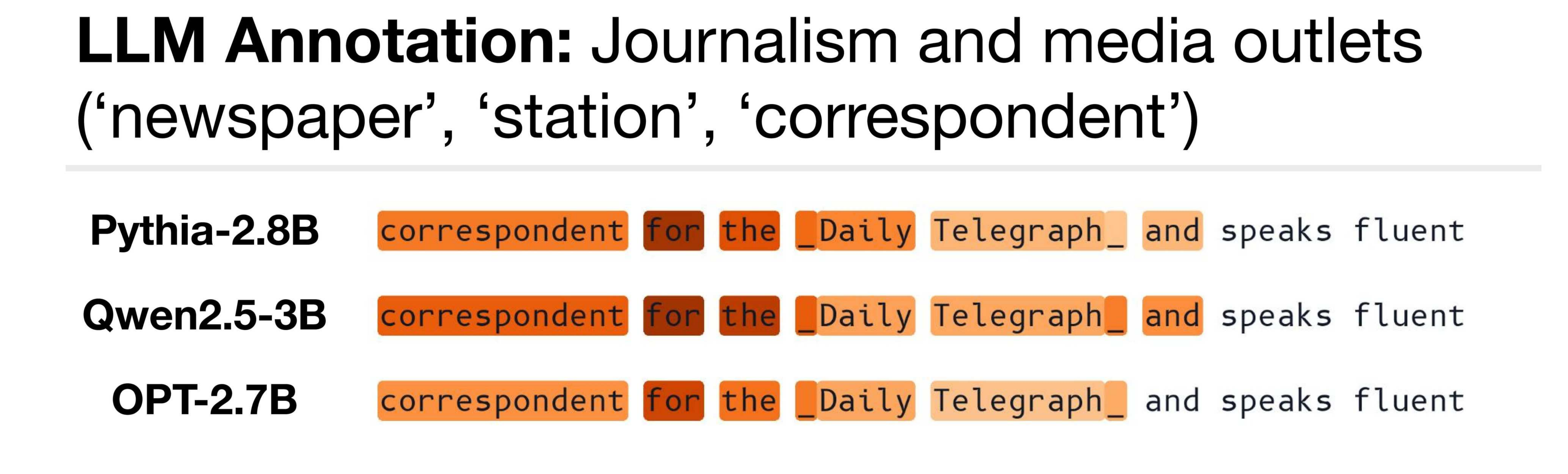}
        \caption{Pythia-2.8B: L22/U6182. Qwen2.5-3B: L31/U1823. OPT-2.7B: L27/U3916.}
        \label{fig:llm_qualitative3_b}
    \end{subfigure}
    
    \vspace{1em}
    
    \begin{subfigure}[b]{0.49\textwidth}
        \centering
        \includegraphics[width=\textwidth]{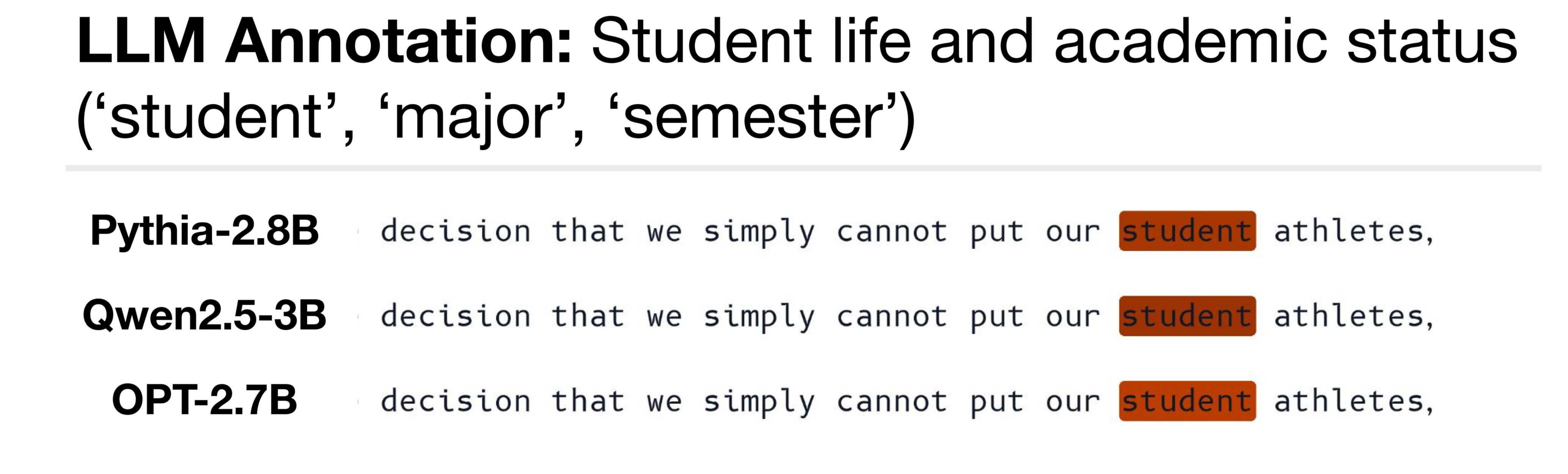}
        \caption{Pythia-2.8B: L12/U5719. Qwen2.5-3B: L26/U2175. OPT-2.7B: L21/U7710.}
        \label{fig:llm_qualitative3_c}
    \end{subfigure}
    \hfill
    \begin{subfigure}[b]{0.49\textwidth}
        \centering
        \includegraphics[width=\textwidth]{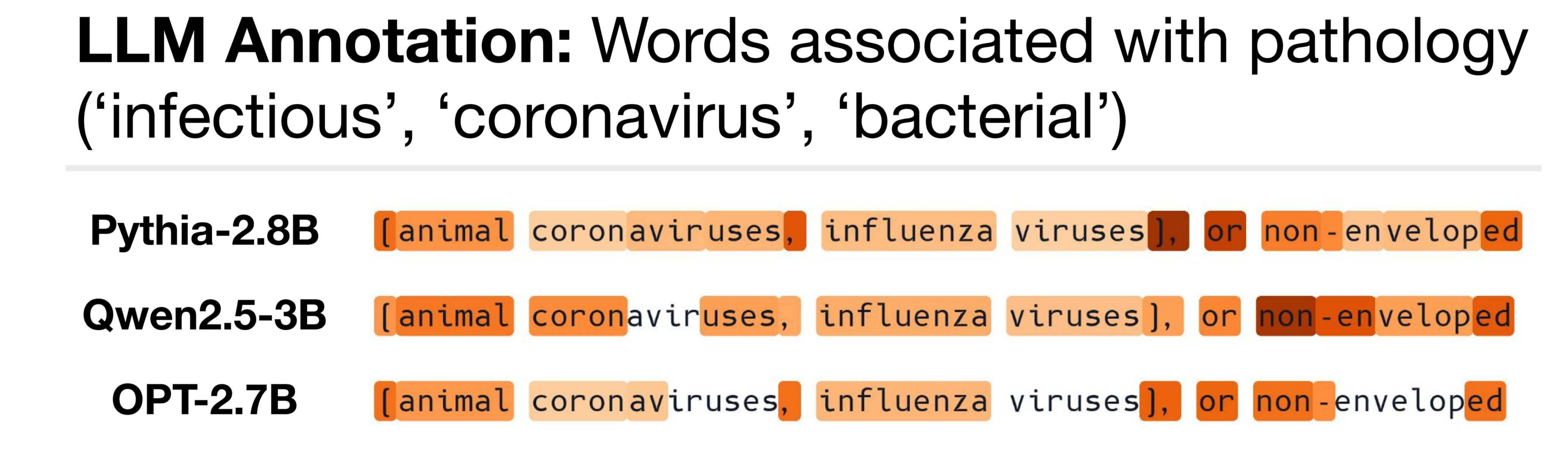}
        \caption{Pythia-2.8B: L28/U10099. Qwen2.5-3B: L31/U6319. OPT-2.7B: L31/U2355.}
        \label{fig:llm_qualitative3_d}
    \end{subfigure}
    
    \caption{\textbf{LLM annotations for Rosetta Neurons.} Comparison between Pythia-2.8B, Qwen2.5-3B, OPT-2.7B.}
    \label{fig:llm_qualitative3}
\end{figure}

\begin{figure}[p]
    \centering
    
    \begin{subfigure}[b]{0.49\textwidth}
        \centering
        \includegraphics[width=\textwidth]{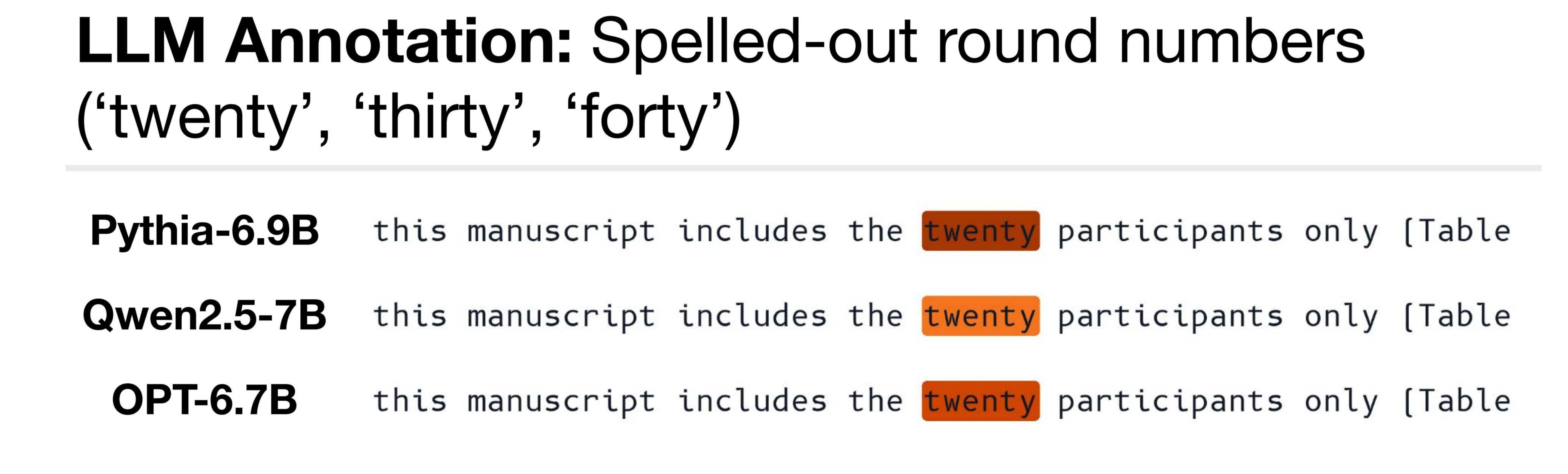}
        \caption{Pythia-6.9B: L1/U1161. Qwen2.5-7B: L22/U12072. OPT-6.7B: L2/U7258.}
        \label{fig:llm_qualitative4_a}
    \end{subfigure}
    \hfill
    \begin{subfigure}[b]{0.49\textwidth}
        \centering
        \includegraphics[width=\textwidth]{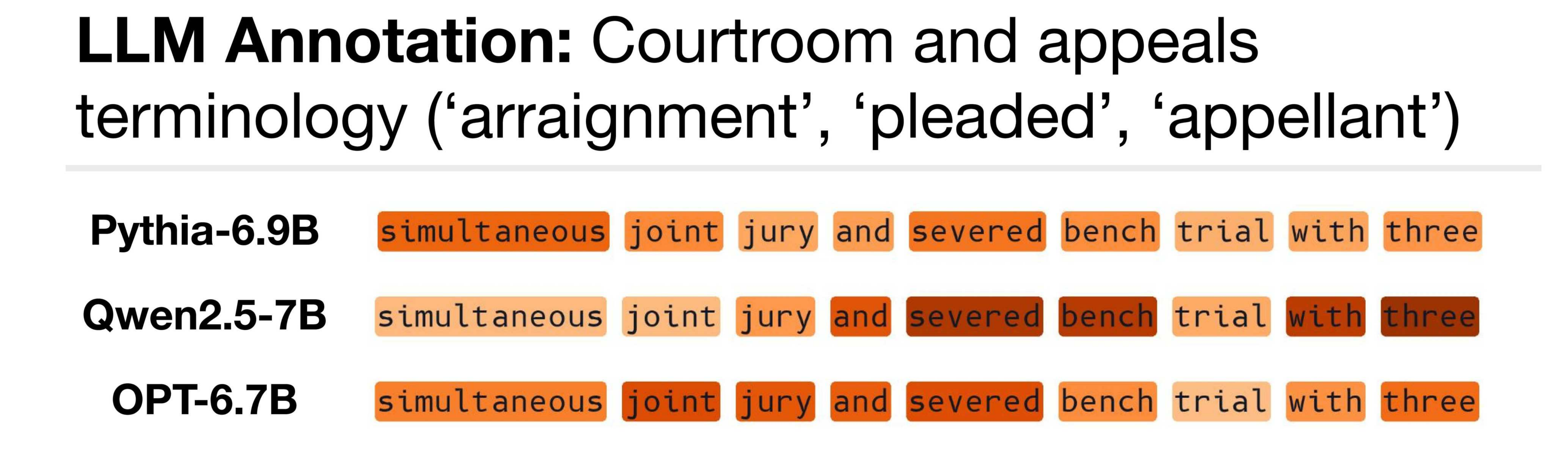}
        \caption{Pythia-6.9B: L14/U15135. Qwen2.5-7B: L23/U5673. OPT-6.7B: L26/U14602.}
        \label{fig:llm_qualitative4_b}
    \end{subfigure}
    
    \vspace{1em}
    
    \begin{subfigure}[b]{0.49\textwidth}
        \centering
        \includegraphics[width=\textwidth]{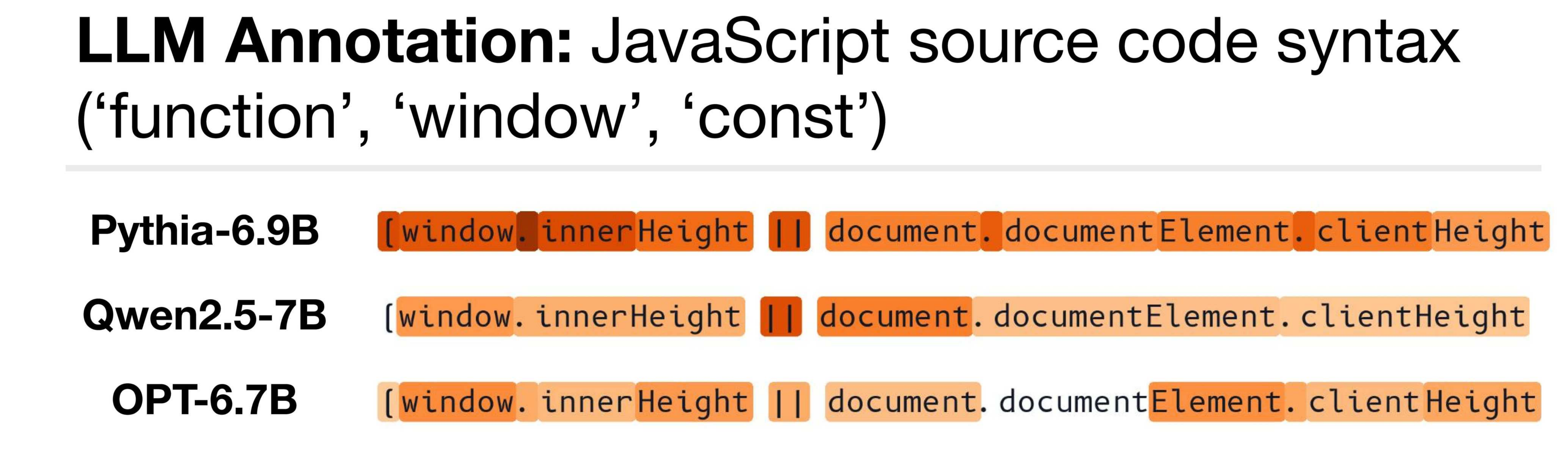}
        \caption{Pythia-6.9B: L16/U11168. Qwen2.5-7B: L22/U14665. OPT-6.7B: L26/U2660.}
        \label{fig:llm_qualitative4_c}
    \end{subfigure}
    \hfill
    \begin{subfigure}[b]{0.49\textwidth}
        \centering
        \includegraphics[width=\textwidth]{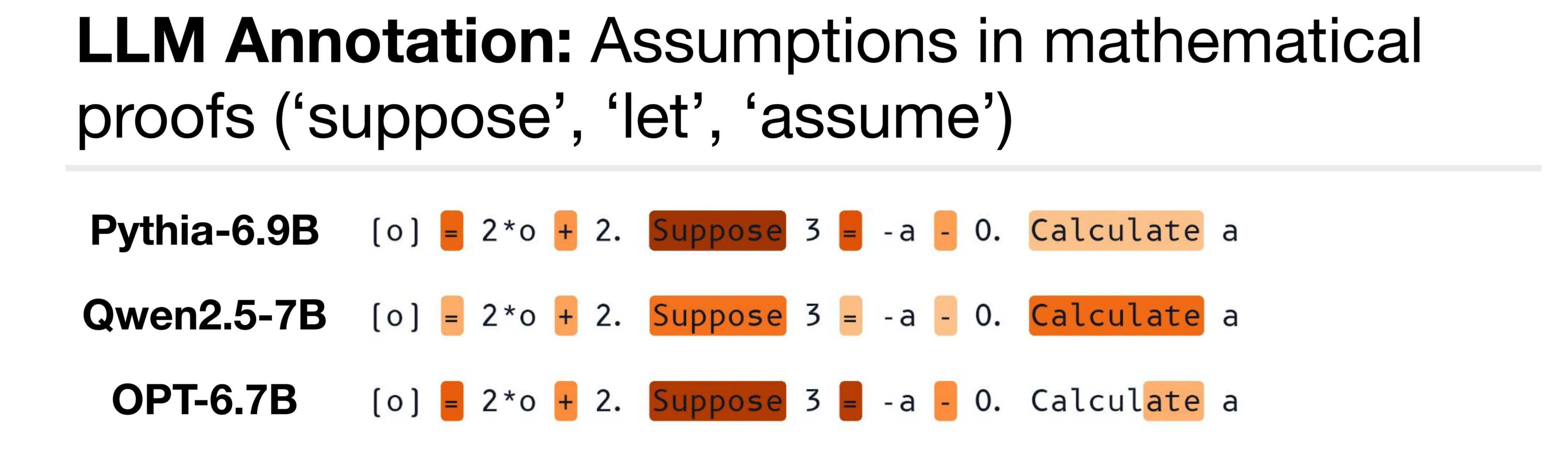}
        \caption{Pythia-6.9B: L27/U826. Qwen2.5-7B: L26/U5796. OPT-6.7B: L29/U14520.}
        \label{fig:llm_qualitative4_d}
    \end{subfigure}
    
    \caption{\textbf{LLM annotations for Rosetta Neurons.} Comparison between Pythia-6.9B, Qwen2.5-7B, OPT-6.7B.}
    \label{fig:llm_qualitative4}
\end{figure}

\begin{figure}[p]
    \centering
    
    \begin{subfigure}[b]{0.49\textwidth}
        \centering
        \includegraphics[width=\textwidth]{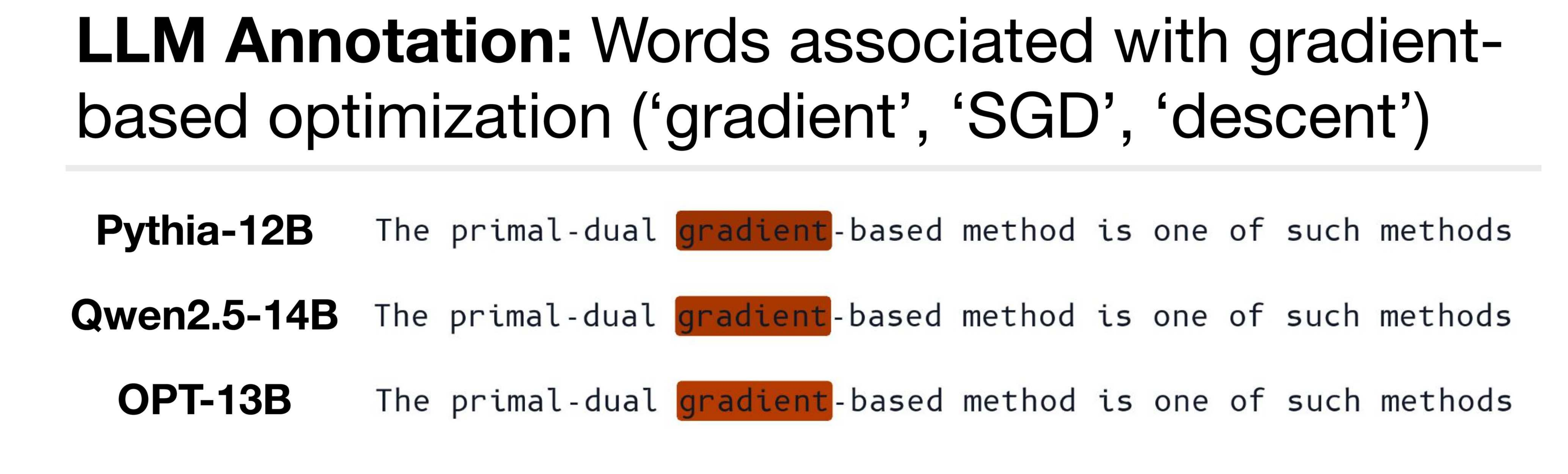}
        \caption{Pythia-12B: L1/U17927. Qwen2.5-14B: L0/U3786. OPT-13B: L0/U4028.}
        \label{fig:llm_qualitative5_a}
    \end{subfigure}
    \hfill
    \begin{subfigure}[b]{0.49\textwidth}
        \centering
        \includegraphics[width=\textwidth]{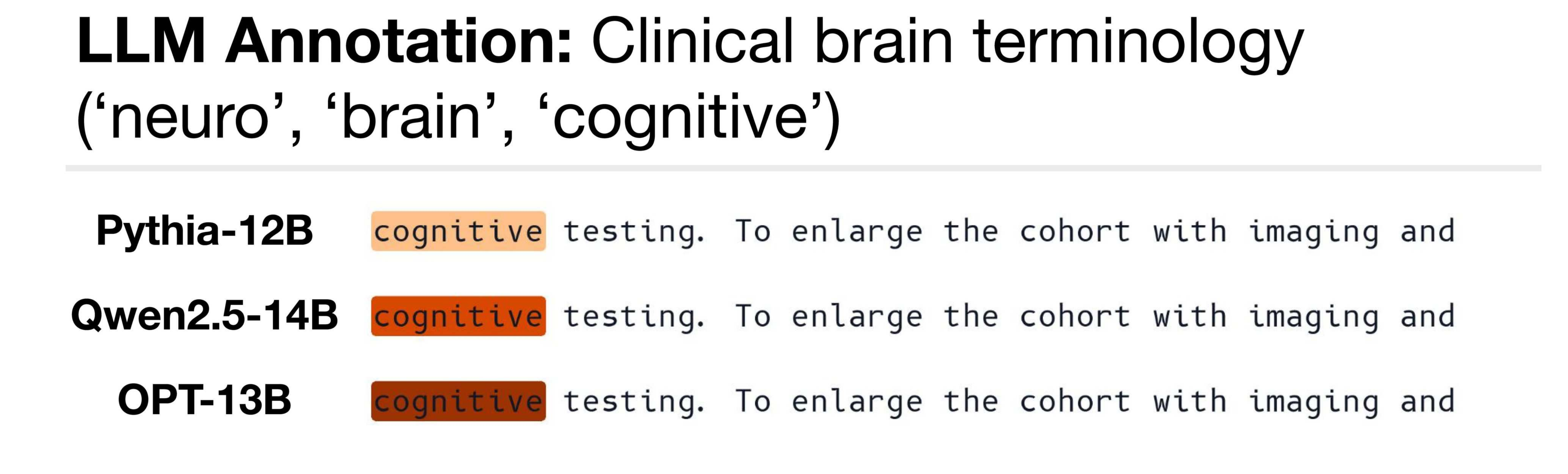}
        \caption{Pythia-12B: L1/U19682. Qwen2.5-14B: L3/U1521. OPT-13B: L0/U2590.}
        \label{fig:llm_qualitative5_b}
    \end{subfigure}
    
    \vspace{1em}
    
    \begin{subfigure}[b]{0.49\textwidth}
        \centering
        \includegraphics[width=\textwidth]{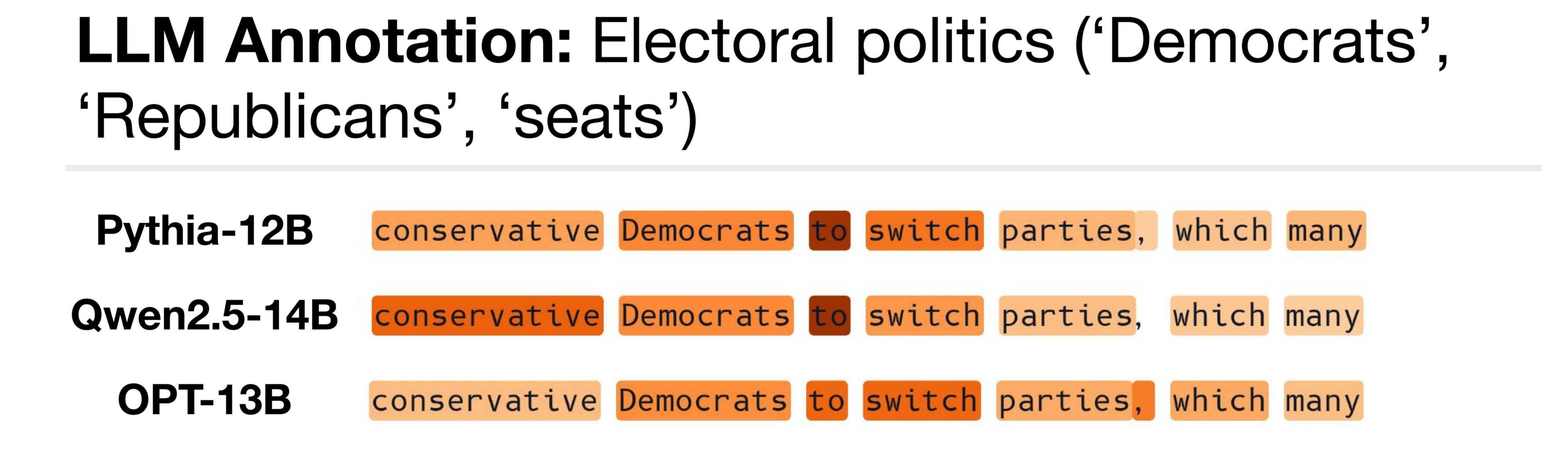}
        \caption{Pythia-12B: L24/U10513. Qwen2.5-14B: L33/U11681. OPT-13B: L29/U14.}
        \label{fig:llm_qualitative5_c}
    \end{subfigure}
    \hfill
    \begin{subfigure}[b]{0.49\textwidth}
        \centering
        \includegraphics[width=\textwidth]{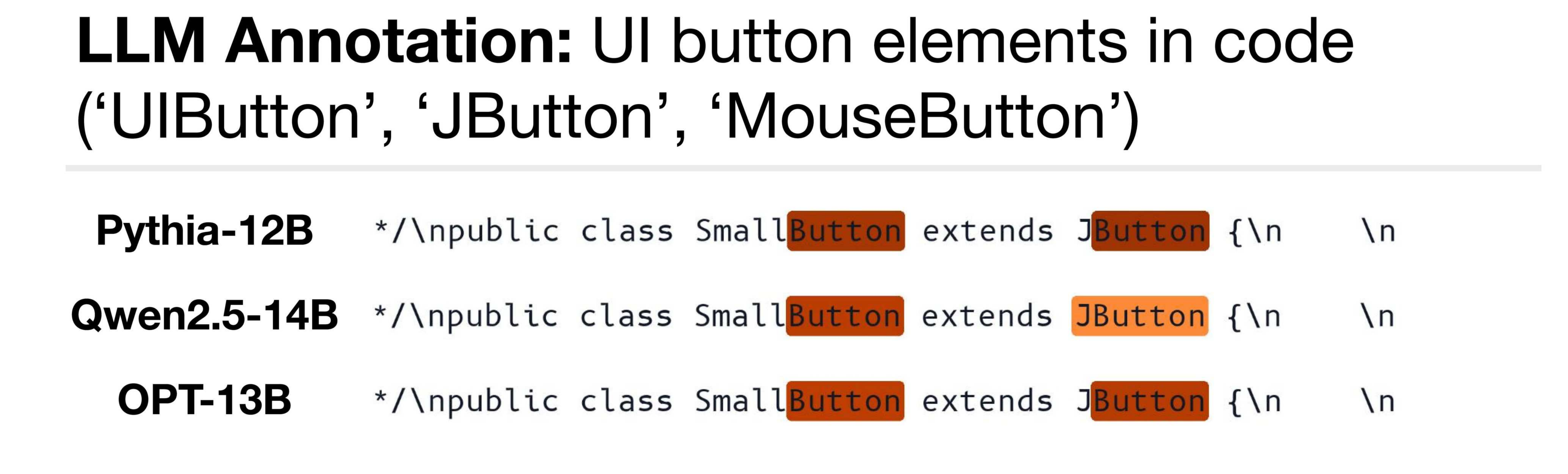}
        \caption{Pythia-12B: L1/U4031. Qwen2.5-14B: L0/U6188. OPT-13B: L3/U11416.}
        \label{fig:llm_qualitative5_d}
    \end{subfigure}
    
    \caption{\textbf{LLM annotations for Rosetta Neurons.} Comparison between Pythia-12B, Qwen2.5-14B, OPT-13B.}
    \label{fig:llm_qualitative5}
\end{figure}

\begin{figure}[p]
    \centering
    
    \begin{subfigure}[b]{0.4\textwidth}
        \centering
        \includegraphics[width=\textwidth]{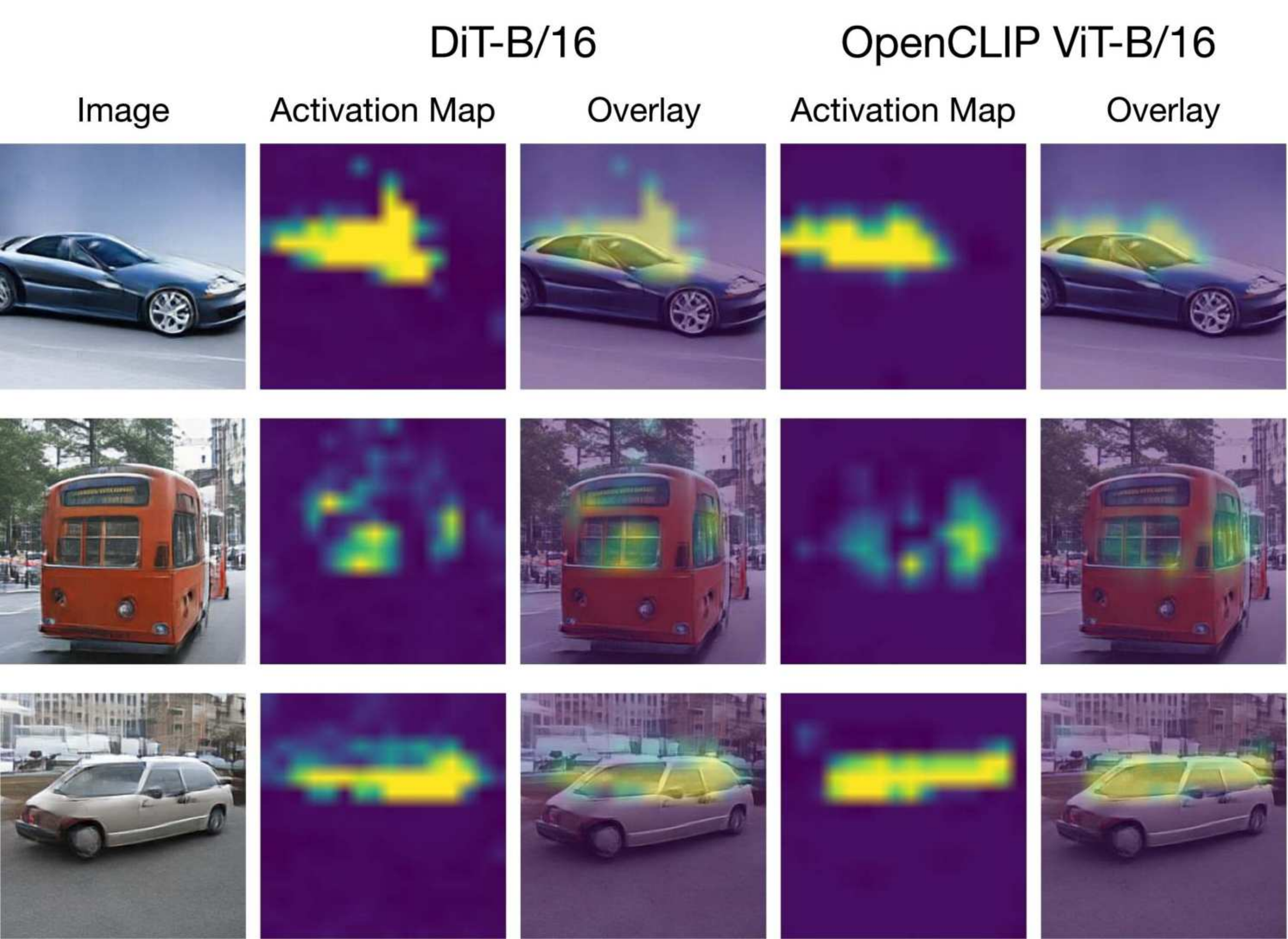}
        \caption{DiT-B/16: L7/U879. OpenCLIP ViT-B/16: L7/U1879.}
        \label{fig:sub_a_B16}
    \end{subfigure}
    \hspace{1em}
    \begin{subfigure}[b]{0.4\textwidth}
        \centering
        \includegraphics[width=\textwidth]{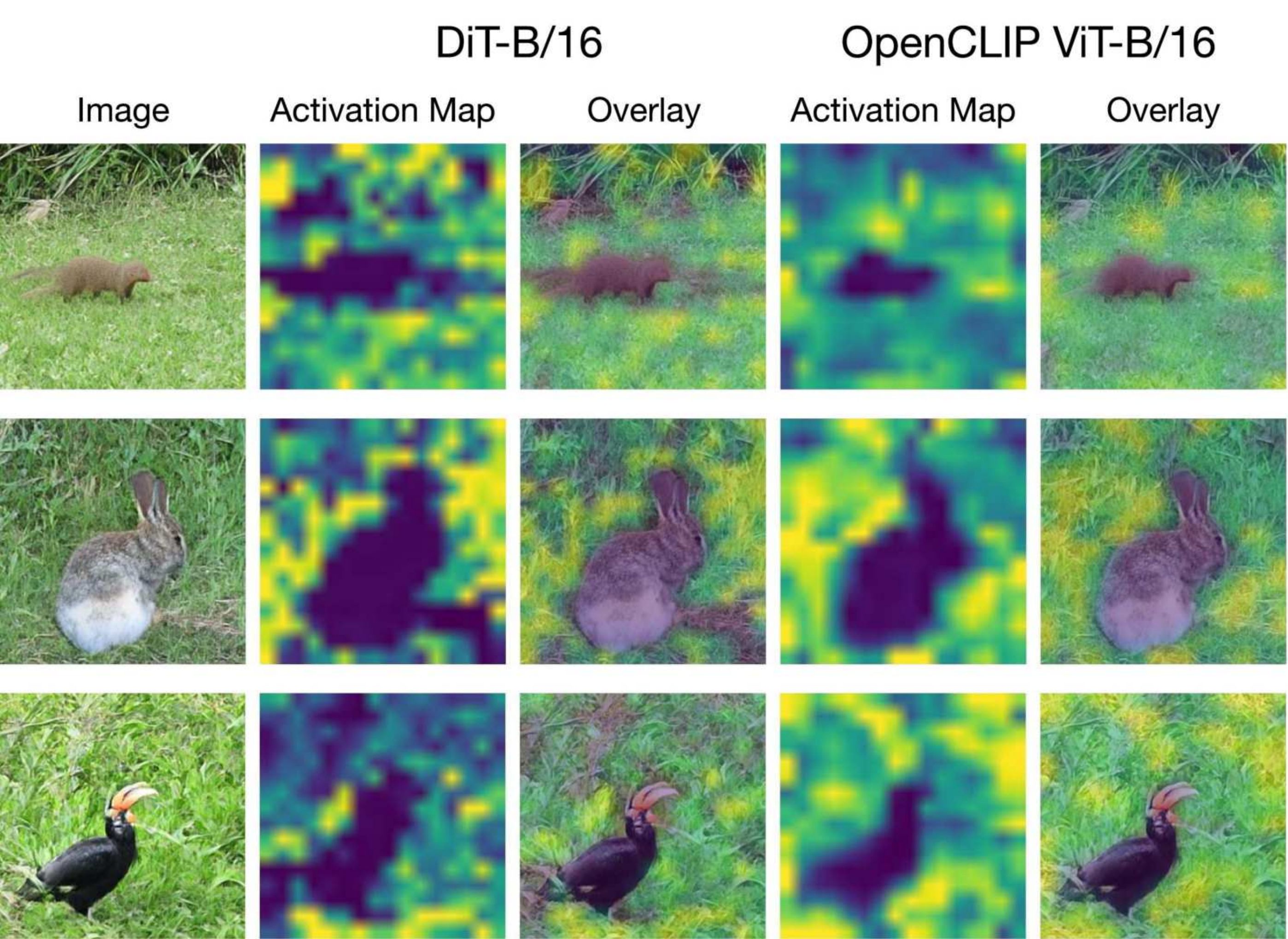}
        \caption{DiT-B/16: L11/U610. OpenCLIP ViT-B/16: L4/U415.}
        \label{fig:sub_b_B16}
    \end{subfigure}
    
    \vspace{0.5em}
    
    \begin{subfigure}[b]{0.4\textwidth}
        \centering
        \includegraphics[width=\textwidth]{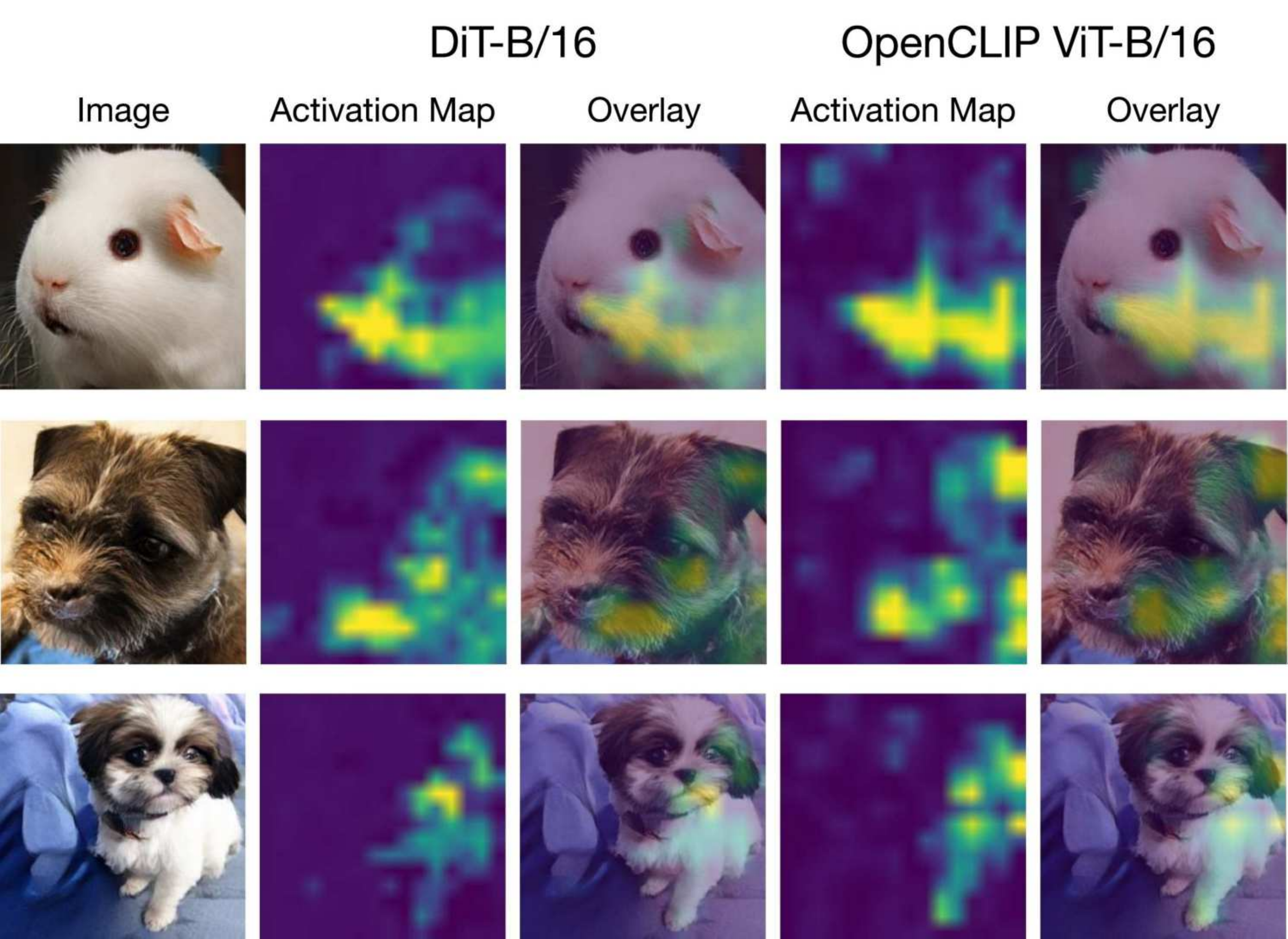}
        \caption{DiT-B/16: L11/U1641. OpenCLIP ViT-B/16: L4/U730.}
        \label{fig:sub_c_B16}
    \end{subfigure}
    \hspace{1em}
    \begin{subfigure}[b]{0.4\textwidth}
        \centering
        \includegraphics[width=\textwidth]{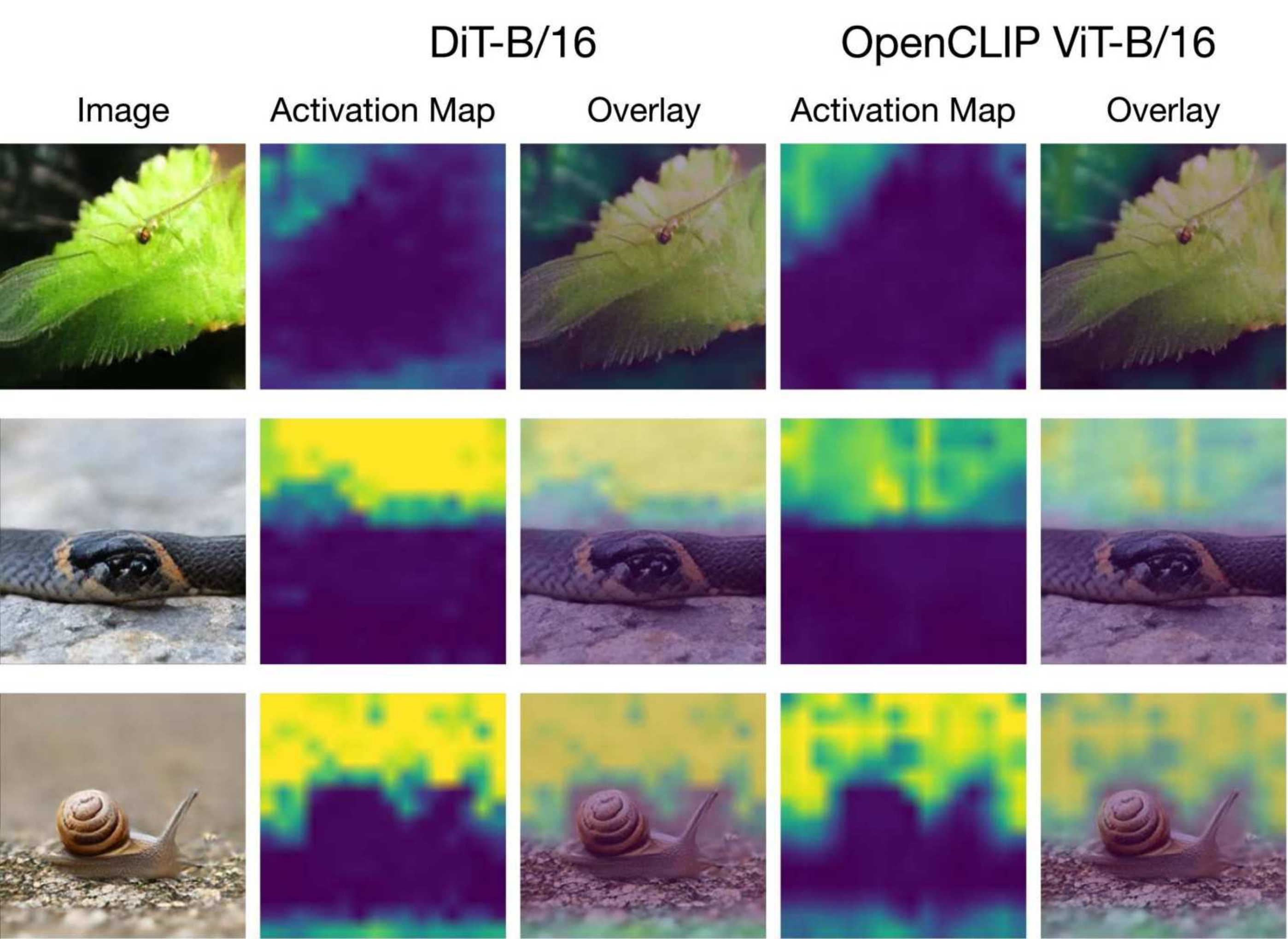}
        \caption{DiT-B/16: L10/U1516. OpenCLIP ViT-B/16: L4/U689.}
        \label{fig:sub_d_B16}
    \end{subfigure}
    
    \caption{\textbf{Top-activating images for Rosetta Neurons.} Comparison between DiT-B/16 and OpenCLIP ViT-B/16.}
    \label{fig:b_16}

    \vspace{2em} 

    \begin{subfigure}[b]{0.4\textwidth}
        \centering
        \includegraphics[width=\textwidth]{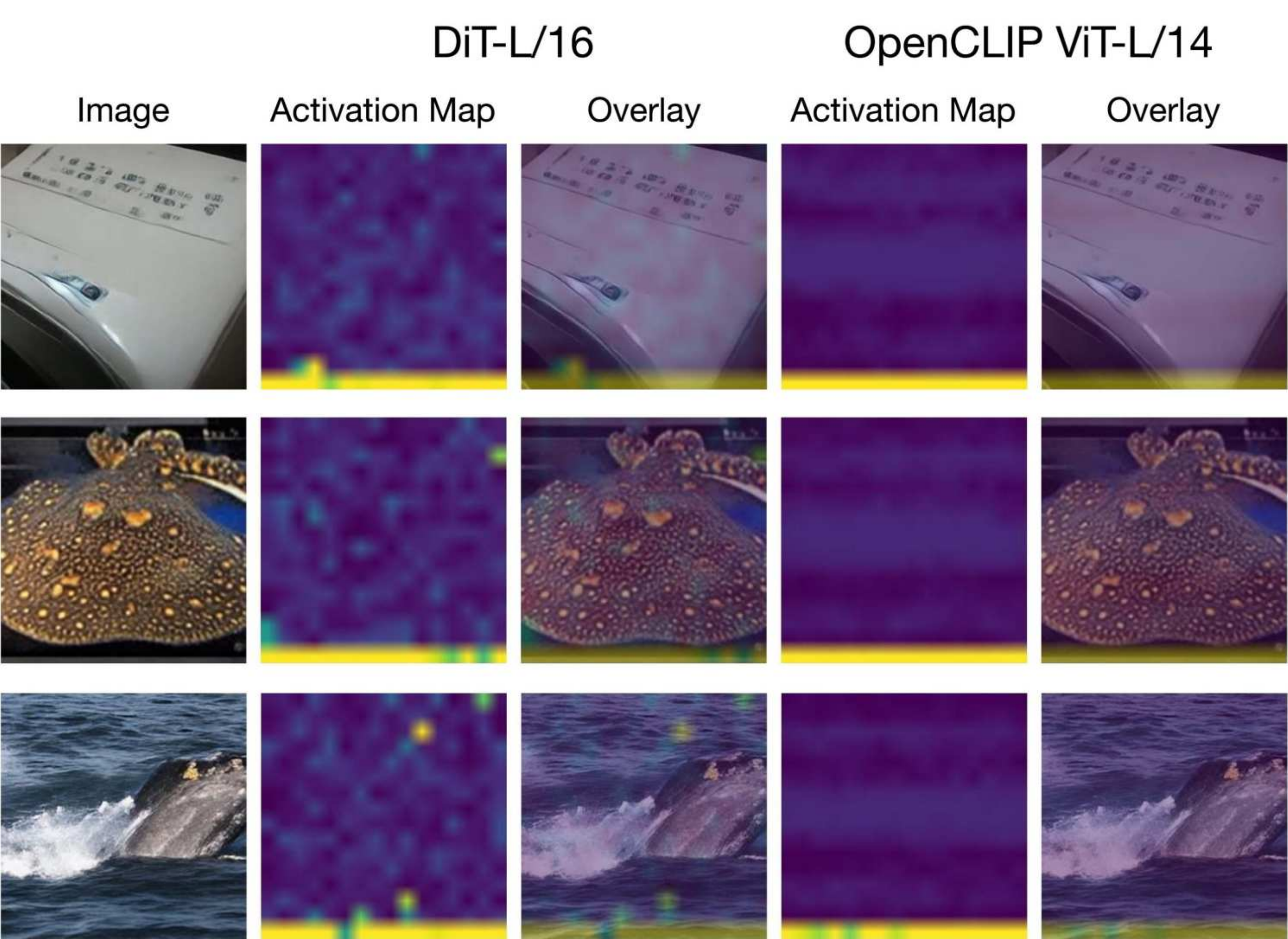}
        \caption{DiT-L/16: L0/U1228. OpenCLIP ViT-L/14: L1/U1669.}
        \label{fig:sub_a_L16}
    \end{subfigure}
    \hspace{1em}
    \begin{subfigure}[b]{0.4\textwidth}
        \centering
        \includegraphics[width=\textwidth]{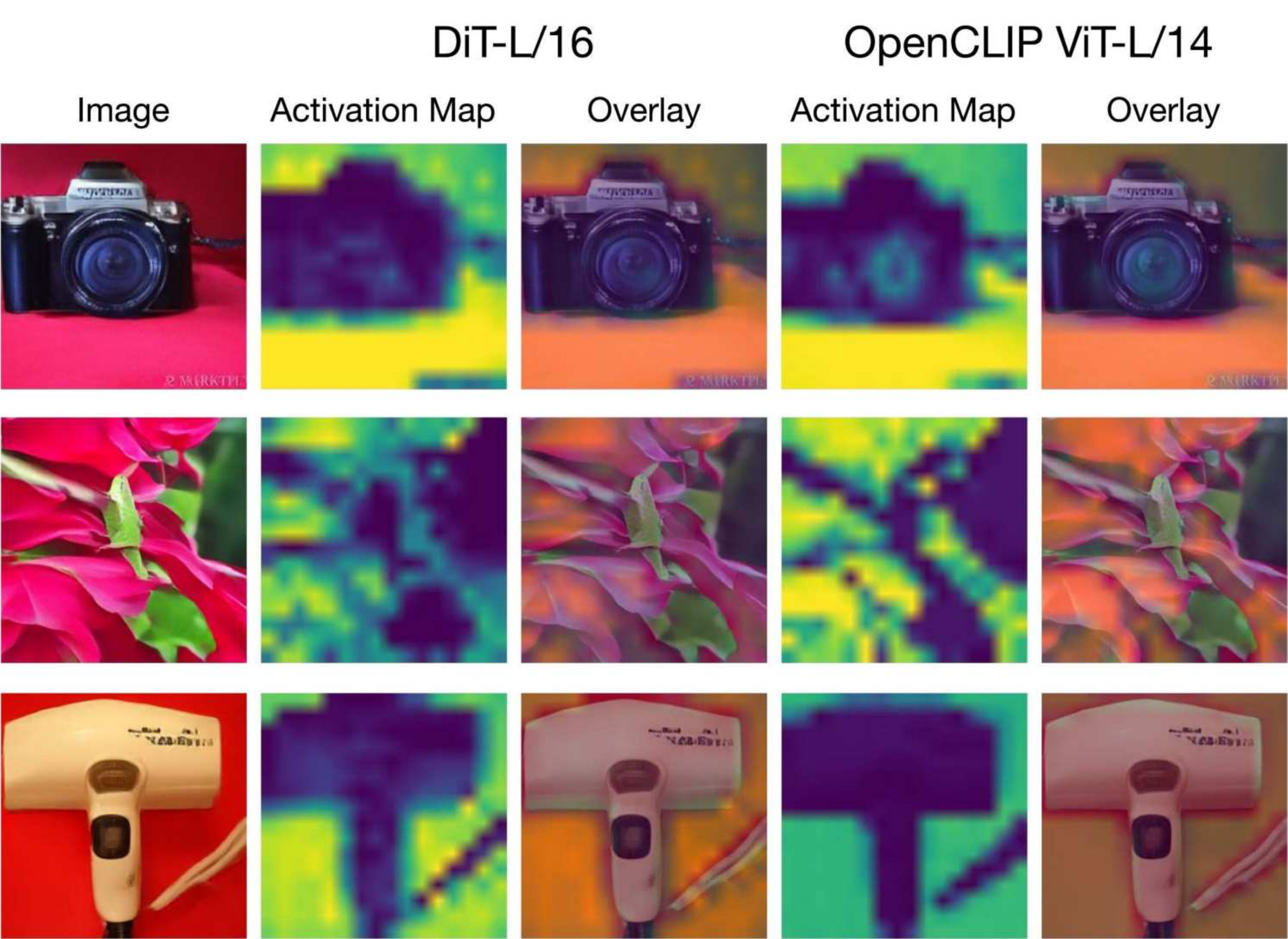}
        \caption{DiT-L/16: L31/U15. OpenCLIP ViT-L/14: L0/U1207.}
        \label{fig:sub_b_L16}
    \end{subfigure}
    
    \vspace{0.5em}
    
    \begin{subfigure}[b]{0.4\textwidth}
        \centering
        \includegraphics[width=\textwidth]{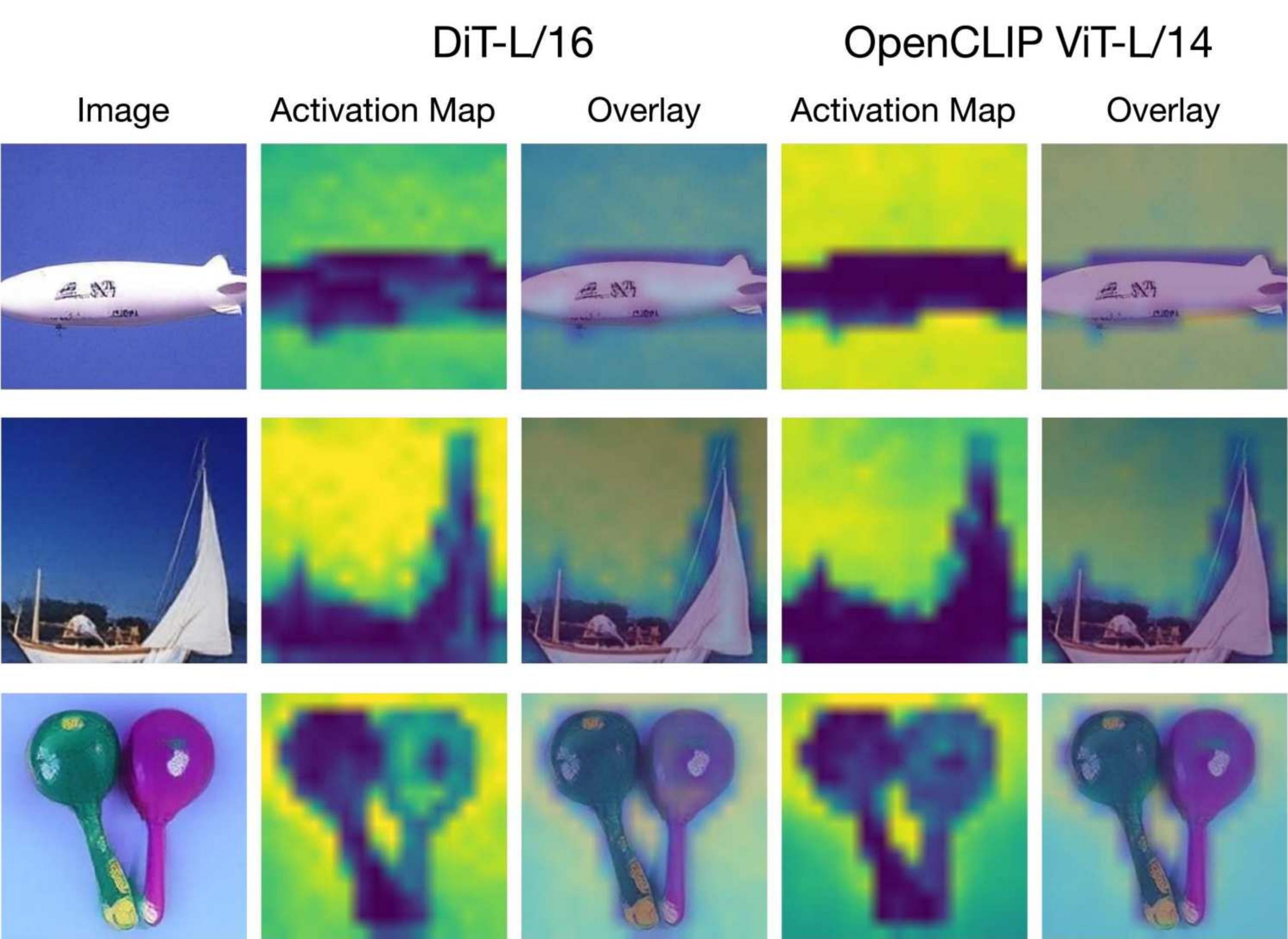}
        \caption{DiT-L/16: L31/U1752. OpenCLIP ViT-L/14: L0/U3894.}
        \label{fig:sub_c_L16}
    \end{subfigure}
    \hspace{1em}
    \begin{subfigure}[b]{0.4\textwidth}
        \centering
        \includegraphics[width=\textwidth]{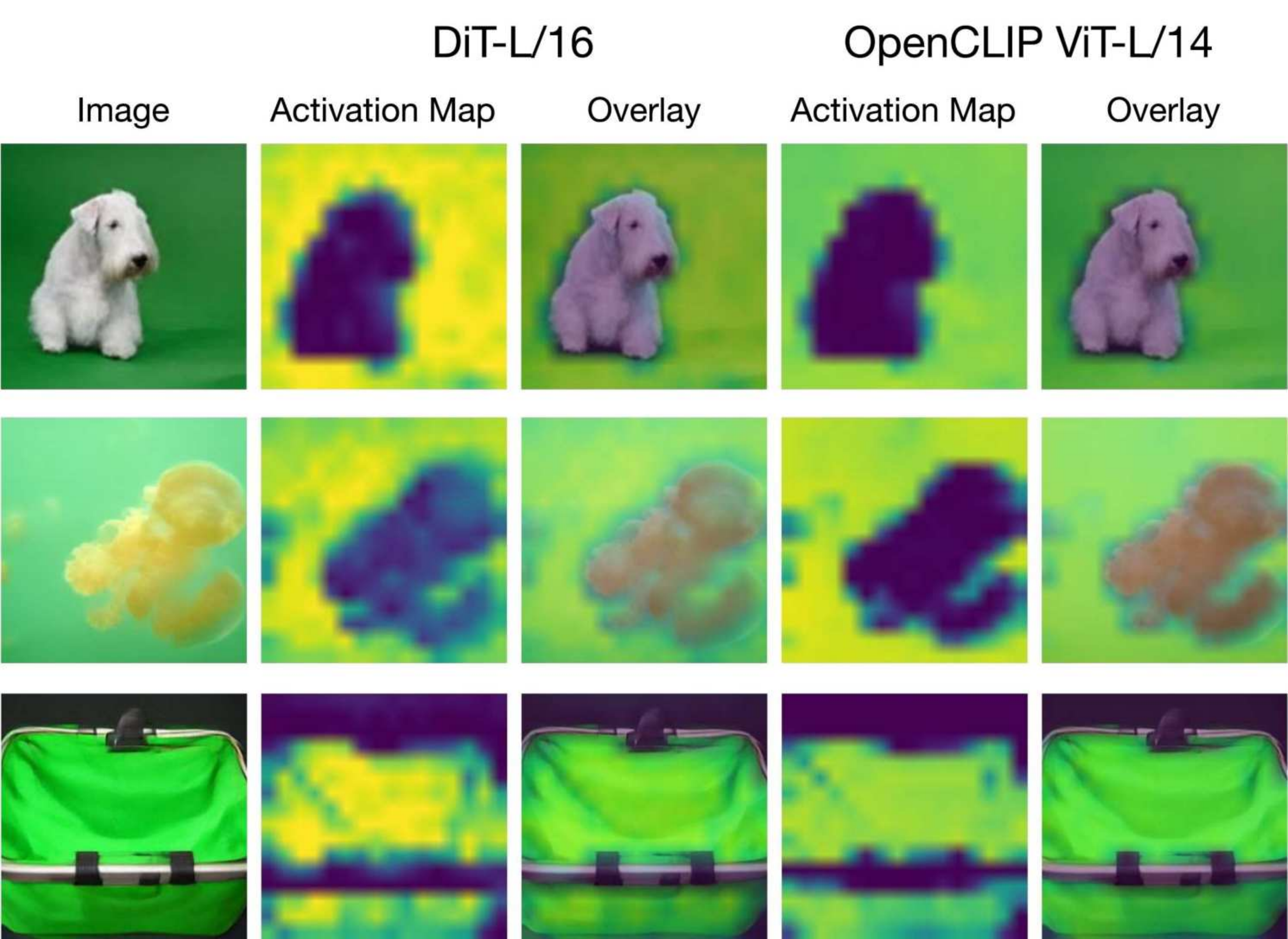}
        \caption{DiT-L/16: L31/U816. OpenCLIP ViT-L/14: L0/U2394.}
        \label{fig:sub_d_L16}
    \end{subfigure}
    
    \caption{\textbf{Top-activating images for Rosetta Neurons.} Comparison between DiT-L/16 and OpenCLIP ViT-L/14.}
    \label{fig:L_16}
\end{figure}

\begin{figure}[p]
    \centering
    
    \begin{subfigure}[b]{0.4\textwidth}
        \centering
        \includegraphics[width=\textwidth]{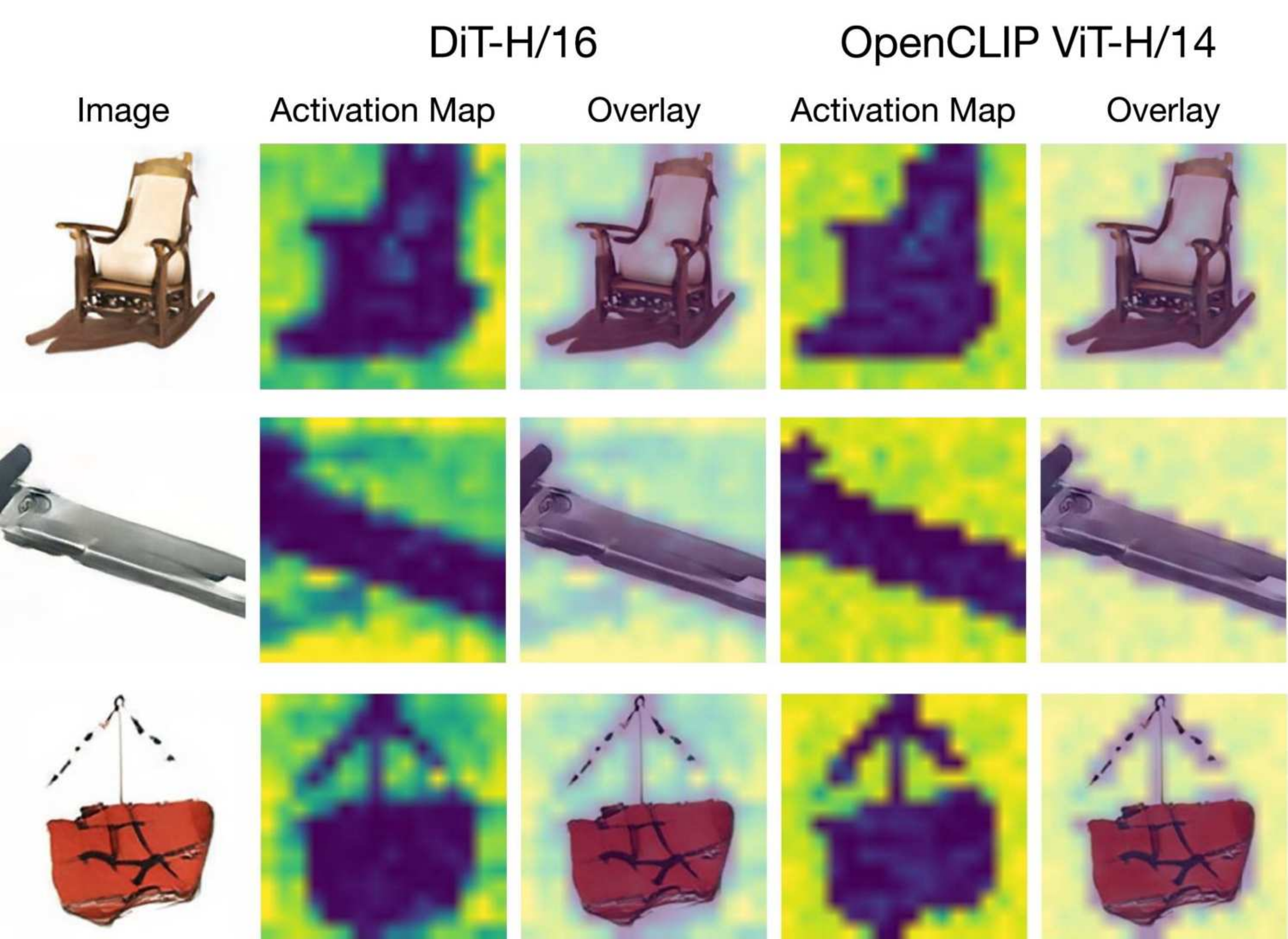}
        \caption{DiT-H/16: L45/U1346. OpenCLIP ViT-H/14: L1/U3383.}
        \label{fig:sub_a_H16}
    \end{subfigure}
    \hspace{1em}
    \begin{subfigure}[b]{0.4\textwidth}
        \centering
        \includegraphics[width=\textwidth]{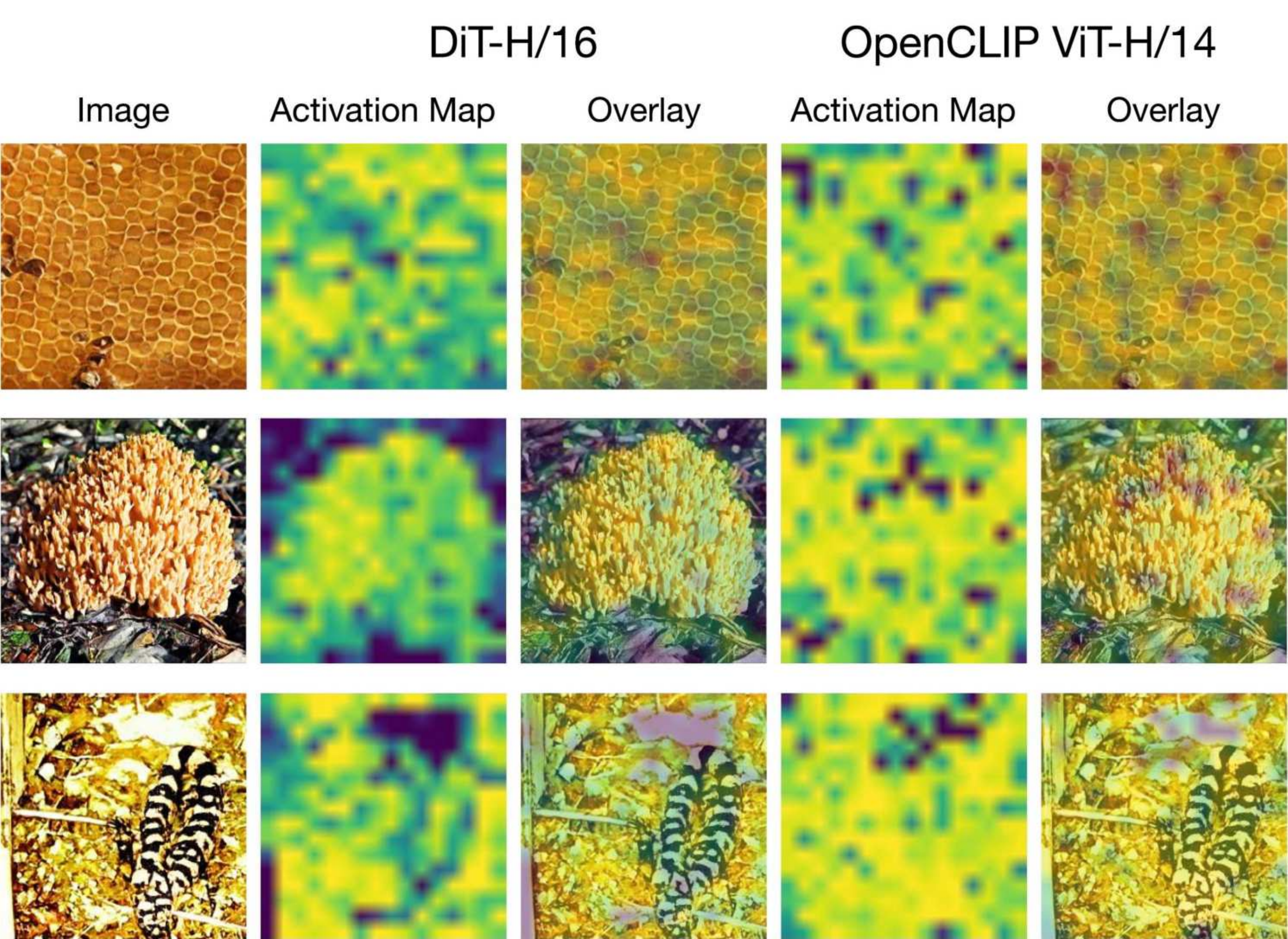}
        \caption{DiT-H/16: L47/U1097. OpenCLIP ViT-H/14: L1/U4324.}
        \label{fig:sub_b_H16}
    \end{subfigure}
    
    \vspace{0.5em}
    
    \begin{subfigure}[b]{0.4\textwidth}
        \centering
        \includegraphics[width=\textwidth]{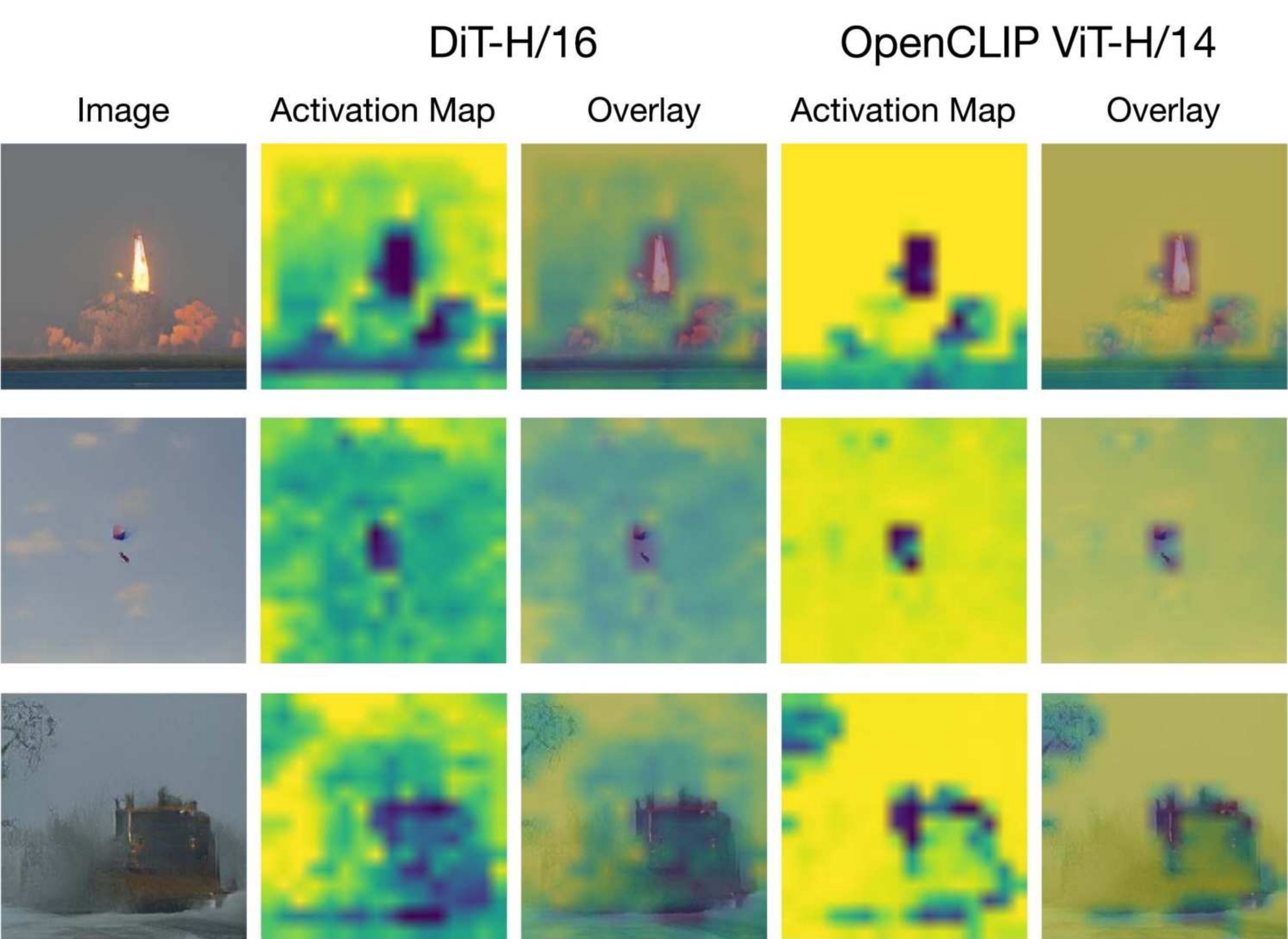}
        \caption{DiT-H/16: L47/U1230. OpenCLIP ViT-H/14: L0/U3495.}
        \label{fig:sub_c_H16}
    \end{subfigure}
    \hspace{1em}
    \begin{subfigure}[b]{0.4\textwidth}
        \centering
        \includegraphics[width=\textwidth]{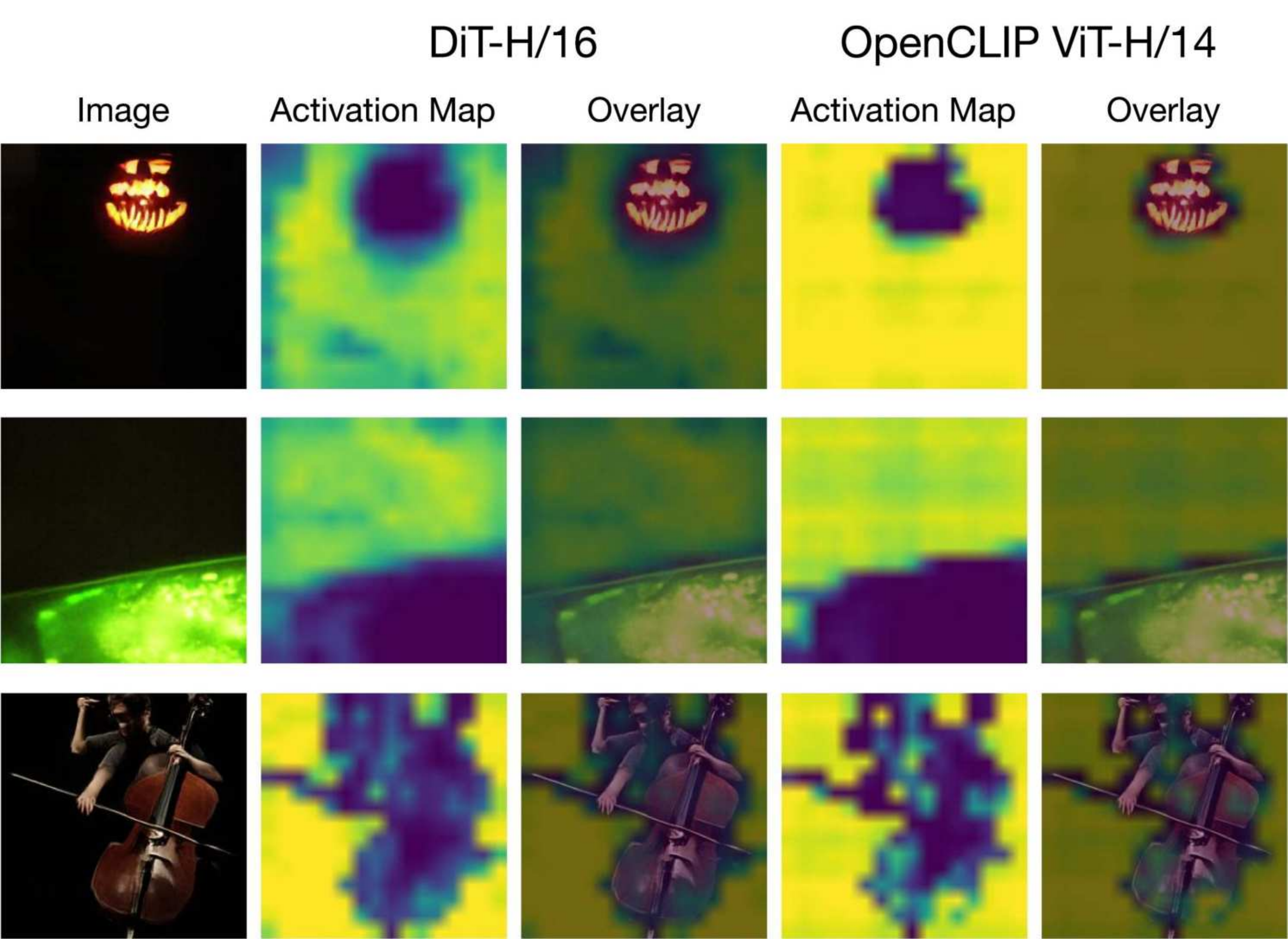}
        \caption{DiT-H/16: L47/U300. OpenCLIP ViT-H/14: L0/U477.}
        \label{fig:sub_d_H16}
    \end{subfigure}
    
    \caption{\textbf{Top-activating images for Rosetta Neurons.} Comparison between DiT-H/16 and OpenCLIP ViT-H/14.}
    \label{fig:H_16}

    \vspace{2em} 

    \begin{subfigure}[b]{0.4\textwidth}
        \centering
        \includegraphics[width=\textwidth]{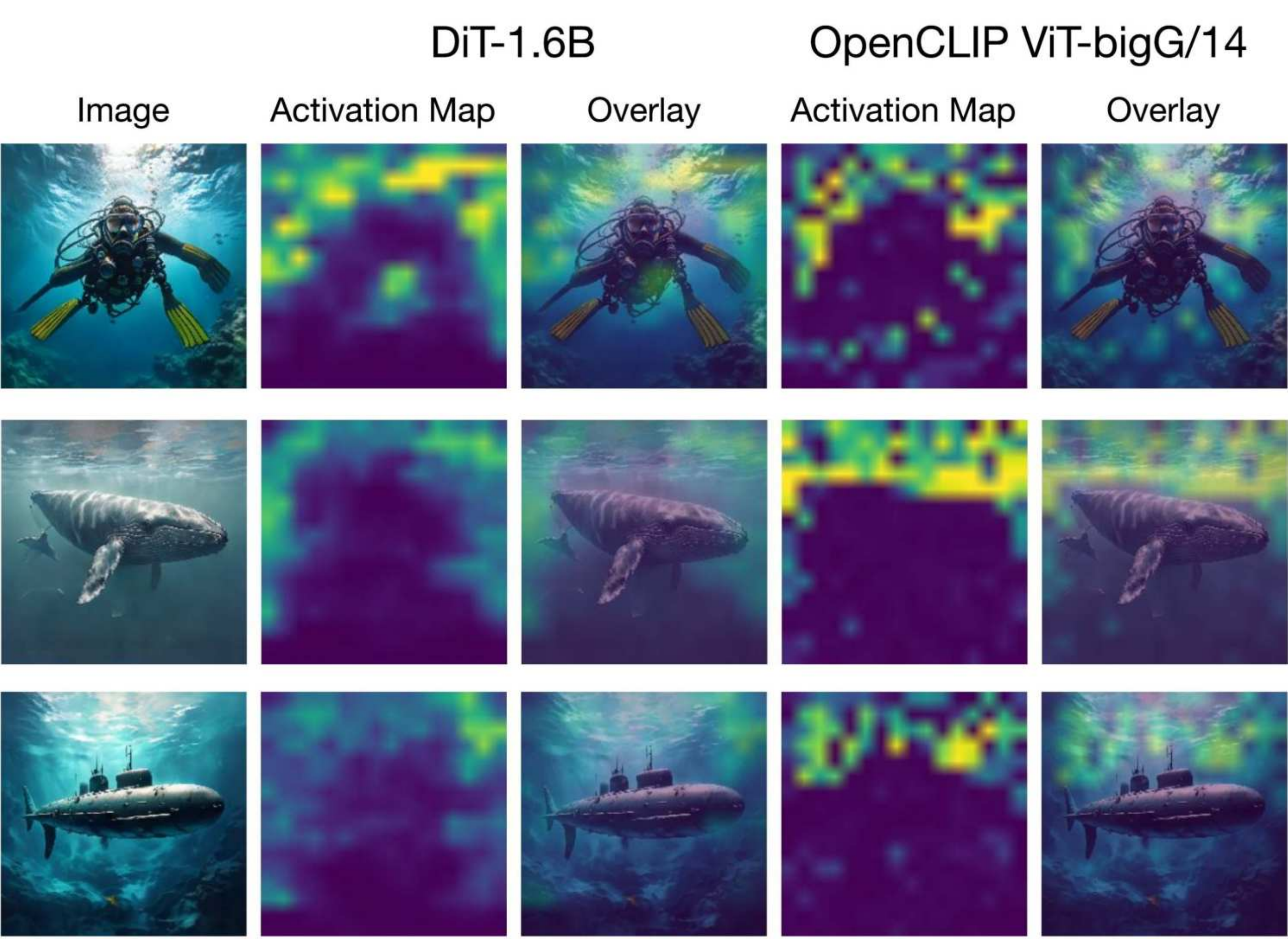}
        \caption{DiT-1.6B: L44/U2447. OpenCLIP ViT-bigG/14: L29/U2762.}
        \label{fig:sub_a_sana}
    \end{subfigure}
    \hspace{1em}
    \begin{subfigure}[b]{0.4\textwidth}
        \centering
        \includegraphics[width=\textwidth]{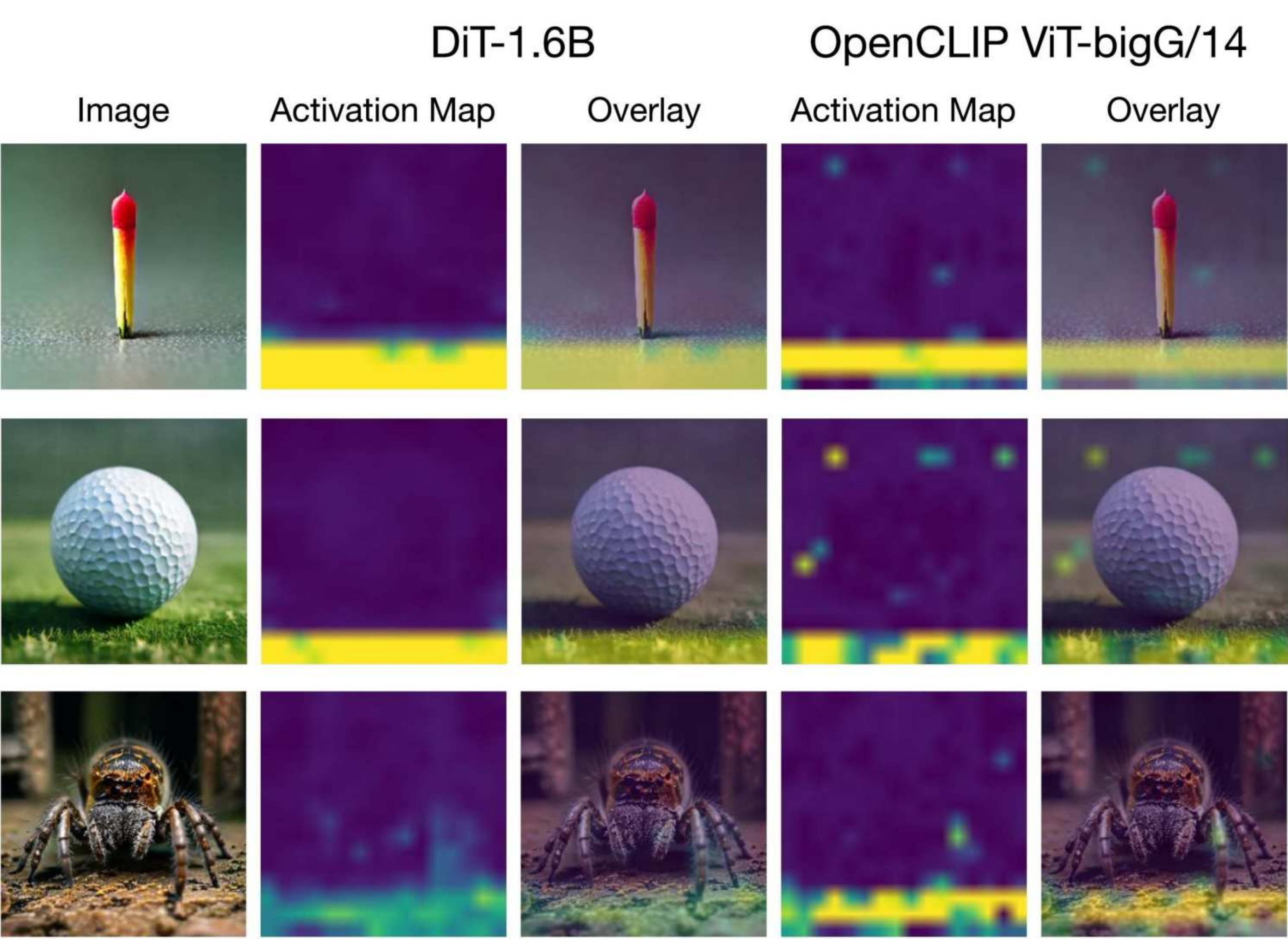}
        \caption{DiT-1.6B: L12/U2668. OpenCLIP ViT-bigG/14: L20/U203.}
        \label{fig:sub_b_sana}
    \end{subfigure}
    
    \vspace{0.5em}
    
    \begin{subfigure}[b]{0.4\textwidth}
        \centering
        \includegraphics[width=\textwidth]{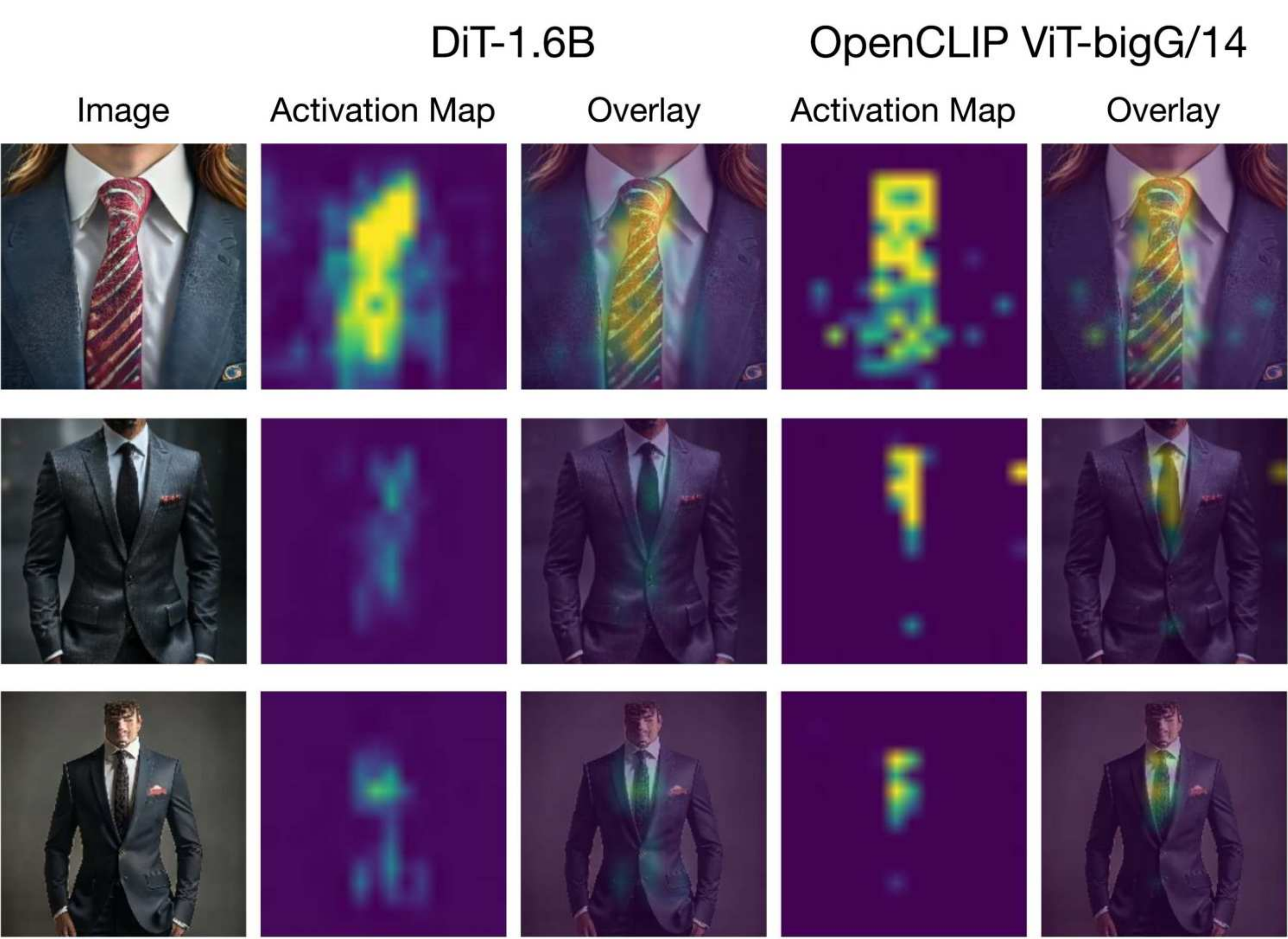}
        \caption{DiT-1.6B: L6/U3382. OpenCLIP ViT-bigG/14: L39/U7045.}
        \label{fig:sub_c_sana}
    \end{subfigure}
    \hspace{1em}
    \begin{subfigure}[b]{0.4\textwidth}
        \centering
        \includegraphics[width=\textwidth]{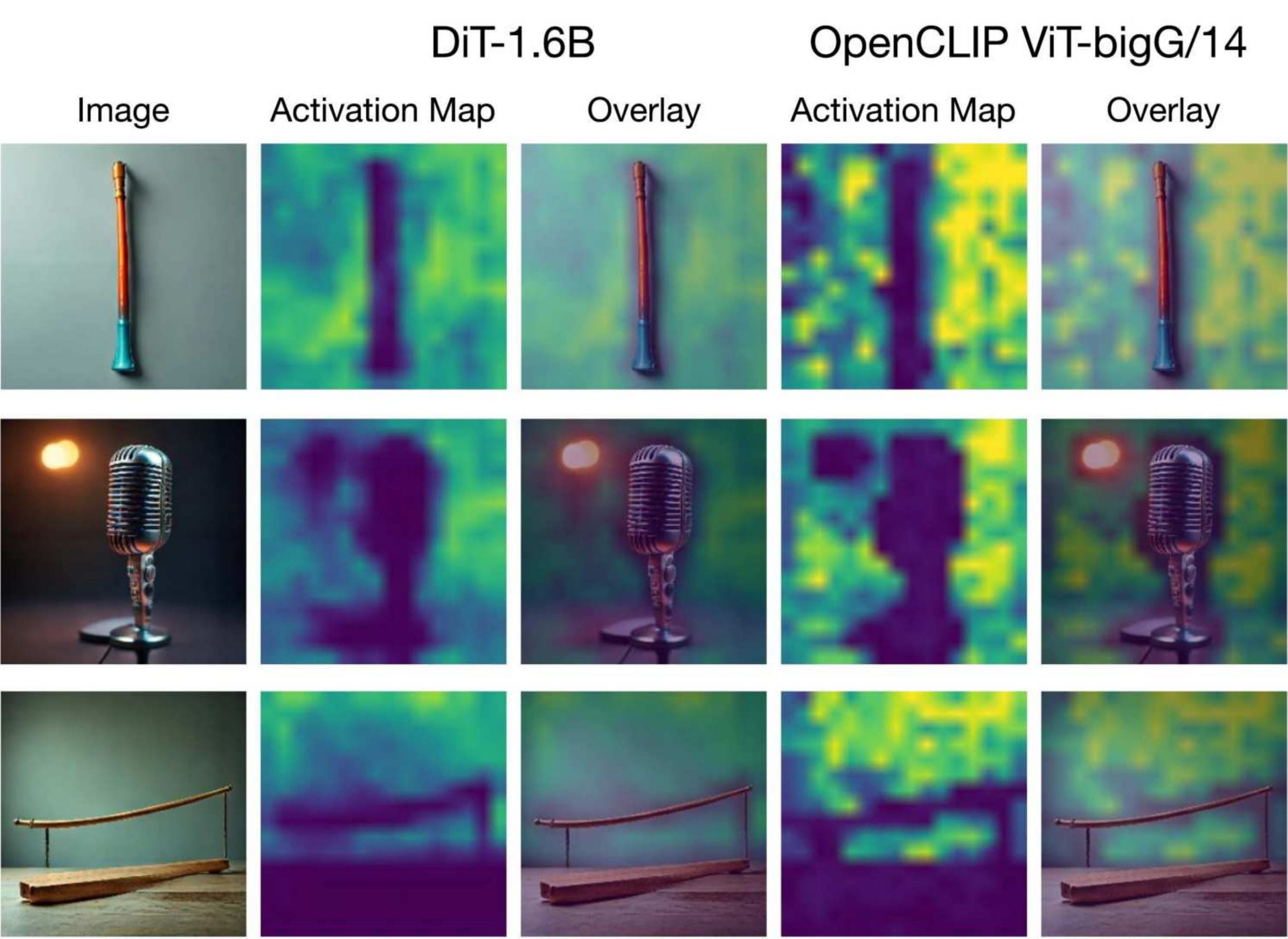}
        \caption{DiT-1.6B: L9/U2940. OpenCLIP ViT-bigG/14: L28/U2265.}
        \label{fig:sub_d_sana}
    \end{subfigure}
    
    \caption{\textbf{Top-activating images for Rosetta Neurons.} Comparison between DiT-1.6B and OpenCLIP ViT-bigG/14.}
    \label{fig:sana}
\end{figure}

\begin{figure}[H]
    \centering
    
    \begin{subfigure}[b]{0.4\textwidth}
        \centering
        \includegraphics[width=\textwidth]{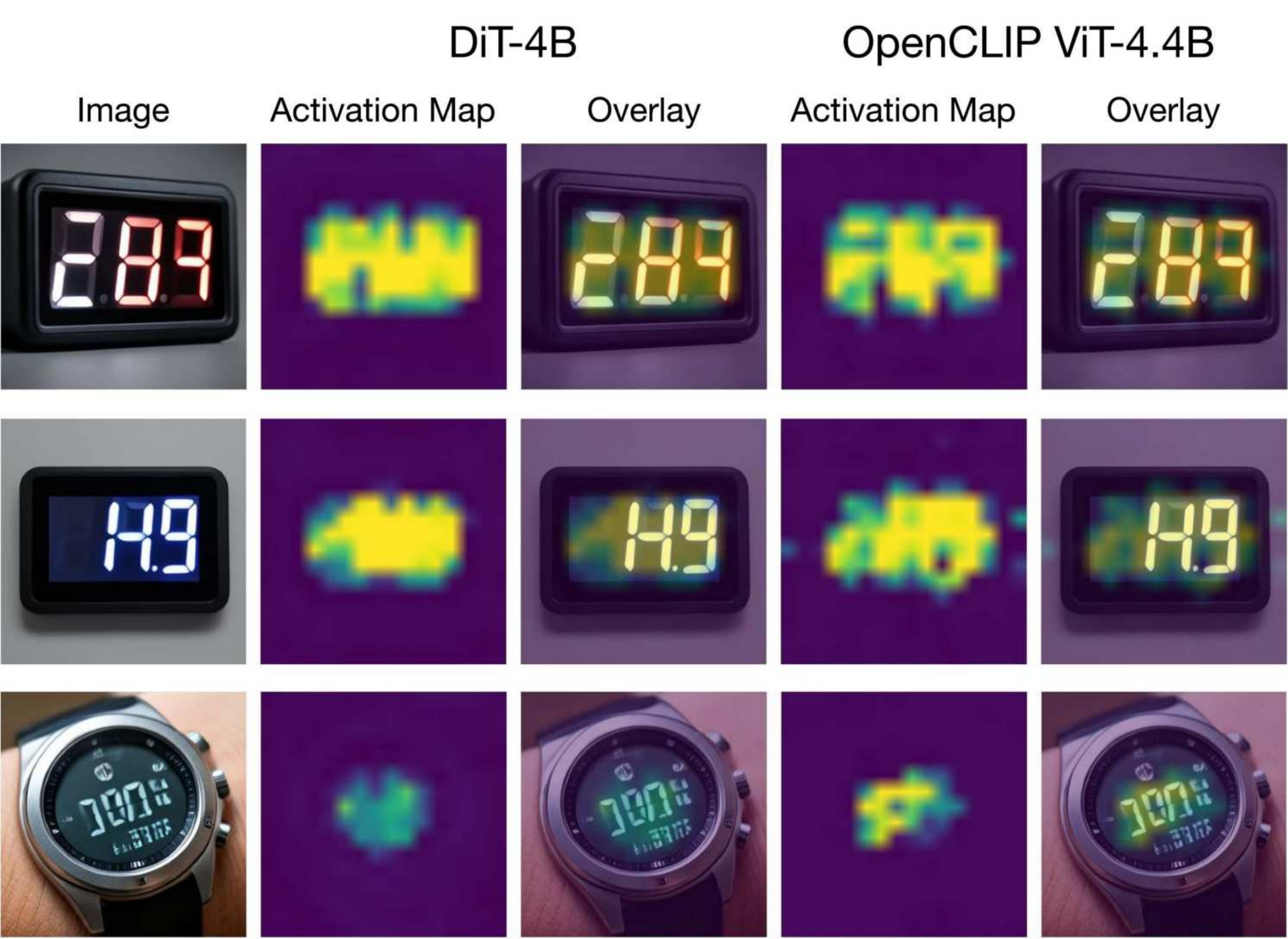}
        \caption{DiT-4B: L8/U4935. OpenCLIP ViT-4.4B: L41/U7020.}
        \label{fig:sub_a_flux}
    \end{subfigure}\hspace{1em} 
    \begin{subfigure}[b]{0.4\textwidth}
        \centering
        \includegraphics[width=\textwidth]{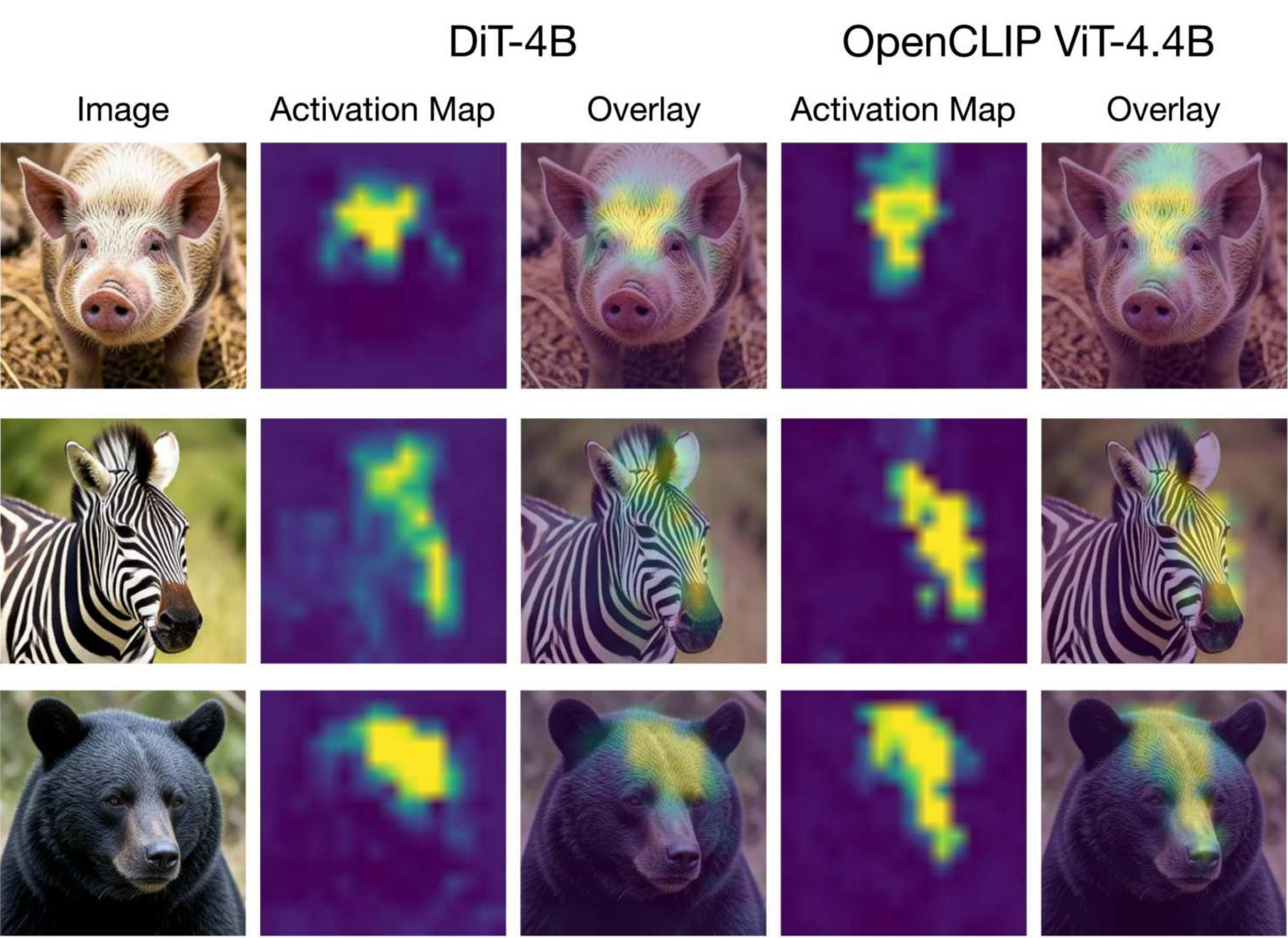}
        \caption{DiT-4B: L2/U4833. OpenCLIP ViT-4.4B: L36/U14930.}
        \label{fig:sub_b_flux}
    \end{subfigure}
    
    \vspace{1em} 
    
    \begin{subfigure}[b]{0.4\textwidth}
        \centering
        \includegraphics[width=\textwidth]{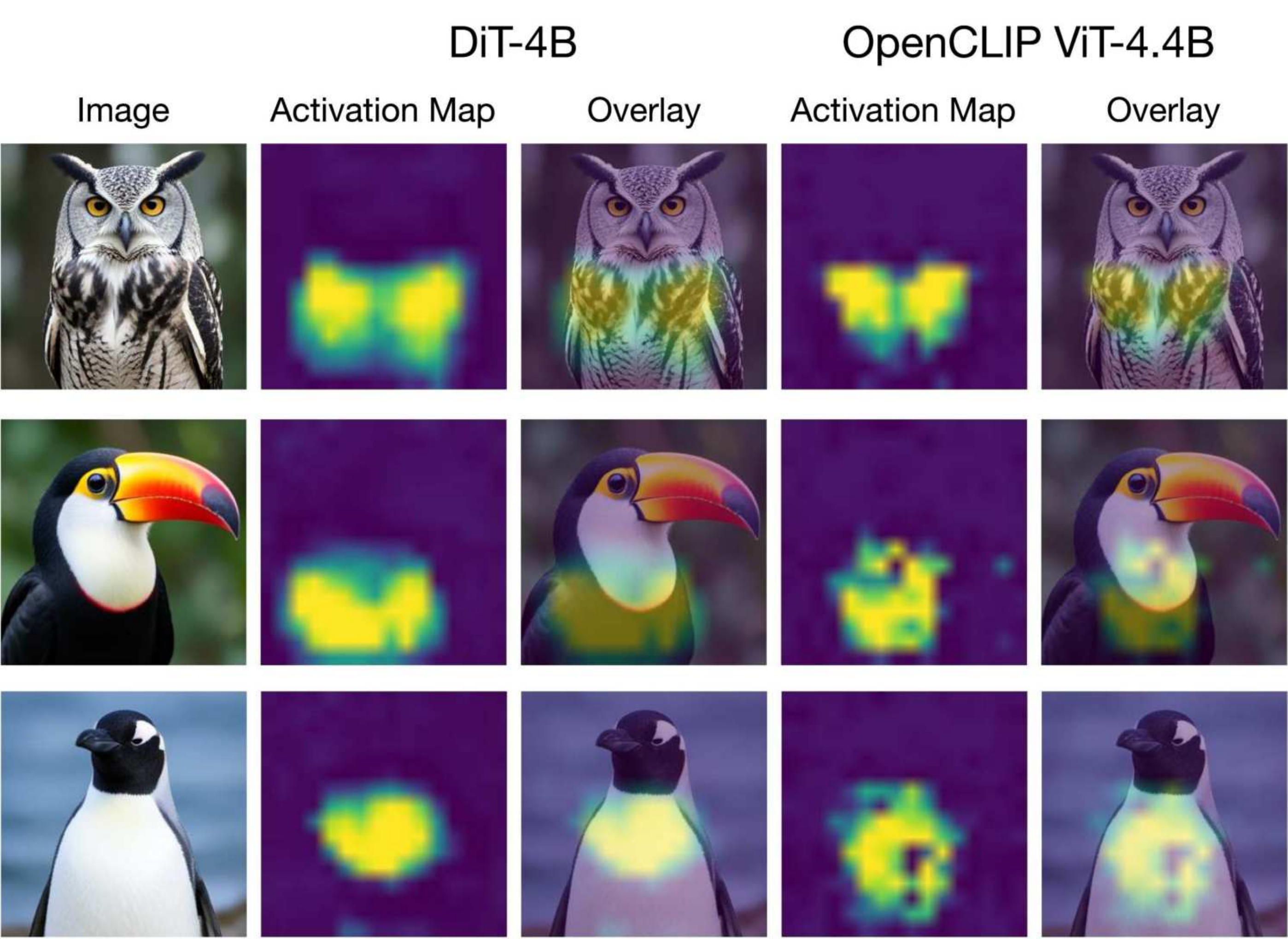}
        \caption{DiT-4B: L9/U3974. OpenCLIP ViT-4.4B: L45/U10399.}
        \label{fig:sub_c_flux}
    \end{subfigure}\hspace{1em} 
    \begin{subfigure}[b]{0.4\textwidth}
        \centering
        \includegraphics[width=\textwidth]{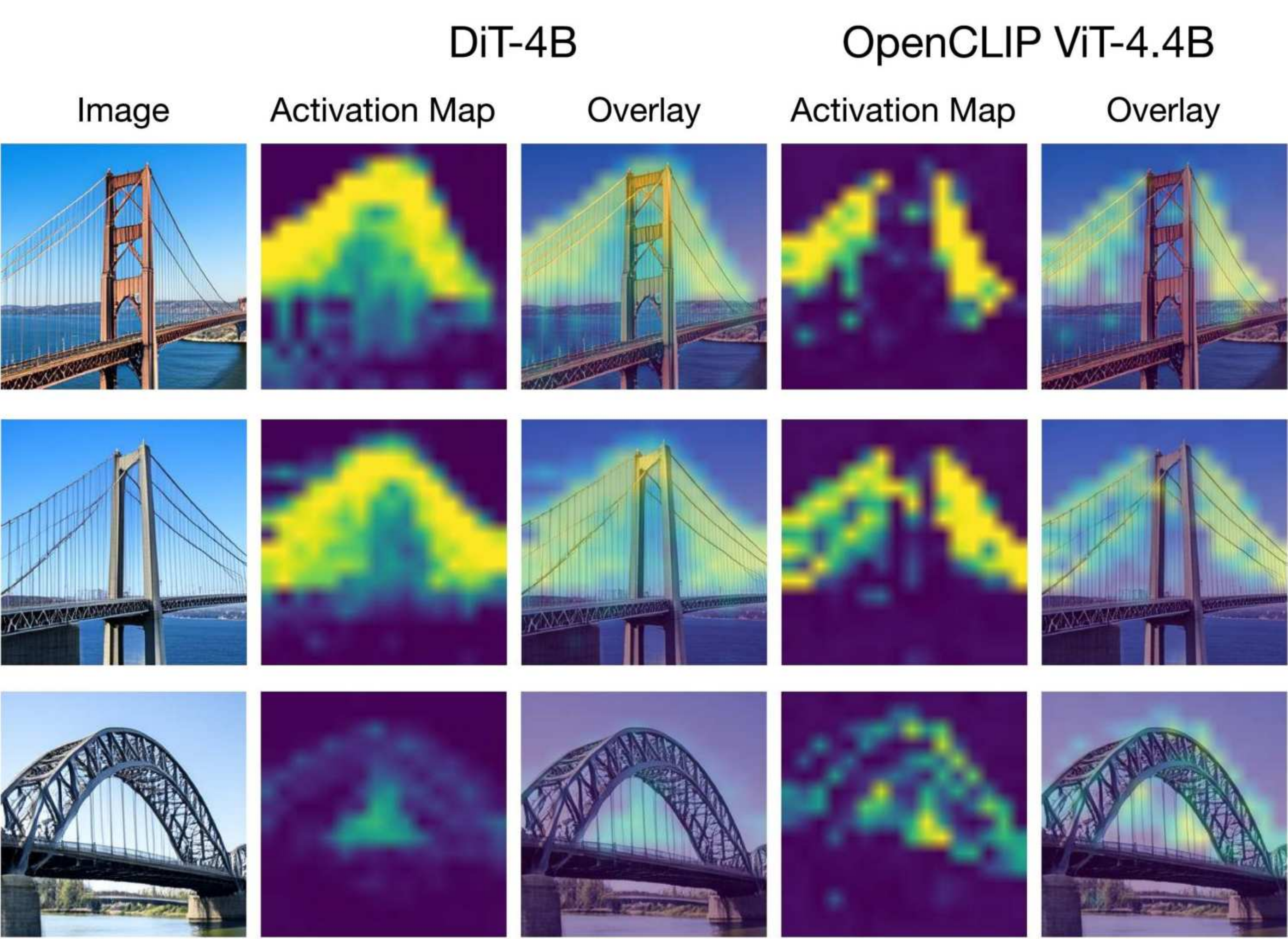}
        \caption{DiT-4B: L9/U8380. OpenCLIP ViT-4.4B: L41/U5498.}
        \label{fig:sub_d_flux}
    \end{subfigure}
    
    \caption{\textbf{Top-activating images for Rosetta Neurons.} Comparison between DiT-4B and OpenCLIP ViT-4.4B.}
    \label{fig:flux}
\end{figure}

We provide qualitative examples of Rosetta Neurons across model scales in both language and vision. For language models, we show LLM annotations for a subset of Rosetta Neurons across matched model families along with representative top-activating input spans  (\crefrange{fig:llm_qualitative0}{fig:llm_qualitative5}). The system prompt used for the LLM annotator is provided below. For vision models, we visualize the top three activating images and corresponding activation maps for a subset of Rosetta Neurons at different scales (\crefrange{fig:b_16}{fig:flux}). We report neuron identities in the captions using the format \texttt{L\{layer\}/U\{unit\}}, together with the coarse model names. Detailed model names are provided in~\Cref{app:model_families}.

\begin{user_prompt}
You are evaluating what a neuron in a language model responds to. Below are the top 20 sequences that most strongly activate this neuron that is shared across multiple language models. Each box shows a text sequence with a heatmap overlay indicating where the corresponding neuron in each model fires (bright = strong activation). Give an overall summary of the concept the neuron responds to in less than 10 words and give up to three examples of what sort of tokens it would fire on.
\end{user_prompt}

\setlength{\floatsep}{8pt plus 1pt minus 2pt}

\clearpage

\clearpage

\clearpage

\clearpage

\subsection{Qualitative Comparison of Rosetta and Non-Rosetta Neurons}
\label{subsec:non-rosetta-vs-rosetta}
\begin{figure}[h]
    \centering
    \includegraphics[width=0.95\linewidth]{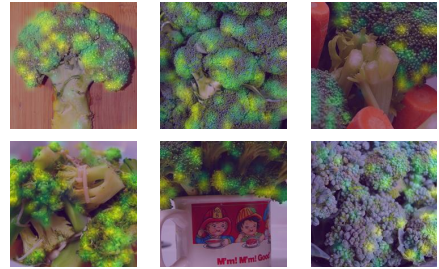}
    \caption{\textbf{Top-5 activating images for OpenCLIP ViT-L/14 Rosetta Neurons.} Rosetta Neurons demonstrate selective firing for coherent concepts.}
    \label{fig:rosetta_monosemantic_vision}
\end{figure}

We compare Rosetta and non-Rosetta neurons by visualizing top-activating examples. In language, we use randomly selected Pythia-6.9B Rosetta Neurons and non-Rosetta neurons from the same layers (\Cref{fig:rosetta_monosemantic_language,fig:random_polysemantic_language}); in vision, we conduct a similar comparison for OpenCLIP ViT-L/14 (\Cref{fig:rosetta_monosemantic_vision,fig:random_polysemantic_vision}). Rosetta Neurons appear more coherent and selective, whereas non-Rosetta neurons often respond to a broader mixture of features. This small sample is not standalone evidence, but it is consistent with the quantitative selectivity trends in~\Cref{sec:properties}.
\begin{figure}[h]
    \centering
    \includegraphics[width=0.9\linewidth]{figures/LLM_polysemanticity_a-compressed.pdf}
    \caption{\textbf{Top-5 activating sequences for Pythia-6.9B Rosetta Neurons.} Rosetta Neurons demonstrate selective firing for coherent concepts.}
    \label{fig:rosetta_monosemantic_language}
\end{figure}
\begin{figure}[h]
    \centering
    \includegraphics[width=0.9\linewidth]{figures/LLM_polysemanticity_b-compressed.pdf}
     \caption{\textbf{Top-5 activating sequences for Pythia-6.9B non-Rosetta neurons.} Neurons are randomly selected from the same layers as those in~\Cref{fig:rosetta_monosemantic_language}. These neurons appear polysemantic, firing for a variety of unrelated concepts.}
     \label{fig:random_polysemantic_language}
\end{figure}

\clearpage
\noindent
\begin{minipage}{\textwidth}
    \centering
    \includegraphics[width=0.95\linewidth]{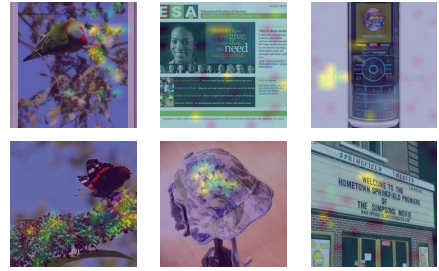}
    \captionof{figure}{\textbf{Top-5 activating images for OpenCLIP ViT-L/14 non-Rosetta neurons.} Neurons are randomly selected from the same layers as those in~\Cref{fig:rosetta_monosemantic_vision}. These neurons appear polysemantic, firing for a variety of unrelated concepts.}
    \label{fig:random_polysemantic_vision}
\end{minipage}

\section{Aligning Token-Wise Activations Across Models}
\label{sec:aligning_tokens}
In this section, we describe how activations are mapped to a shared set of aligned positions. This enables direct comparison of neuron activations across models, even when they use different tokenizers or patch grids.

\begin{figure}[h!]
    \centering
    \includegraphics[width=1.0\linewidth]{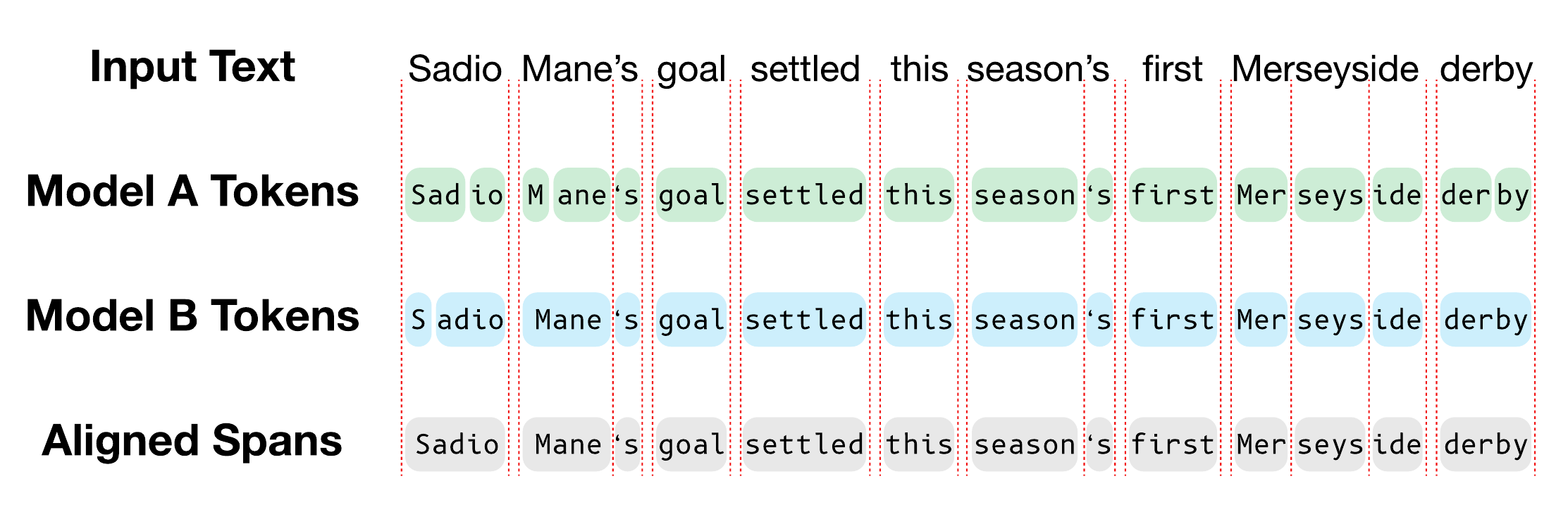}
   \caption{\textbf{Aligning text tokens via shared byte boundaries.} We align the tokens from Model $A$ and Model $B$ by keeping only the byte boundaries that both tokenizers share (red dashed lines). By finding the tokens that live in these new \emph{Aligned Spans}, we can pool the activations on these tokens, creating a shared set of positions that we can then use to compare neuron responses.}
    \label{fig:align_language_tokens}
\end{figure}

\paragraph{Language models.}
The same text may be tokenized differently across language models. To obtain a tokenizer-independent alignment, we represent the input text in UTF-8 byte space. For a text sequence \(x\), let the tokenizers for models $A$ and $B$ split this into tokens \(a_1,\dots,a_{T_A}\) and \(b_1,\dots,b_{T_B}\), respectively. Each token is associated with a contiguous byte span in the original text. We then define a canonical sequence of byte spans by taking the shared byte boundaries induced by the two tokenizations. These are represented by the dotted red lines in~\Cref{fig:align_language_tokens}. This procedure results in aligned text positions \(t=1,\dots,T^\ast\) that are independent of either tokenizer. For each canonical span \(t\), we mean pool the activations of any tokens overlapping that span, yielding one activation value \(m_t^u(x)\) per neuron \(u\) and aligned position \(t\). These pooled activations are then used for comparison according to the neuron similarity metric defined in \Cref{eq:pearson_correlation}.

\paragraph{Vision models.}

\begin{figure}[h!]
    \centering
    \includegraphics[width=1.0\linewidth]{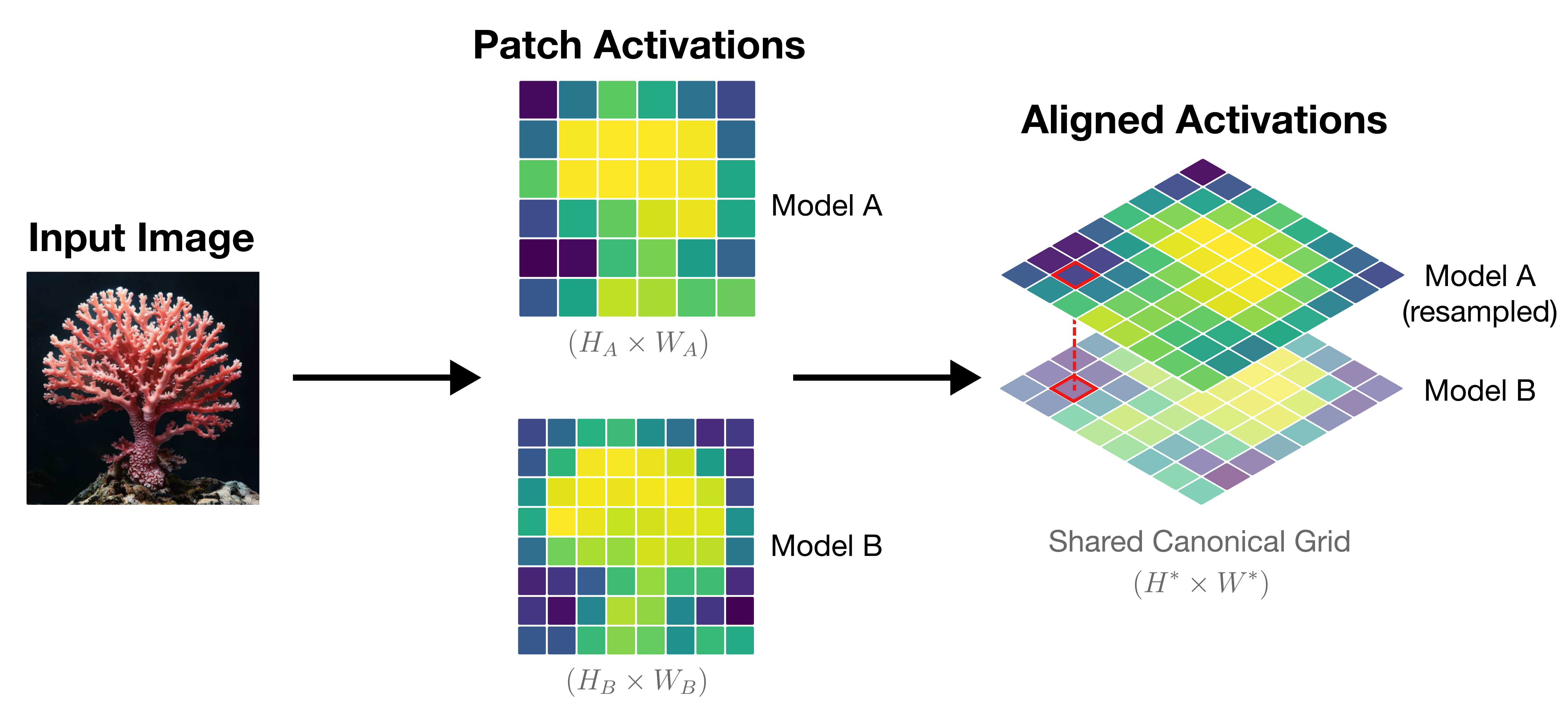}
    \caption{\textbf{Aligning spatial grids for neuron comparison.} We align the different patch grids from Model $A$ and Model $B$ by choosing a single target resolution (the canonical grid defined by Model $B$ in this case). Any mismatched native grids are resampled using bilinear interpolation to fit this shared grid. This results in aligned activation maps that can be used for measuring neuron similarity.}
    \label{fig:align_vision_tokens}
\end{figure}

Vision Transformers may use different patch sizes or input resolutions, producing activations on different spatial grids. For an image \(x\), let model \(A\) produce patch-token activations on a grid of size \(H_A \times W_A\), and model \(B\) produce activations on a grid of size \(H_B \times W_B\). We choose one of these grids as a canonical grid of size \(H^\ast \times W^\ast\). If a model's native grid differs from the canonical grid, we reshape its patch-token activations into a spatial feature map and resample that map to resolution \(H^\ast \times W^\ast\) using bilinear interpolation as illustrated in~\Cref{fig:align_vision_tokens}. This yields an aligned activation map \(m_{p,q}^u(x)\) for each neuron \(u\), defined on the same spatial cells \((p,q)\). We then flatten the canonical grid into a sequence of aligned spatial positions \(t=1,\dots,H^\ast W^\ast\), and use the resulting activations for neuron comparison according to \Cref{eq:pearson_correlation}. In practice, we remove any non-spatial prefix tokens such as class or register tokens, and then bilinearly interpolate the activation maps to have the same spatial dimensions according to the maximum of the two map sizes. 

\clearpage
\section{Further Details on Rosetta-Neuron Scaling}
\label{sec:rosetta_neuron_scaling_appendix}
We provide additional details on the model families used in the Rosetta Neuron scaling experiments from~\Cref{scaling_law_section}. We then present robustness checks for the matching procedure, including an ablation of the mutual-$k$ nearest-neighbor criterion and a dataset-permutation null. 

\subsection{Model Families}
\label{app:model_families}
\textbf{Language Model Families.} For language models, we only consider pretrained models that have not undergone post-training. We conduct the neuron matching using models from the Pythia, GPT-2, OPT, and Qwen-2.5 model families, spanning roughly 100 million to 30 billion parameters \citep{gpt2,pythia,opt,qwen2.5}. All models are included in~\Cref{tab:model_combo_language}.

\textbf{Vision Model Families.} We analyze discriminative vision models from the OpenCLIP, DINOv2, and Pixio families, spanning scales from approximately 80 million to 5 billion parameters~\citep{openclip, radford2021learning, dinov2, pixio}. For the generative model, we use one-step diffusion models built on the Diffusion Transformer architecture~\citep{peebles2023scalable}. We use models from the Pixel Mean Flow (pMF), Sana, and Flux families for diffusion~\citep{lu2026one, chen2025sana, flux-2-2025}. All models are included in~\Cref{tab:model_combo_vision}.

\vspace{5em}
\begin{table}[h]
\centering
\label{tab:run-configurations-language}
\begin{tabular}{@{}cl@{\hspace{2em}}cl@{}}
\toprule
Run \# & Models & Run \# & Models \\
\midrule
1  & OPT-1.3B, Qwen2.5-1.5B          & 12 & Pythia-160M, GPT2-124M              \\
2  & OPT-2.7B, Qwen2.5-3B          & 13 & Pythia-410M, GPT2-355M              \\
3  & OPT-6.7B, Qwen2.5-7B          & 14 & Pythia-1.4B, GPT2-1.5B              \\
4  & OPT-13B, Qwen2.5-14B          & 15 & OPT-125M, Pythia-160M, GPT2-124M       \\
5  & OPT-30B, Qwen2.5-32B          & 16 & OPT-350M, Pythia-410M, GPT2-355M       \\
6  & Pythia-160M, OPT-125M          & 17 & OPT-1.3B, Pythia-1.4B, GPT2-1.5B       \\
7  & Pythia-410M, OPT-350M          & 18 & Pythia-1.4B, OPT-1.3B, Qwen2.5-1.5B    \\
8  & Pythia-1.4B, OPT-1.3B           & 19 & Pythia-2.8B, OPT-2.7B, Qwen2.5-3B     \\
9  & Pythia-2.8B, OPT-2.7B           & 20 & Pythia-6.9B, OPT-6.7B, Qwen2.5-7B     \\
10 & Pythia-6.9B, OPT-6.7B           & 21 & Pythia-12B, OPT-13B, Qwen2.5-14B    \\
11 & Pythia-12B, OPT-13B           &    &                               \\
\bottomrule
\end{tabular}
\vspace{0.5em}
\caption{Model combinations used in the Rosetta Neuron scaling laws in~\Cref{fig:language_scaling}.}
\label{tab:model_combo_language}
\end{table}

\begin{table}[h]
\centering
\footnotesize
\setlength{\tabcolsep}{4pt}
\renewcommand{\arraystretch}{0.95}
\begin{tabular}{@{}cl@{\hspace{1.5em}}cl@{}}
\toprule
Run \# & Models & Run \# & Models \\
\midrule
1  & pMF DiT-B/16, DINOv2 ViT-B/14        & 12 & Sana DiT-1.6B, OpenCLIP ViT-bigG/14 \\
2  & pMF DiT-L/16, DINOv2 ViT-L/14        & 13 & Flux.2-Klein-DiT-4B, OpenCLIP EVA-02-CLIP-E/14 \\
3  & pMF DiT-H/16, DINOv2 ViT-g/14        & 14 & pMF DiT-B/16, Pixio ViT-B/16, OpenCLIP ViT-B/16 \\
4  & pMF DiT-B/16, Pixio ViT-B/16         & 15 & pMF DiT-L/16, Pixio ViT-L/16, OpenCLIP ViT-L/14 \\
5  & pMF DiT-L/16, Pixio ViT-L/16         & 16 & pMF DiT-H/16, Pixio ViT-H/16, OpenCLIP ViT-H/14 \\
6  & pMF DiT-H/16, Pixio ViT-H/16         & 17 & Sana DiT-1.6B, Pixio ViT-1B/16, OpenCLIP ViT-bigG/14 \\
7  & Sana DiT-1.6B, Pixio ViT-1B/16       & 18 & Flux.2-DiT-4B, Pixio ViT-5B/16, OpenCLIP EVA-02-E/14 \\
8  & Flux.2-Klein-DiT-4B, Pixio ViT-5B/16 & 19 & pMF DiT-B/16, DINOv2 ViT-B/14, OpenCLIP ViT-B/16 \\
9  & pMF DiT-B/16, OpenCLIP ViT-B/16      & 20 & pMF DiT-L/16, DINOv2 ViT-L/14, OpenCLIP ViT-L/14 \\
10 & pMF DiT-L/16, OpenCLIP ViT-L/14      & 21 & pMF DiT-H/16, DINOv2 ViT-g/14, OpenCLIP ViT-H/14 \\
11 & pMF DiT-H/16, OpenCLIP ViT-H/14      &    & \\
\bottomrule
\end{tabular}
\vspace{0.5em}
\caption{Model combinations used in the Rosetta Neuron scaling laws in~\Cref{fig:vision_scaling}.}
\label{tab:model_combo_vision}
\end{table}

\clearpage
\newpage

\begin{figure}[t]
    \centering

    \begin{subfigure}[t]{0.49\textwidth}
        \centering
        \includegraphics[
            width=\linewidth,
            trim={0bp 0bp 785.585bp 0bp},
            clip
        ]{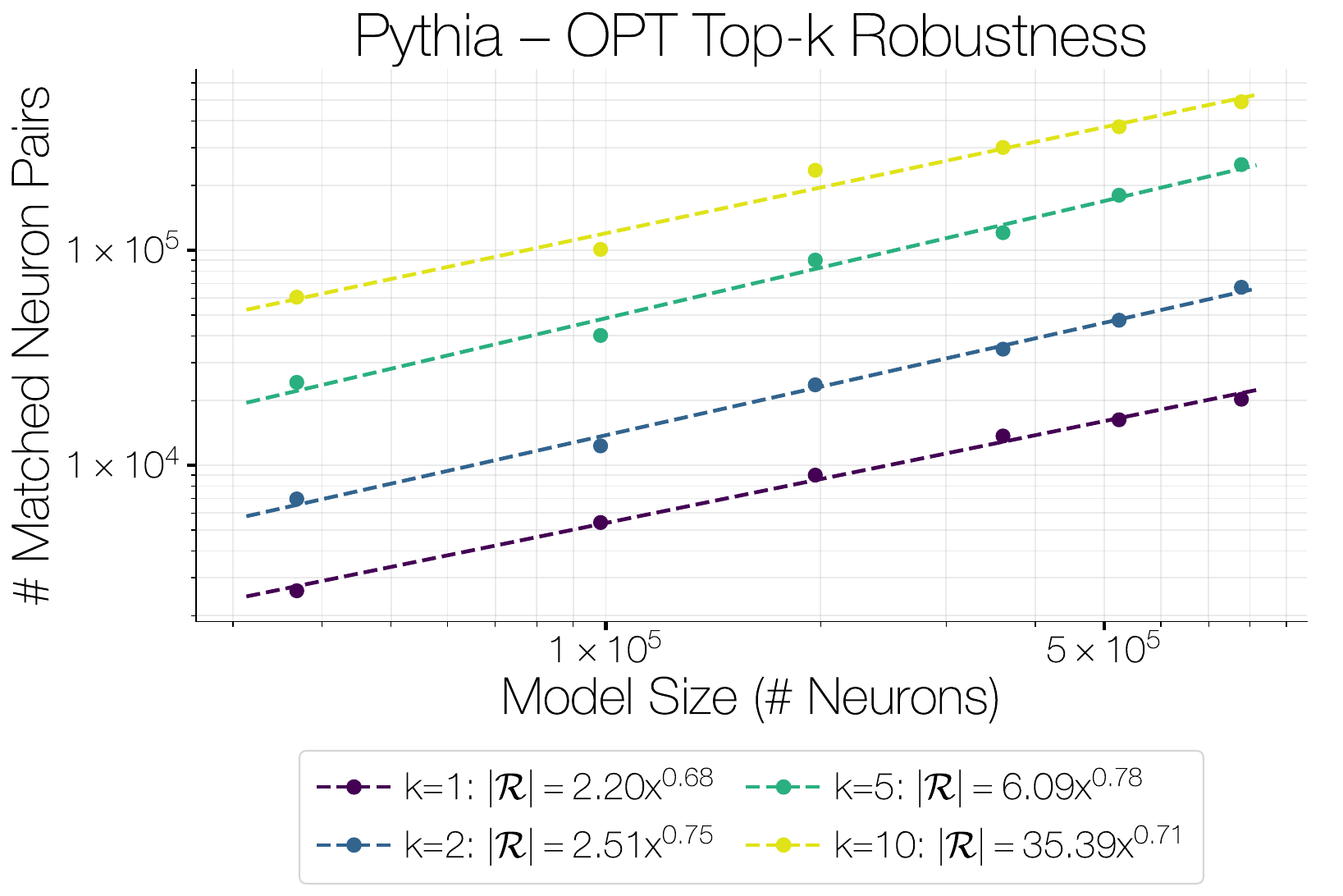}
    \end{subfigure}
    \hfill
    \begin{subfigure}[t]{0.49\textwidth}
        \centering
        \includegraphics[
            width=\linewidth,
            trim={785.585bp 0bp 0bp 0bp},
            clip
        ]{figures/topk_robustness.pdf}
    \end{subfigure}

    \caption{\textbf{Robustness to the mutual top-$k$ matching criterion.} We repeat the scaling analysis for Pythia--OPT and Diffusion--OpenCLIP for different values of $k$ in the nearest neighbor criterion. Increasing $k$ results in more discovered neuron pairs, but the fitted power-law exponents remain within a narrow sublinear range.}
    \label{fig:topk_ablation}
    \vspace{-1em}
\end{figure}

\subsection{Robustness to the Mutual Top-$k$ Criterion}
\label{subsec:topk_ablation}
Our main scaling experiments in~\Cref{scaling_law_section} identify Rosetta Neurons using mutual top-1 nearest-neighbor matching. To test whether the observed scaling behavior depends on this particular choice, we repeat the scaling analysis while varying the mutual top-$k$ parameter on one representative scaling trajectory per modality: Pythia--OPT for language models and Diffusion--OpenCLIP for vision models. As seen in~\Cref{fig:topk_ablation}, increasing $k$ makes the matching rule more permissive and increases the absolute number of discovered neuron pairs. However, across all tested values of $k$, the fitted power-law exponents remain within a narrow sublinear range in both modalities. This suggests that the Rosetta Neuron scaling trend is not an artifact of the default top-$k$ matching rule. It is also consistent with the detectability-threshold view in our analytical model (\Cref{sec:theory_supp}): varying the permissiveness of the matching criterion changes which features are counted as detectable and changes the prefactor of the scaling law, but should not change the underlying scaling exponent.

\subsection{Input-Permutation Null}
\label{supp:permutation_null}
In~\Cref{subsec:rosetta_powerlaw}, we applied the neuron matching pipeline to untrained random networks of different sizes to test whether Rosetta Neuron scaling laws could arise from the matching procedure. That baseline suggests the trend is not simply due to architecture, initialization, or the larger number of candidate neurons in bigger networks. Here, we test whether the observed scaling could still arise from activation marginals or high-dimensional nearest-neighbor effects even when input alignment is corrupted.
\begin{wrapfigure}{r}{0.45\textwidth}
    \vspace{5pt}
    \centering
    \includegraphics[width=0.45\textwidth]
    {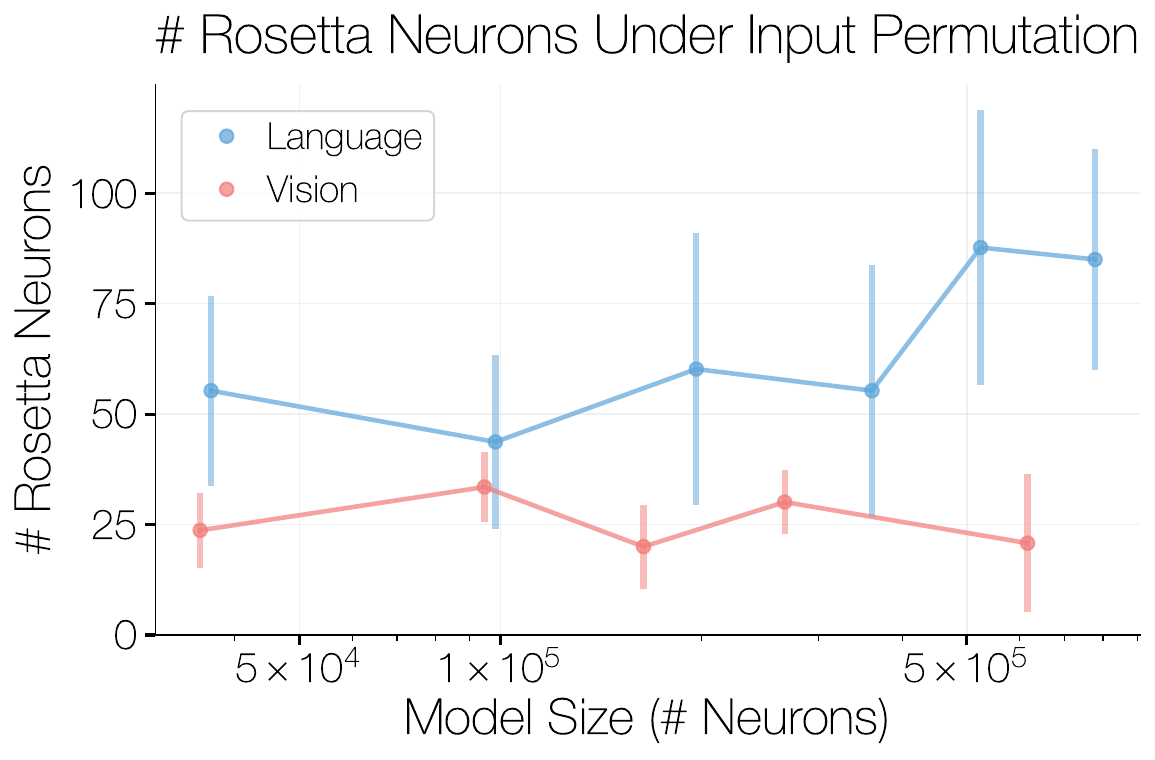}
    \vspace{-15pt}
    \caption{\textbf{Rosetta Neuron counts under input permutation lack systematic scaling.}}
   \label{fig:permutation_null}
    \vspace{-12pt}
\end{wrapfigure}

\textbf{Power-law scaling is absent under dataset permutation.} For each model pair, we first compute token-level activations on the same dataset used in the main experiments. Before computing cross-model correlations, we randomly permute the flattened activation positions for one model so that correlations are computed between mismatched input positions across the two models. This preserves each neuron's marginal activation distribution while corrupting input-wise alignment. We then apply the same mutual nearest-neighbor matching procedure with $k=1$ as in the main experiments. We run this null over three random permutations at each model scale and report means with 95\% confidence intervals in~\Cref{fig:permutation_null}. Under this null, the number of discovered Rosetta Neurons collapses to roughly 20--100 matches. Moreover, these counts no longer exhibit the systematic sublinear power-law trend observed with aligned activations. This suggests that Rosetta Neuron scaling depends on shared responses to the aligned inputs, rather than being induced by the matching procedure or activation statistics alone.

\clearpage
\section{Detailed Derivation of the Rosetta Neuron Scaling Model}
\label{sec:theory_supp}

This appendix gives the formal version of the analytical model for Rosetta Neuron scaling in~\Cref{subsec:analytical_model}. The model provides an explanation for the empirical observation that the number of Rosetta Neurons grows as a sublinear power law with scale. It further predicts that scaling separates an increasingly selective Rosetta population from a more crowded non-Rosetta background, a phenomenon we call the \textit{Neuron Polarization Effect}. We organize this section into four conceptual parts. We first state and solve a minimal allocation problem over latent features. Second, we motivate the assumptions of this allocation problem by interpreting scalar neuron readouts as Gaussian information channels in a superposition model. Third, we connect the idealized Rosetta Neuron detectability criterion to our empirical cross-model neuron matching. 
Finally, we use simulations to verify that the theory reproduces the predicted scaling and polarization trends when paired with our empirical matching procedure.

\subsection{A Minimal Capacity-Allocation Model}
\label{app:minimal-allocation-model}

We begin with a minimal capacity-allocation model that abstracts away architectural and optimization details while capturing the central constraint: a growing set of latent features must share a finite budget of clean neuron capacity.

\textbf{Feature spectrum.}
Let \(N\) denote the number of MLP neuron coordinates\ under consideration.\footnote{Throughout this appendix, we use “neuron” and “coordinate” interchangeably to refer to an individual MLP activation coordinate.} Let \(A(N)\) be the number of active latent features representing patterns from a data distribution \(D\). A network of size \(N\) allocates its neuron capacity across this feature set. We index these features by decreasing importance:
\[
r \in \{1,2,\ldots,A(N)\}.
\]

Since modern scaling regimes increase both model size and effective data complexity~\citep{hoffmann2022training}, we assume that the feature set does not saturate with model size. Specifically, we take
\begin{equation}    
    A(N)=\Omega(N),
    \qquad
    A(N)>N
\end{equation}

for sufficiently large $N$. Thus the model remains in a superposition regime asymptotically~\citep{liu2025superposition}: there are more latent features than neuron coordinates. Following prior spectral scaling theories
\citep{michaud2023quantization,bordelon2020spectrum, bahri2024explaining}, we assume a power-law feature-importance spectrum for these features: 
\begin{equation}
    w_r = C r^{-\beta},
    \qquad
    \beta>1,
    \qquad
    C>0.
    \label{eq:app-feature-importance-spectrum}
\end{equation}
Here, $w_r$ measures the importance, or downstream value of making feature $r$ cleanly readable, reflecting factors such as frequency, predictiveness, or task relevance.

\textbf{Feature isolation and Rosetta detectability.}
Let \(s_r \ge 0\) measure how cleanly feature \(r\) is isolated in its most dedicated neuron coordinate. Larger values of \(s_r\) indicate that this coordinate is dominated by feature \(r\) and is more monosemantic, rather than reflecting a mixture of unrelated features. At this stage, \(s_r\) is an abstract isolation score; in \cref{app:carrier-sir-model}, we derive \(s_r\) as a signal-to-interference ratio in a linear superposition model.

We assume that independently trained models of similar scale and modality are trained on comparable data distributions \(D\), and therefore share approximately the same coarse latent feature spectrum induced by \(D\). Under this shared-spectrum assumption, we analyze a representative network in isolation: features that are cleanly isolated in one model are expected to be similarly isolated in another matched-scale model, while model-specific interference averages out in cross-model correlations over aligned inputs. We make this connection explicit in \cref{supp:empirical-matching}. We therefore model Rosetta detectability by thresholding the isolation score. For a fixed detectability threshold \(\tau>1\), we define the idealized Rosetta count as
\begin{equation}
    R_\tau(N)
    =
    \sum_{r=1}^{A(N)} \mathbf{1}\{s_r\ge \tau\}.
    \label{eq:app-rosetta-count-abstract}
\end{equation}
The thresholded features are those cleanly isolated enough to be reproducible and detectable across independently trained models. For \(\tau>1\), this feature count corresponds to distinct dedicated neuron coordinates in the superposition model introduced in \cref{app:carrier-sir-model}, so we interpret \(R_\tau(N)\) as an idealized Rosetta Neuron count.

\textbf{Utility and budget.}
We now specify the allocation rule used in our analytical model. We state the allocation objective and constraint in~\Cref{eq:app-discrete-allocation}, and derive them in~\cref{app:objective-and-budget-microfoundation} from a simple superposition picture: neuron coordinates mix many latent features, and each coordinate provides a noisy scalar readout of any one feature. Under this view, the logarithmic utility \(w_r\log(1+s_r)\) comes from the importance-weighted information gain from improving a noisy readout; the linear budget \(\sum_{r=1}^{A(N)} s_r \le \kappa N\), for a constant \(\kappa>0\), comes from the bounded total variance of an \(N\)-coordinate activation vector. These ingredients give the following capacity-allocation problem:
\begin{equation}
    \max_{s_r\ge 0}\;\sum_{r=1}^{A(N)} w_r\log(1+s_r)
    \qquad
    \text{subject to}
    \qquad
    \sum_{r=1}^{A(N)} s_r\le \kappa N.
    \label{eq:app-discrete-allocation}
\end{equation}
The objective captures a coverage--purity tradeoff. Because the utility is logarithmic, increasing the isolation of a noisy feature initially gives substantial gain, but further purifying an already clean feature gives diminishing returns. The optimum therefore allocates high isolation to the most important features while also extending partial isolation into the lower-importance tail.

\subsection{Solving the Allocation Problem}
\label{app:solving-allocation}

\paragraph{Continuum approximation.}
To analyze the asymptotic scaling behavior, we approximate the ranked feature spectrum by a continuous variable $r\in[1,A(N)]$. The continuum version of \cref{eq:app-discrete-allocation} yields the following variational problem: 
\begin{equation}
    \max_{s(r)\ge 0}\;\int_1^{A(N)} C r^{-\beta}\log(1+s(r))\,dr
    \qquad
    \text{subject to}
    \qquad
    \int_1^{A(N)} s(r)\,dr\le \kappa N.
    \label{eq:app-continuum-allocation}
\end{equation}
The corresponding effective Rosetta Neuron count is
\begin{equation}
    R_\tau(N)
    =
    \int_1^{A(N)} \mathbf{1}\{s(r;N)\ge \tau\}\,dr.
    \label{eq:app-continuum-rosetta-count}
\end{equation}

\textbf{Proposition 1: Sublinear Rosetta Neuron scaling.}
Assume \(\beta>1\) and \(A(N)=\Omega(N)\), with \(A(N)>N\) for sufficiently large \(N\). Then the solution to the allocation problem in \cref{eq:app-continuum-allocation} is
\begin{equation}
    s^\star(r;N)
    =
    \left[
    \left(\frac{r_0(N)}{r}\right)^\beta - 1
    \right]_+,
    \qquad [u]_+ := \max\{u,0\},
    \label{eq:app-optimal-isolation-profile}
\end{equation}
where \(r_0(N)=\Theta(N^{1/\beta})\) is the largest feature rank receiving positive isolation. For every fixed threshold \(\tau>1\), the effective Rosetta Neuron count satisfies
\begin{equation}
    R_\tau(N)=\Theta\!\left(N^{1/\beta}\right).
    \label{eq:app-rosetta-scaling}
\end{equation}
Since \(\beta>1\), the Rosetta Neuron count grows sublinearly.

\paragraph{Proof.}
Introduce a Lagrange multiplier \(\lambda\) for the budget constraint. For any feature rank with positive isolation, \(s(r)>0\), the first-order optimality condition gives
\begin{equation}
    \frac{C r^{-\beta}}{1+s(r)}=\lambda.
    \label{eq:app-stationarity}
\end{equation}
Solving for \(s(r)\) gives \(s(r)=Cr^{-\beta}/\lambda-1\). Let \(r_0(N)\) denote the boundary of the positive-support region, so that \(s(r_0(N))=0\). Then \(\lambda=C r_0(N)^{-\beta}\), and therefore
\begin{equation}
    s^\star(r;N)
    =
    \left[\left(\frac{r_0(N)}{r}\right)^\beta-1\right]_+.
\end{equation}

Because \(\log(1+s(r))\) is strictly increasing, the optimum uses the full available isolation budget. Substituting the solution \(s^\star(r;N)\) into the budget constraint gives
\begin{align}
    \kappa N
    &=
    \int_1^{r_0(N)}
    \left[\left(\frac{r_0(N)}{r}\right)^\beta-1\right]dr \\
    &=
    r_0(N)^\beta\int_1^{r_0(N)} r^{-\beta}\,dr-(r_0(N)-1) \\
    &=
    \frac{r_0(N)^\beta}{\beta-1}
    -
    \frac{\beta r_0(N)}{\beta-1}
    +1
    =
    \Theta\!\left(r_0(N)^\beta\right).
    \label{eq:app-r0-budget}
\end{align}
Hence \(r_0(N)=\Theta(N^{1/\beta})\). Since \(\beta>1\), this cutoff grows sublinearly in \(N\), while \(A(N)=\Omega(N)\) grows at least linearly. Therefore \(r_0(N)=o(A(N))\), so for sufficiently large \(N\), the positively isolated features form an initial segment \(1 \le r \le r_0(N)\) within the full feature range \([1,A(N)]\).

Let \(r_\tau(N)\) denote the Rosetta frontier, defined by
\(s^\star(r_\tau(N);N)=\tau\). Using the optimal allocation profile,
this frontier satisfies
\begin{equation}
    \left(\frac{r_0(N)}{r_\tau(N)}\right)^\beta - 1 = \tau,
\end{equation}
and therefore
\begin{equation}
    r_\tau(N)
    =
    r_0(N)(1+\tau)^{-1/\beta}
    =
    \Theta(N^{1/\beta}).
    \label{eq:app-rosetta-frontier}
\end{equation}
Thus, counting the features above the Rosetta threshold amounts to counting feature ranks up to order \(r_\tau(N)\), so
\begin{equation}
    R_\tau(N)
    =
    \int_1^{A(N)}\mathbf{1}\{s^\star(r;N)\ge \tau\}\,dr
    =
    \Theta(r_\tau(N))
    =
    \Theta(N^{1/\beta}).
\end{equation}
Since \(\beta>1\), this growth is sublinear. \hfill\(\square\)

The optimal allocation partitions the spectrum into Rosetta-detectable features
    with $s_r \ge \tau$, partially isolated features with $0 < s_r < \tau$,
    strongly superposed features with $s_r = 0$, and features beyond the represented
    set $A(N)$. The frontiers $r_\tau(N)$ and $r_0(N)$ scale as
    $\Theta(N^{1/\beta})$, yielding the sublinear Rosetta Neuron count
    $R_\tau(N)=\Theta(N^{1/\beta})$. This is illustrated in~\Cref{fig:feature-isolation-frontier}.


\subsection{Prediction: Neuron Polarization}
\label{app:polarization-prediction}

The same allocation profile predicts a polarization effect. As \(N\) grows, the number of detectable Rosetta Neurons increases, and their average isolation also increases. At the same time, a growing tail of latent features remains weakly isolated, corresponding to a more crowded non-Rosetta background.

\textbf{Rosetta purification.}
Consider any fixed top-ranked feature \(r_\star\) (i.e., a feature with small rank index and high importance). Its isolation in the dedicated neuron coordinate scales as
\begin{equation}
    s^\star(r_\star;N)
    =
    \left(\frac{r_0(N)}{r_\star}\right)^\beta-1
    =
    \Theta\left(\frac{N}{r_\star^\beta}\right)-1
    \to \infty
    \qquad
    \text{as } N\to\infty.
    \label{eq:app-fixed-feature-purification}
\end{equation}
Thus, for any fixed high-importance feature, its isolation grows on the order of \(N\), so its dedicated neuron coordinate becomes increasingly dominated by that feature. Intuitively, this increasing isolation can be interpreted as greater selectivity and monosemanticity.

The Rosetta-detectable features are those with ranks \(1\le r\le r_\tau(N)\). We therefore define their average isolation as
\begin{align}
    \bar s_{\mathrm{Rosetta}}(N)
    &:=
    \frac{1}{r_\tau(N)}
    \int_1^{r_\tau(N)}
    \left[\left(\frac{r_0(N)}{r}\right)^\beta-1\right]dr \\
    &=
    \Theta\left(\frac{r_0(N)^\beta}{r_\tau(N)}\right)
    =
    \Theta\left(r_0(N)^{\beta-1}\right)
    =
    \Theta\left(N^{(\beta-1)/\beta}\right),
    \label{eq:app-rosetta-average-sir}
\end{align}
where we used \(r_\tau(N)=r_0(N)(1+\tau)^{-1/\beta}\). Since \(\beta>1\), this average isolation increases with scale, corresponding to a Rosetta population that becomes more selective and monosemantic on average.

\paragraph{Non-Rosetta crowding.}
We now analyze the isolation trend in the non-Rosetta feature tail. For Rosetta-detectable features, the threshold \(\tau>1\) lets us associate each feature with a distinct dedicated neuron coordinate. Below the threshold, this one-to-one interpretation no longer applies: non-Rosetta features may instead be distributed across multiple shared, superposed coordinates. So, we first analyze crowding on the feature side by measuring the average clean isolation assigned to features outside the Rosetta set:
\begin{equation}
    \bar s_{\mathrm{tail}}(N)
    :=
    \frac{1}{A(N)-R_\tau(N)}
    \int_{r_\tau(N)}^{A(N)} s^\star(r;N)\,dr.
    \label{eq:app-nonrosetta-feature-average}
\end{equation}

Only the partially isolated non-Rosetta region \(r\in[r_\tau(N),r_0(N)]\) contributes to the integral, since features beyond \(r_0(N)\) have zero isolation. Thus,
\begin{align}
    \int_{r_\tau(N)}^{A(N)} s^\star(r;N)\,dr
    &=
    \int_{r_\tau(N)}^{r_0(N)}
    \left[\left(\frac{r_0(N)}{r}\right)^\beta-1\right]dr \\
    &=
    r_0(N)
    \int_{(1+\tau)^{-1/\beta}}^1
    \left(u^{-\beta}-1\right)du \\
    &=
    \Theta(r_0(N))
    =
    \Theta(N^{1/\beta}),
    \label{eq:app-nonrosetta-numerator}
\end{align}
using the change of variables \(u=r/r_0(N)\). Since \(A(N)=\Omega(N)\), \(R_\tau(N)=\Theta(N^{1/\beta})\), and \(\beta>1\), the measure of the non-Rosetta feature tail, \(A(N)-R_\tau(N)\), grows asymptotically faster than \(N^{1/\beta}\). Therefore,
\begin{equation}
    \bar s_{\mathrm{tail}}(N)
    =
    O\left(\frac{N^{1/\beta}}{A(N)-R_\tau(N)}\right)
    =
    O\left(\frac{N^{1/\beta}}{A(N)}\right)
    \to 0.
    \label{eq:app-nonrosetta-average-vanishes}
\end{equation}

Equation~\eqref{eq:app-nonrosetta-average-vanishes} shows feature-side crowding: the average clean isolation assigned to an unresolved tail feature vanishes. We can also translate this into a neuron-level interpretation. The \(R_\tau(N)\) Rosetta-detectable features correspond to distinct dedicated neurons leaving \(N-R_\tau(N)=\Theta(N)\) non-Rosetta coordinates. The total clean isolation assigned to the unresolved tail is \(\Theta(N^{1/\beta})\), so the average tail-isolation mass per non-Rosetta neuron scales as
\begin{equation}
    O\!\left(\frac{N^{1/\beta}}{N}\right)
    =
    O\!\left(N^{(1-\beta)/\beta}\right)
    \to 0,
\end{equation}
since \(\beta>1\). Thus, on average, non-Rosetta coordinates receive vanishing clean tail-isolation mass. By definition, \(A(N)\) counts the active latent features that the network allocates capacity to represent, so the unresolved tail features are still part of the represented feature population. Since they are not cleanly isolated into dedicated coordinates, they must be represented in a distributed or superposed form. This gives the neuron-level interpretation: the non-Rosetta population forms a crowded, polysemantic background of weakly isolated features. 


\subsection{Deriving the Isolation Score from Superposition}
\label{app:carrier-sir-model}

The minimal model above treats $s_r$ as an abstract isolation score. We now derive this score from a simple linear superposition model of neuron activations.

\paragraph{Linear feature decomposition.}
Let \(x\) denote an aligned input position, such as a text token or image patch, and let \(\mathbf h(x)\in\mathbb R^N\) be the centered vector of MLP neuron activations at that position, with centering taken over the data distribution \(D\). For each latent feature \(r\in\{1,\ldots,A(N)\}\), let \(\tilde z_r(x)\in\mathbb R\) be the centered, input-dependent scalar response of feature \(r\). Let \(\mathbf v_r\in\mathbb R^N\) describe how that feature response is distributed across neuron coordinates. We model the activation vector as a linear superposition of feature responses,
\begin{equation}
    \mathbf h(x)
    =
    \sum_{r=1}^{A(N)} \tilde z_r(x)\mathbf v_r.
    \label{eq:app-linear-superposition}
\end{equation}
We normalize the feature directions as \(\|\mathbf v_r\|_2^2=1\), so \(\mathbf v_r\) specifies the relative distribution of feature \(r\) across coordinates, while the overall strength of the feature response on input \(x\) is carried by the scale of \(\tilde z_r(x)\).

For the isolation-score calculation that follows, we first standardize each feature activation over the data distribution. Since \(\tilde z_r(x)\) is centered, define
\begin{equation}
    g_r^2 := \mathbb E_{x\sim D}[\tilde z_r(x)^2],
    \qquad
    z_r(x):=\tilde z_r(x)/g_r,
    \label{eq:app-feature-standardization}
\end{equation}

for represented features with \(g_r>0\). Then \(z_r(x)\) has mean zero and unit variance over \(D\), while \(g_r\) carries the overall activation scale of feature \(r\). For tractability, we use a diagonal second-moment approximation, keeping only the variances of the standardized feature activations and neglecting cross-feature covariances:

\begin{equation}
    \mathbb E_{x\sim D}[z_r(x)]=0,
    \qquad
    \mathbb E_{x\sim D}[z_r(x)^2]=1,
    \qquad
    \mathbb E_{x\sim D}[z_r(x)z_s(x)]=0
    \quad (r\neq s).
    \label{eq:app-feature-decorrelation}
\end{equation}
The representation can be written as
\begin{equation}
    \mathbf h(x)
    =
    \sum_{r=1}^{A(N)} \tilde z_r(x)\mathbf v_r
    =
    \sum_{r=1}^{A(N)} g_r z_r(x)\mathbf v_r.
    \label{eq:app-feature-decomp}
\end{equation}
This normalization will let us define the variance contribution of each feature to each neuron coordinate in the next step.

Define the effective loading of feature $r$ on neuron coordinate $j$ by
\begin{equation}
    W_{jr}:=g_r\mathbf e_j^\top\mathbf v_r,
    \qquad
    a_{jr}:=W_{jr}^2.
    \label{eq:app-effective-loading}
\end{equation}
Note that \(W_{jr}\) is an effective feature loading onto neuron coordinates in the superposition model, not a learned MLP weight matrix.

Because $\|\mathbf v_r\|_2^2=1$,
\begin{equation}
    \sum_{j=1}^N a_{jr}=g_r^2.
    \label{eq:app-loading-sums-to-amplitude}
\end{equation}
The activation of coordinate $j$ is
\begin{equation}
    h_j(x)=\sum_{r=1}^{A(N)} W_{jr}z_r(x).
    \label{eq:app-coordinate-activation}
\end{equation}
Using the decorrelation approximation in \cref{eq:app-feature-decorrelation},
\begin{equation}
    \operatorname{Var}_x(h_j)
    =
    \sum_{r=1}^{A(N)} W_{jr}^2
    =
    \sum_{r=1}^{A(N)} a_{jr}.
    \label{eq:app-coordinate-variance}
\end{equation}
Thus $a_{jr}$ is the variance contribution of feature $r$ to neuron coordinate $j$. We next use this to bound the total variance across coordinates. We assume the normalized activation vector has bounded average variance per coordinate:
\begin{equation}
    \sum_{j=1}^N \operatorname{Var}_x(h_j)
    =
    \sum_{j=1}^N\sum_{r=1}^{A(N)} W_{jr}^2
    =
    \sum_{r=1}^{A(N)} g_r^2
    \le C_h N,
    \label{eq:app-activation-energy}
\end{equation}
for a constant $C_h$ independent of $N$. This says that an $N$-coordinate normalized activation vector carries $O(N)$ total variance, consistent with standard scale-preserving initialization and normalization schemes~\citep{he2015delving, yang2021tuning, ba2016layer}.

\paragraph{Signal-to-interference isolation score.}
A single MLP neuron coordinate is a scalar readout of a richer, potentially high-dimensional latent feature state. When using coordinate \(j\) to read out feature \(r\), we decompose the coordinate into the desired contribution from feature \(r\) and a residual term:
\begin{equation}
    y_j = W_{jr}z_r+\eta_{jr},
    \label{eq:app-scalar-readout}
\end{equation}
where \(\eta_{jr}\) contains all variation in coordinate \(j\) not explained by feature \(r\). We model this residual as having two sources: interference from other modeled features that also contribute to coordinate \(j\), and a fixed finite-resolution noise floor \(\sigma_\infty^2>0\) that captures unmodeled variation. Under the diagonal second-moment approximation above, the effective residual variance is
\begin{equation}
    \nu_{jr}
    =
    \sigma_\infty^2+
    \sum_{k\neq r} W_{jk}^2.
    \label{eq:app-residual-variance}
\end{equation}

We define the signal-to-interference ratio for reading feature \(r\) from coordinate \(j\) as
\begin{equation}
    s_{jr}
    :=
    \frac{W_{jr}^2}
    {\sigma_\infty^2+\sum_{k\neq r}W_{jk}^2}.
    \label{eq:app-coordinate-sir}
\end{equation}
The most dedicated neuron coordinate for feature \(r\) is the coordinate with maximal signal-to-interference ratio,
\begin{equation}
    c(r)=\arg\max_j s_{jr},
    \label{eq:app-canonical-carrier}
\end{equation}
and the feature-level isolation score is
\begin{equation}
    s_r:=s_{c(r),r}
    =
    \frac{W_{c(r),r}^2}
    {\sigma_\infty^2+\sum_{k\neq r}W_{c(r),k}^2}.
    \label{eq:app-sir-definition}
\end{equation}
Large \(s_r\) means that the most dedicated coordinate for feature \(r\) is dominated by that feature and more monosemantic; small \(s_r\) means that the feature is read out through a more polysemantic mixture.

This definition also explains why the threshold \(\tau>1\) in \cref{eq:app-rosetta-count-abstract} gives a one-to-one feature-to-coordinate count in the idealized model. Suppose two distinct features \(r\) and \(r'\) shared the same threshold-passing most dedicated coordinate \(j\). Since \(s_r>1\),
\[
    W_{jr}^2
    >
    \sigma_\infty^2+
    \sum_{k\neq r} W_{jk}^2
    \ge
    W_{jr'}^2.
\]
Similarly, \(s_{r'}>1\) implies \(W_{jr'}^2>W_{jr}^2\), a contradiction. Therefore, each threshold-passing feature has its own dedicated neuron coordinate.

\subsection{Deriving the Logarithmic Utility and Linear Isolation Budget}
\label{app:objective-and-budget-microfoundation}
We now derive the logarithmic utility and linear isolation budget used in \cref{eq:app-discrete-allocation} based on the signal-to-interference model above.

\paragraph{Logarithmic utility from predictive loss reduction.}
Recall that \(z_r\) has been standardized so that \(\mathbb E_{x\sim D}[z_r(x)]=0\) and \(\operatorname{Var}_{x\sim D}(z_r(x))=1\). For tractability, we use a Gaussian approximation to obtain a closed-form relationship between readout quality and predictive uncertainty. Specifically, we model \(z_r\) as Gaussian and treat the residual interference $\eta$ in the most dedicated coordinate as independent Gaussian noise. After choosing the most dedicated coordinate \(c(r)\), its readout is
\begin{equation}
    y_{c(r)}
    =
    W_{c(r),r}z_r+\eta_{c(r),r}.
\end{equation}
We divide by the feature coefficient \(W_{c(r),r}\) and obtain the normalized observation model
\begin{equation}
    \tilde y_r
    :=
    \frac{y_{c(r)}}{W_{c(r),r}}
    =
    z_r+\epsilon_r,
    \qquad
    z_r\sim\mathcal N(0,1),
    \qquad
    \epsilon_r\sim\mathcal N(0,1/s_r).
    \label{eq:app-rescaled-carrier-observation}
\end{equation}
For this Gaussian model, the posterior variance is
\begin{equation}
    \operatorname{Var}(z_r\mid \tilde y_r)=\frac{1}{1+s_r}.
    \label{eq:app-posterior-variance}
\end{equation}
Thus the optimal Gaussian negative log-likelihood, or equivalently the conditional entropy, is
\begin{equation}
    H(z_r\mid \tilde y_r)
    =
    \frac12\log\left(2\pi e\,\frac{1}{1+s_r}\right).
    \label{eq:app-posterior-entropy}
\end{equation}
Relative to having no informative readout, the reduction in optimal predictive loss is
\begin{equation}
    H(z_r)-H(z_r\mid \tilde y_r)
    =
    \frac12\log(1+s_r).
    \label{eq:app-uncertainty-reduction}
\end{equation}
We therefore model a feature with downstream importance \(w_r\) contributing utility proportional to \(w_r\log(1+s_r)\), with the factor \(1/2\) absorbed into the overall scale.\footnote{Equivalently, this is the standard scalar Gaussian-channel calculation underlying the Shannon--Hartley theorem: mutual information grows as \(\frac12\log(1+\mathrm{SNR})\), with \(s_r\) playing the role of the SNR.}

\paragraph{Linear Isolation Budget.}
From the definition of the feature-level isolation score in \cref{eq:app-sir-definition},
\begin{equation}
    s_r
    \le
    \frac{W_{c(r),r}^2}{\sigma_\infty^2}.
    \label{eq:app-sir-upper-bound}
\end{equation}
Summing over features and using the activation-energy bound in \cref{eq:app-activation-energy},
\begin{equation}
    \sum_{r=1}^{A(N)} s_r
    \le
    \sigma_\infty^{-2}
    \sum_{r=1}^{A(N)} W_{c(r),r}^2
    \le
    \sigma_\infty^{-2}
    \sum_{j=1}^{N}\sum_{r=1}^{A(N)} W_{jr}^2
    \le
    \frac{C_h}{\sigma_\infty^2}N.
    \label{eq:app-derived-sir-budget}
\end{equation}
Therefore
\begin{equation}
    \sum_{r=1}^{A(N)} s_r\le \kappa N,
    \qquad
    \kappa:=\frac{C_h}{\sigma_\infty^2}.
    \label{eq:app-sir-budget}
\end{equation}
The linear budget thus follows from normalized activation energy and an irreducible finite-resolution floor for scalar neuron readouts.

\subsection{Connection to Empirical Rosetta Matching}
\label{supp:empirical-matching}

The theory defines Rosetta-detectable features by thresholding the feature-level isolation score \(s_r\). In the experiments, however, Rosetta Neurons are identified by cross-model activation matching rather than by directly observing \(s_r\). In this section, we relate these two notions by showing that, in an idealized setting, the same isolation score that defines Rosetta detectability also controls the cross-model activation correlation used for empirical matching. A high-dimensional random-packing approximation captures the effect of model-specific superposition: independent models mix unresolved tail features in nearly orthogonal ways, so reliable cross-model neuron matches require sufficiently isolated common features.

\textbf{Superposition as model-specific interference.} Consider two independently trained matched-scale models \(M\) and \(M'\), trained on comparable data distributions \(D\). As in the allocation model, we assume these models share approximately the same coarse latent feature spectrum induced by \(D\), although their model-specific interference patterns may differ because low-isolation tail features can be packed in many approximately equivalent ways. Let \(c_M(r)\) and \(c_{M'}(r)\) denote their most dedicated neuron coordinates for feature \(r\), with feature-level isolation scores \(s_r^M\) and \(s_r^{M'}\).

Since empirical matching uses Pearson correlation, arbitrary rescalings of a neuron activation do not affect the match score. For feature \(r\), the relevant neurons are the most dedicated coordinates \(c_M(r)\) and \(c_{M'}(r)\), whose raw readouts are
\begin{equation}    
    y^M_{c_M(r)}
    =
    W^M_{c_M(r),r} z_r+\eta^M_{c_M(r),r},
    \qquad
    y^{M'}_{c_{M'}(r)}
    =
    W^{M'}_{c_{M'}(r),r} z_r+\eta^{M'}_{c_{M'}(r),r}.
\end{equation}

We divide each readout by the corresponding feature loading,
\(W^M_{c_M(r),r}\) or \(W^{M'}_{c_{M'}(r),r}\), which removes arbitrary activation scale that does not affect Pearson correlation. This leaves normalized readouts consisting of the shared feature signal plus rescaled model-specific interference:
\begin{equation}
    \tilde y^M_{c_M(r)} = z_r+\epsilon_M,
    \qquad
    \tilde y^{M'}_{c_{M'}(r)} = z_r+\epsilon_{M'}.
\end{equation}

Here \(z_r\) is the shared feature signal, while \(\epsilon_M\) and \(\epsilon_{M'}\) are model-specific interference terms. Recall that the shared feature \(z_r\) has unit variance, and that the rescaled interference variances are determined by the feature-level isolation scores:
\[
    \operatorname{Var}(z_r)=1,
    \qquad
    \operatorname{Var}(\epsilon_M)=1/s_r^M,
    \qquad
    \operatorname{Var}(\epsilon_{M'})=1/s_r^{M'}.
\]

\textbf{Tail leakage and cross-model covariance.} We separate the feature-dependent part of the interference by writing it as a mixture over unresolved tail features:
\begin{equation}
    \epsilon^{\mathrm{feat}}_M
    =
    \sum_{k\in T_r^M} b^M_{c_M(r),k}z_k,
    \qquad
    \epsilon^{\mathrm{feat}}_{M'}
    =
    \sum_{k\in T_r^{M'}} b^{M'}_{c_{M'}(r),k}z_k ,
\end{equation}
Here \(T_r^M\) and \(T_r^{M'}\) denote the non-target features that leak into the scalar readout of feature \(r\) from the selected coordinates \(c_M(r)\) and \(c_{M'}(r)\). The coefficients \(b^M_{c_M(r),k}\) and \(b^{M'}_{c_{M'}(r),k}\) are the corresponding normalized leakage coefficients, obtained after rescaling by the target-feature loadings \(W^M_{c_M(r),r}\) and \(W^{M'}_{c_{M'}(r),r}\). This tail-packing calculation focuses on the feature-dependent part of the interference; the residual noise-floor variation is treated as independent across models and therefore does not contribute to cross-model covariance. Since the total normalized interference variances are \(1/s_r^M\) and \(1/s_r^{M'}\), the feature-leakage coefficients satisfy
\begin{equation}
    \sum_{k\in T_r^M} \left(b^M_{c_M(r),k}\right)^2 \le 1/s_r^M,
    \qquad
    \sum_{k\in T_r^{M'}} \left(b^{M'}_{c_{M'}(r),k}\right)^2 \le 1/s_r^{M'}.
\end{equation}
To compute \(\operatorname{Cov}(\epsilon^{\mathrm{feat}}_M,\epsilon^{\mathrm{feat}}_{M'})\), we expand the two residual mixtures, which gives a double sum over pairs of leaked features. Under the diagonal feature approximation, \(\operatorname{Cov}(z_k,z_\ell)=0\) for \(k\neq \ell\) and \(\operatorname{Var}(z_k)=1\). Therefore, only the same latent feature appearing in both residual mixtures contributes to cross-model covariance:
\begin{equation}
    \operatorname{Cov}(\epsilon^{\mathrm{feat}}_M,\epsilon^{\mathrm{feat}}_{M'})
    =
    \sum_{k\in T_r^M\cap T_r^{M'}}
    b^M_{c_M(r),k}
    b^{M'}_{c_{M'}(r),k}.
\end{equation}

\textbf{Random tail packing.} This term is small in a simple random tail-packing approximation. Suppose the tail available to the readout of feature \(r\) has effective size \(L_r\), where \(L_r\) is the effective number of unresolved non-target features that could contribute interference to this scalar readout. We view each set of leakage coefficients as a vector over this \(L_r\)-dimensional tail, with zero entries for features absent from the mixture. If the two models choose tail-mixing directions approximately independently and isotropically in this space, then
\begin{equation}
    \mathbb{E}\!\left[
    \operatorname{Cov}(\epsilon^{\mathrm{feat}}_M,\epsilon^{\mathrm{feat}}_{M'})
    \right]
    =0,
\end{equation}
and
\begin{equation}
    \operatorname{Var}\!\left[
    \operatorname{Cov}(\epsilon^{\mathrm{feat}}_M,\epsilon^{\mathrm{feat}}_{M'})
    \right]
    =
    O\!\left(
    \frac{\|b^M_{c_M(r)}\|_2^2\|b^{M'}_{c_{M'}(r)}\|_2^2}{L_r}
    \right)
    \le
    O\!\left(\frac{1}{s_r^M s_r^{M'}L_r}\right).
\end{equation}
\textbf{Isolation controls matching.} Thus, as the unresolved-tail dimension \(L_r\) grows, independently packed feature-dependent interference becomes asymptotically uncorrelated across models. Any remaining noise-floor variation is also treated as independent across models. We therefore model \(\epsilon_M\) and \(\epsilon_{M'}\) as uncorrelated across models and uncorrelated with the shared feature signal. Under this approximation, we obtain
\begin{equation}
    \operatorname{Corr}(\tilde y^M_{c_M(r)},\tilde y^{M'}_{c_{M'}(r)})
    =
    \frac{1}
    {\sqrt{(1+1/s_r^M)(1+1/s_r^{M'})}}.
    \label{eq:app-sir-correlation}
\end{equation}

Cross-model activation correlation is monotonically increasing in the feature-level isolation score of each model. It follows that a high-isolation feature is more likely to produce a high-correlation cross-model neuron score. We next connect this score-level relationship to the notion of Rosetta Neuron matching.

For two models and a fixed detectability threshold \(\tau>1\), the corresponding idealized two-model Rosetta Neuron count is
\begin{equation}
    R_\tau^{M,M'}(N)
    =
    \sum_{r=1}^{A(N)}
    \mathbf{1}\{\min(s_r^M,s_r^{M'})\ge \tau\}.
    \label{eq:app-two-model-rosetta-count}
\end{equation}

The condition \(\tau>1\) gives the idealized count a one-feature, one-coordinate interpretation: threshold-passing features have distinct dedicated neuron coordinates. This mirrors the empirical matching rule, where we use mutual nearest-neighbor matching with \(k=1\) under Pearson correlation. This enforces one-to-one cross-model neuron matches. By \cref{eq:app-sir-correlation}, higher isolation scores imply higher cross-model correlation. So, mutual nearest-neighbor matching can be interpreted as an empirical detectability filter: it prefers to keep features whose shared signal dominates model-specific interference in both models.


\subsection{Synthetic Validation of the Analytical Model}
\label{supp:simulation}

We perform a synthetic feature-packing experiment to test whether our phenomenological model produces sublinear Rosetta Neuron scaling and neuron polarization. We embed the theory's predicted isolation profile into synthetic activations for two independent networks. We then recover Rosetta Neurons using the same Pearson-correlation mutual-nearest-neighbor procedure used in our main experiments and compare the recovered scaling and polarization trends to the theory's predictions.

\textbf{Experimental setup.}
For each neuron budget \(N\), we instantiate \(A=\gamma N\) latent features with importance weights \(w_r \propto r^{-\beta}\). We then solve the isolation-allocation problem with budget \(\sum_r s_r \le \kappa N\) to obtain predicted isolation scores \(s_r\). For each synthetic input \(x\), we sample latent feature responses \(z_r(x)\) for all features \(r\). We assign every feature with positive isolation, \(s_r>0\), to a synthetic neuron coordinate. Features with \(s_r=0\) are not assigned as target features, but may still appear as background interference. For a neuron \(j\) assigned to feature \(r\), we sample a background set \(\mathcal{B}_j\) of non-target features and write the activation as
\begin{equation}
h_j(x)
=
W_{j,r}z_r(x)
+
\sum_{k\in\mathcal{B}_j} W_{j,k}z_k(x)
+
\xi_j(x),
\end{equation}
where \(\xi_j(x)\) is an irreducible noise-floor term with variance \(\sigma_\infty^2\). We draw the off-target loadings \(W_{j,k}\) for \(k\in\mathcal{B}_j\) as a random isotropic Gaussian direction and normalize them to a fixed interference norm. We then choose the target-feature loading \(W_{j,r}\) so that the resulting signal-to-interference ratio matches the predicted isolation score:
\begin{equation}
\frac{W_{j,r}^2}
{\sum_{k\in\mathcal{B}_j} W_{j,k}^2+\sigma_\infty^2}
\approx
s_r.
\end{equation}
The background mixtures and noise are sampled independently across the two synthetic models, while the latent feature responses \(z_r(x)\) are shared.

\textbf{Evaluation protocol.} We run the simulation across neuron budgets \(N \in \{2^{10}, \ldots, 2^{17}\}\) and feature-spectrum exponents \(\beta \in \{1.5, 1.75, 2, 2.25, 2.5, 3\}\). For each \((N,\beta)\) setting, we report the mean over 10 independent runs; error bars are omitted because the standard errors are smaller than the plotted markers. For each run, we sample \(M\) synthetic inputs by drawing a shared latent feature-response matrix \(Z \in \mathbb{R}^{M \times A}\) with i.i.d. Gaussian entries, then standardize each feature column across inputs. The same \(Z\) is used for both synthetic models, while the neuron-level feature mixtures and noise are sampled independently across models. We then apply Pearson-correlation mutual-nearest-neighbor matching between the two synthetic models. We compare the recovered Rosetta count against the theoretical scaling prediction \(R_\tau(N)=\Theta(N^{1/\beta})\). To test the polarization effect, we separately track the mean isolation of recovered Rosetta neurons and non-Rosetta neurons as \(N\) increases.

\begin{figure}[h]
    \centering

    \begin{subfigure}{0.58\textwidth}
        \centering
        \includegraphics[width=\linewidth]{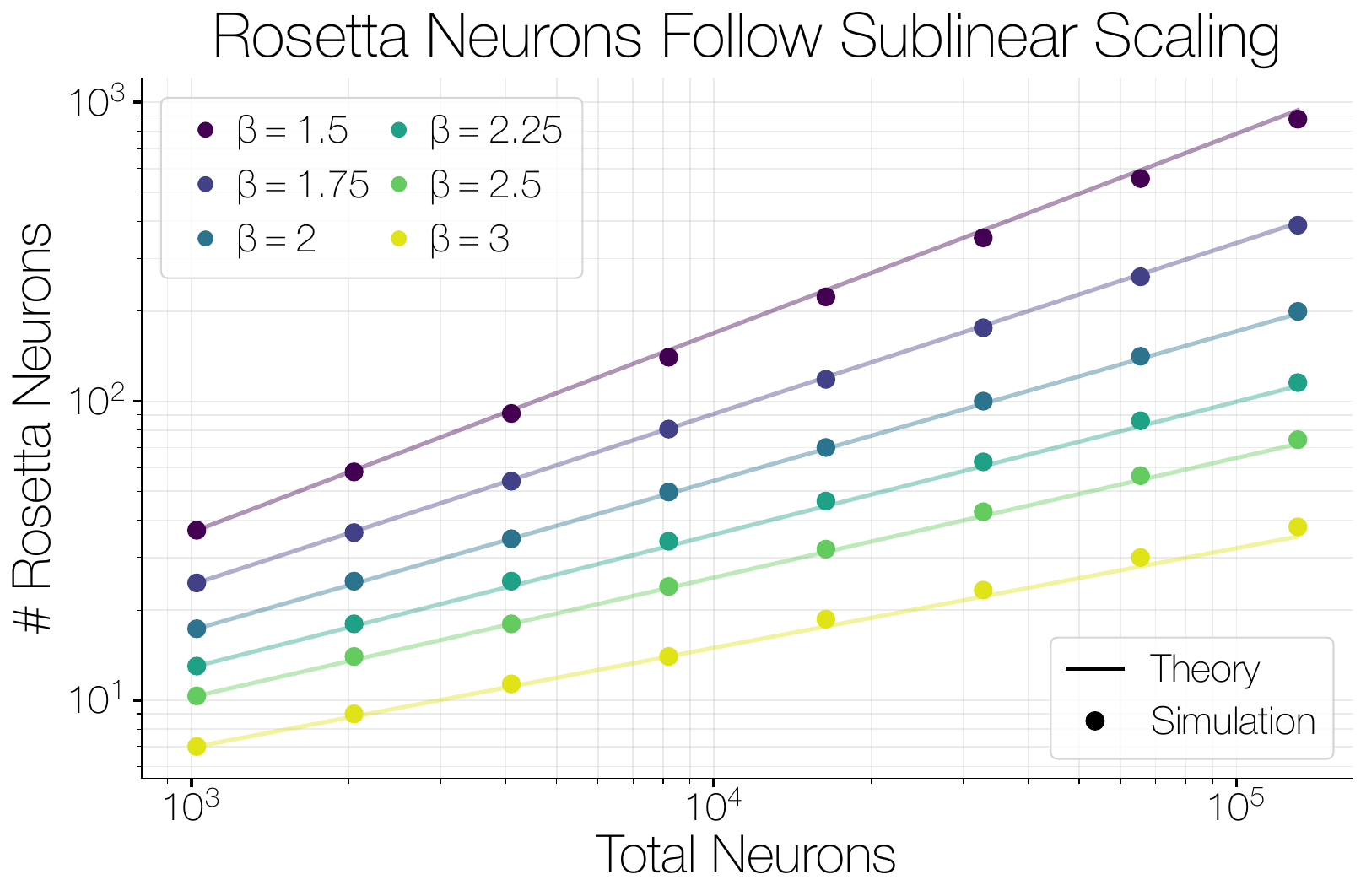}
        \caption{Rosetta Neuron counts scale as \(N^{1/\beta}\).}
        \label{fig:rosetta-count-scaling-simulation}
    \end{subfigure}

    \vspace{1.0em}

    \begin{subfigure}{0.58\textwidth}
        \centering
        \includegraphics[width=\linewidth]{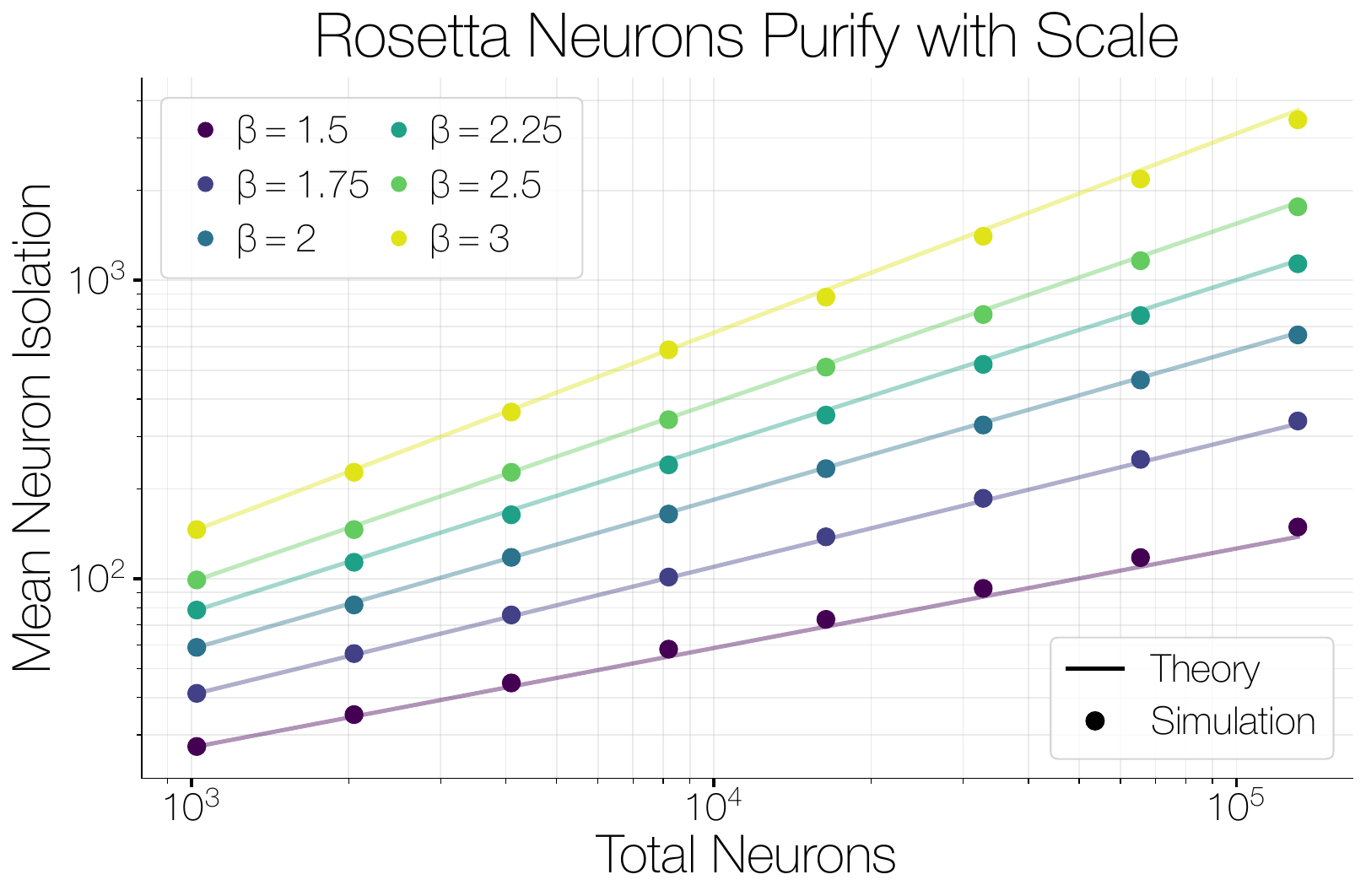}
        \caption{Rosetta Neuron isolation scales as \(N^{(\beta-1)/\beta}\).}
        \label{fig:rosetta-purification-simulation}
    \end{subfigure}

    \vspace{1.0em}

    \begin{subfigure}{0.58\textwidth}
        \centering
        \includegraphics[width=\linewidth]{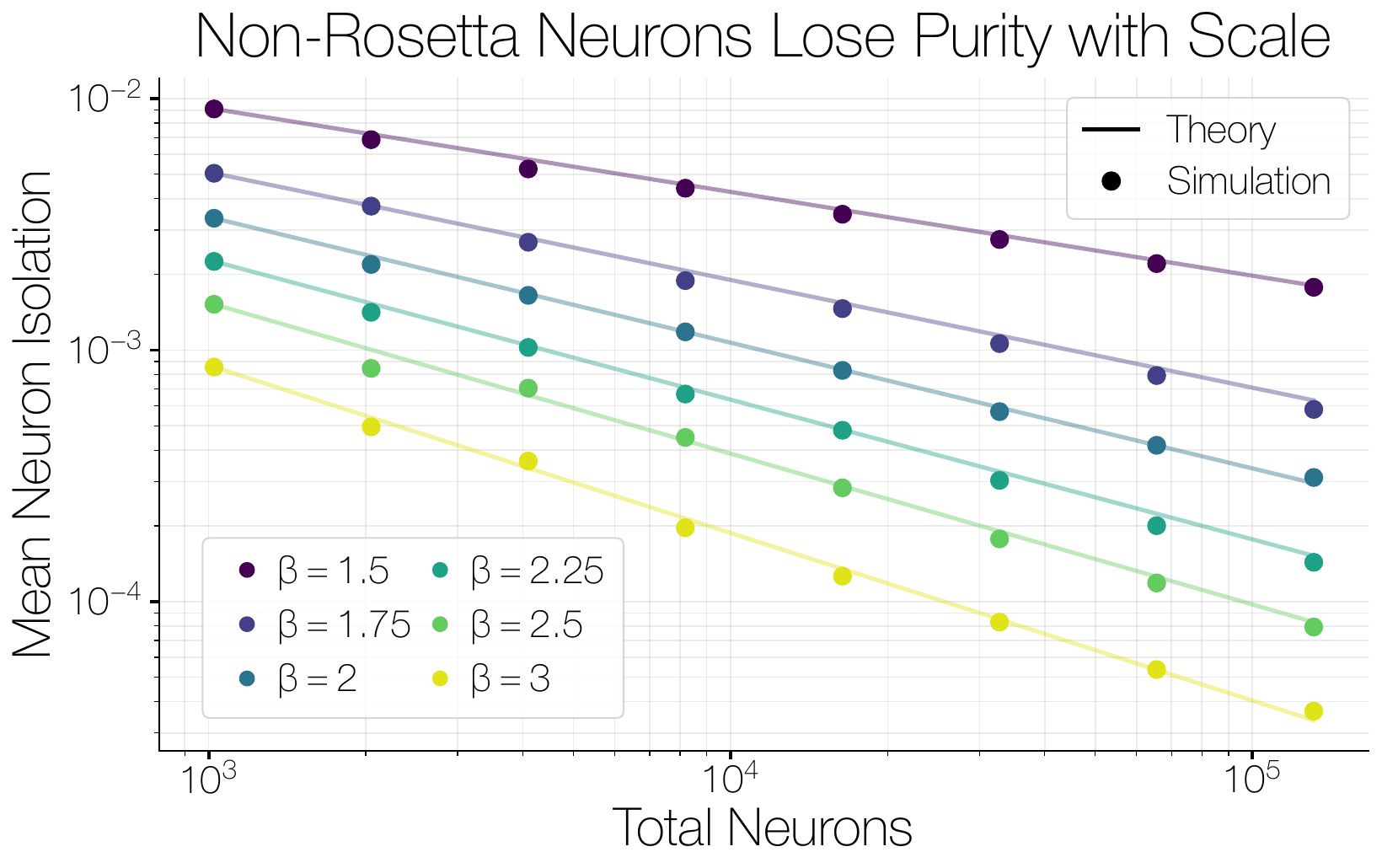}
        \caption{Non-Rosetta neuron isolation scales as \(N^{(1-\beta)/\beta}\).}
        \label{fig:nonrosetta-depurification-simulation}
    \end{subfigure}

    \caption{
    \textbf{Scaling behavior of Rosetta and non-Rosetta neurons in simulation}.
    Top: the number of Rosetta Neurons follows the predicted scaling law according to our analytical model.
    Middle: Rosetta Neurons become more isolated with scale.
    Bottom: Non-Rosetta neurons become less isolated with scale, as predicted by our theory.
    }
    \label{fig:rosetta-scaling-simulations}
\end{figure}

\textbf{Rosetta Neuron matching recovers theoretical predictions.}
The simulations closely match the analytical predictions as seen in~\Cref{fig:rosetta-scaling-simulations}. Across the sweep of feature-spectrum exponents $\beta$, the recovered Rosetta Neuron count follows the predicted scaling $R_\tau(N)=\Theta(N^{1/\beta})$: the fitted empirical slopes align with the theoretical exponents $1/\beta$ (\Cref{fig:rosetta-count-scaling-simulation}). Since the theory predicts the scaling exponent but not the prefactor, we plot theoretical reference curves with slope $1/\beta$ and choose the intercept so that each curve passes through the first empirical point. This agreement suggests that Pearson-correlation mutual-nearest-neighbor matching recovers the theory-predicted Rosetta detectability threshold.

The simulation also recovers the predicted polarization behavior. Our analytical model predicts that the mean isolation of Rosetta neurons grows as $\Theta\!\left(N^{(\beta-1)/\beta}\right)$, while the average clean isolation associated with the non-Rosetta neuron population decays as $O\!\left(N^{(1-\beta)/\beta}\right)$. Consistent with these predictions, Rosetta Neuron matches have increasing assigned-feature isolation with slope close to $(\beta-1)/\beta$ (\Cref{fig:rosetta-purification-simulation}).  The unmatched non-Rosetta neurons have decreasing average isolation with slope close to the theoretical $-(\beta-1)/\beta$ (\Cref{fig:nonrosetta-depurification-simulation}). 

\textbf{Additional Details.} For the reported results, we used $\kappa=1.0$, $\gamma=4.0$, $M=10,000$. We additionally verified that the exponent recovery is robust to these simulation hyperparameters. Across \(\kappa \in \{2^{-1}, 2^{0},\ldots, 2^{5}\}\), \(\gamma \in \{2^{-1}, 2^{0},\ldots, 2^{5}\}\), and \(M \in \{10^{3}, 10^4, 10^5\}\), all the fitted exponents remained within \(\epsilon\approx0.05\) of the corresponding theoretical predictions.
\newpage

\newpage
\section{Additional Results on Rosetta Neuron Properties}
\label{sec:rosetta_properties_supp}

In this section, we provide additional analyses supporting the neuron-level trends described in~\Cref{sec:properties}. We first extend the language-model selectivity results from~\Cref{polsemanticity_section} to additional model families. We then provide further details and results for the document-type firing analysis introduced in~\Cref{specialization_section}. Finally, we analyze where Rosetta Neurons appear across network depth in both language and vision models.

\subsection{Additional Vocabulary-Space Selectivity Results}
\label{app:kurtosis_results}
\begin{figure}[h]
    \centering
    \begin{subfigure}[b]{0.49\textwidth}
        \centering
        \includegraphics[width=\textwidth]{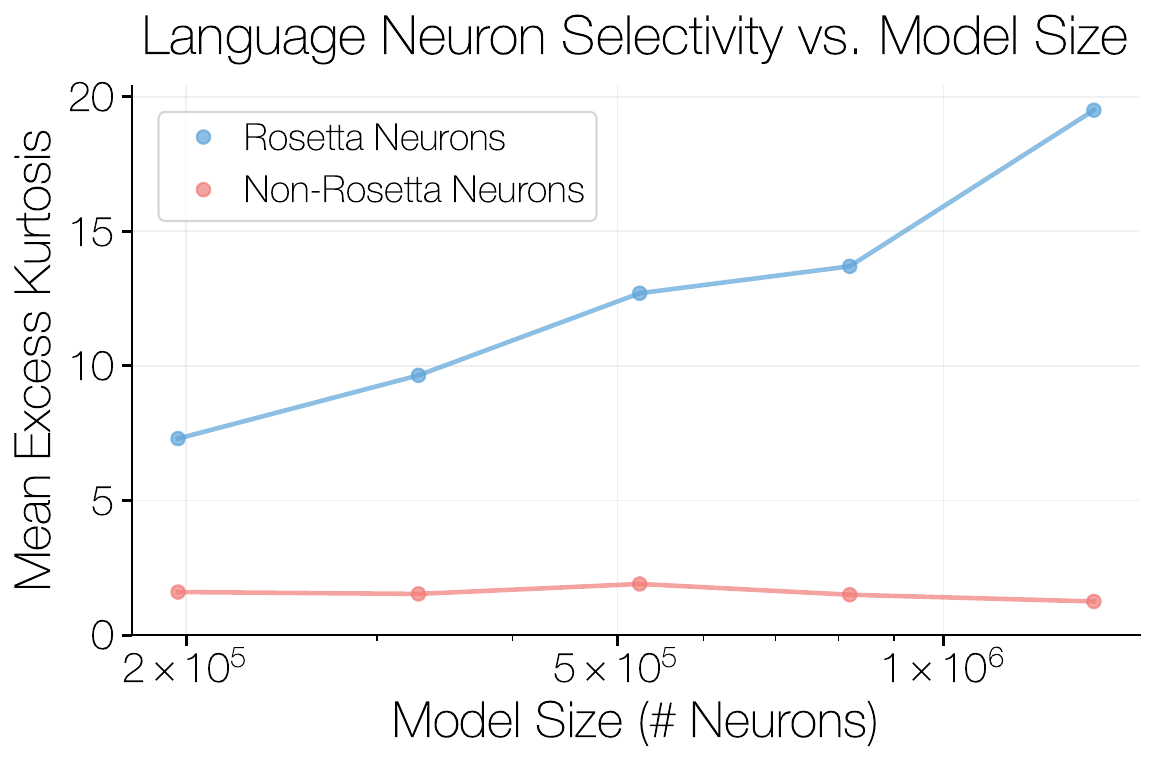}
        \caption{Vocabulary-space neuron selectivity in OPT.}
    \end{subfigure}
    \hfill 
    \begin{subfigure}[b]{0.49\textwidth}
        \centering
        \includegraphics[width=\textwidth]{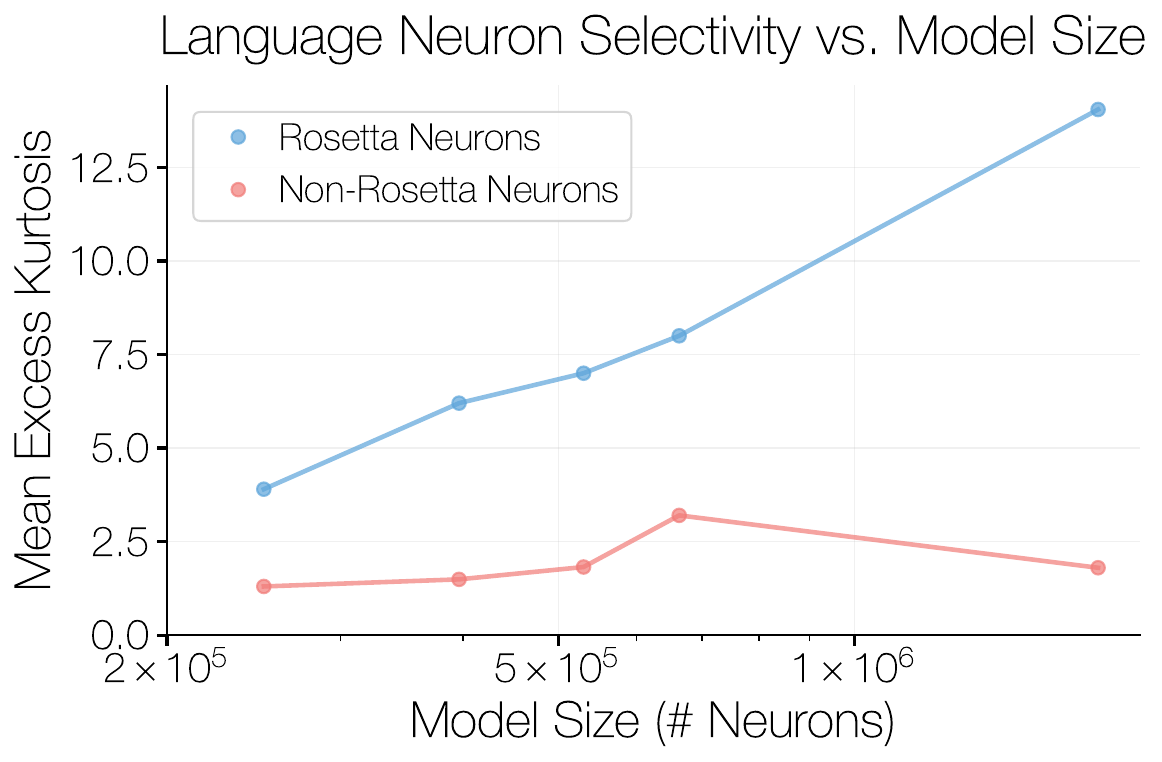}
        \caption{Vocabulary-space neuron selectivity in Qwen2.5.}

    \end{subfigure}
    
    \caption{\textbf{The Neuron Polarization Effect in Language Models.}  Rosetta Neurons exhibit increasing mean excess kurtosis of vocabulary-space projections with scale, suggestive of monosemantic function. Non-Rosetta neurons remain near zero, consistent with weaker vocabulary-level selectivity under this metric.}
    \label{fig:language_selectivity_supp}
\end{figure}

We measure output-side selectivity using the same vocabulary-space excess-kurtosis metric from~\Cref{polsemanticity_section}. As previously shown in~\Cref{fig:neuron_polarization_effect_figure}, a polarization occurs: the mean excess kurtosis of the Rosetta population increases with scale, while non-Rosetta neurons remain near zero, suggesting weak selectivity.  We observe the same qualitative trend for OPT and Qwen-2.5 in~\Cref{fig:language_selectivity_supp}. Across both families, Rosetta Neurons become increasingly selective with scale, while non-Rosetta neurons remain close to the floor. This supports the interpretation that the Neuron Polarization Effect is not tied to a single model family, but reflects a more general trend across language models.

\subsection{Document-Type Firing Analysis}
\label{app:document_firing_details}

We provide additional details and results for the document-type firing analysis in~\Cref{specialization_section}. For each Rosetta Neuron in a given model, we retrieve its top 20 activating contexts from the validation set of The Pile. Each context is assigned the Pile subset label of its source document (e.g., ``GitHub''). We aggregate the original Pile subsets into five general source categories: code, math, formal/scientific text, general prose, and conversational text. The mapping from Pile subsets to categories, together with each category's token share of the validation cache, is shown in~\Cref{tab:pile_category_mapping}.

For each category $c$, we compute the fraction of Rosetta Neuron top activations assigned to that category, and then normalize by the category's token frequency in the validation cache:
\[
\mathrm{NormalizedFire}(c)
=
\frac{
\#\{\text{top activations from category } c\} / \#\{\text{all top activations}\}
}{
\#\{\text{tokens from category } c\} / \#\{\text{all tokens}\}
}.
\]
A value of one corresponds to the corpus baseline, meaning that top activations fall in category $c$ at the same rate as expected from its token frequency. Values above one indicate over-representation among top activations, while values below one indicate under-representation.

\begin{figure}[h]
    \centering

    \begin{subfigure}{0.95\textwidth}
        \centering
        \includegraphics[width=\textwidth]{figures/pythia_rosetta_firing_type.pdf}
        \caption{Pythia Rosetta Neurons.}
        \label{fig:pythia_rosetta_firing}
    \end{subfigure}

    \vspace{0.75em}

    \begin{subfigure}{0.95\textwidth}
        \centering
        \includegraphics[width=\textwidth]{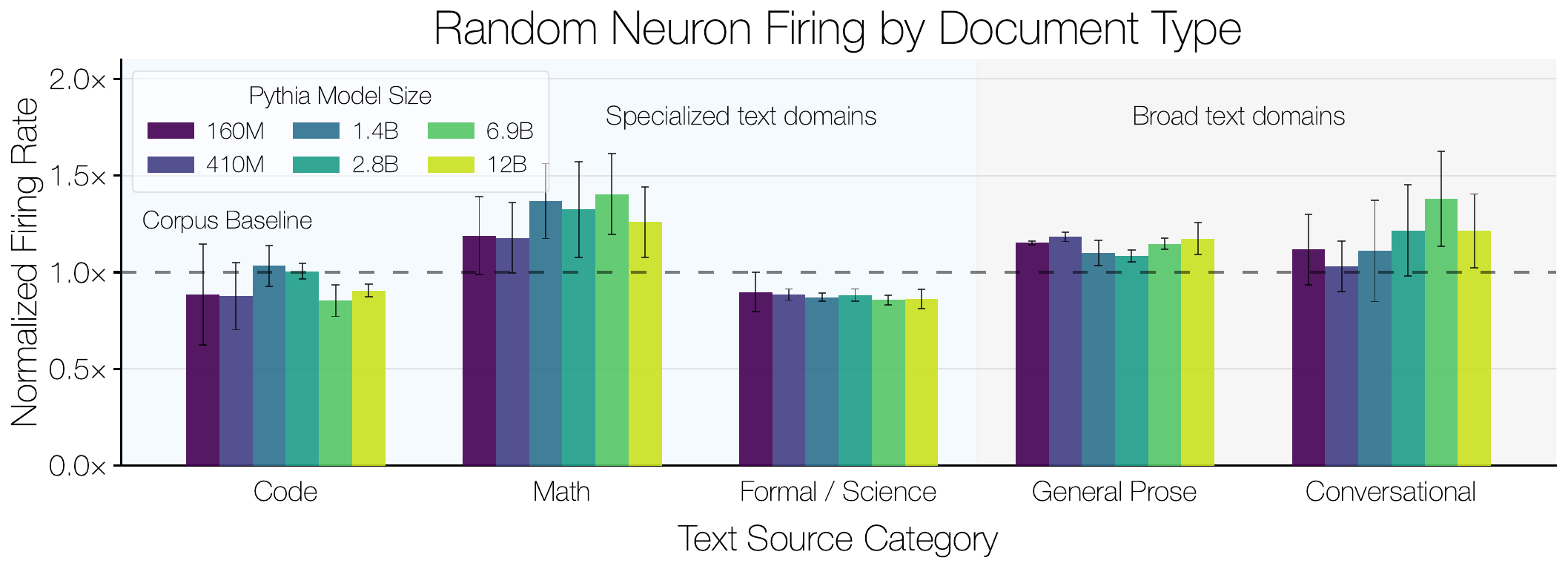}
        \caption{Pythia random non-Rosetta neurons.}
        \label{fig:pythia_random_firing}
    \end{subfigure}

   \caption{\textbf{Rosetta Neuron document-type firing in Pythia.}
For each Pythia model size, each bar shows how often top-activating Rosetta Neuron contexts fall into a document category, normalized by that category's token frequency in the validation cache. The dashed line marks the corpus baseline. With scale, Rosetta Neuron firing shifts toward specialized categories such as code and math.}
    \label{fig:pythia_document_firing_supp}
\end{figure}

\begin{figure}[h]
    \centering

    \begin{subfigure}{0.95\textwidth}
        \centering
        \includegraphics[width=\textwidth]{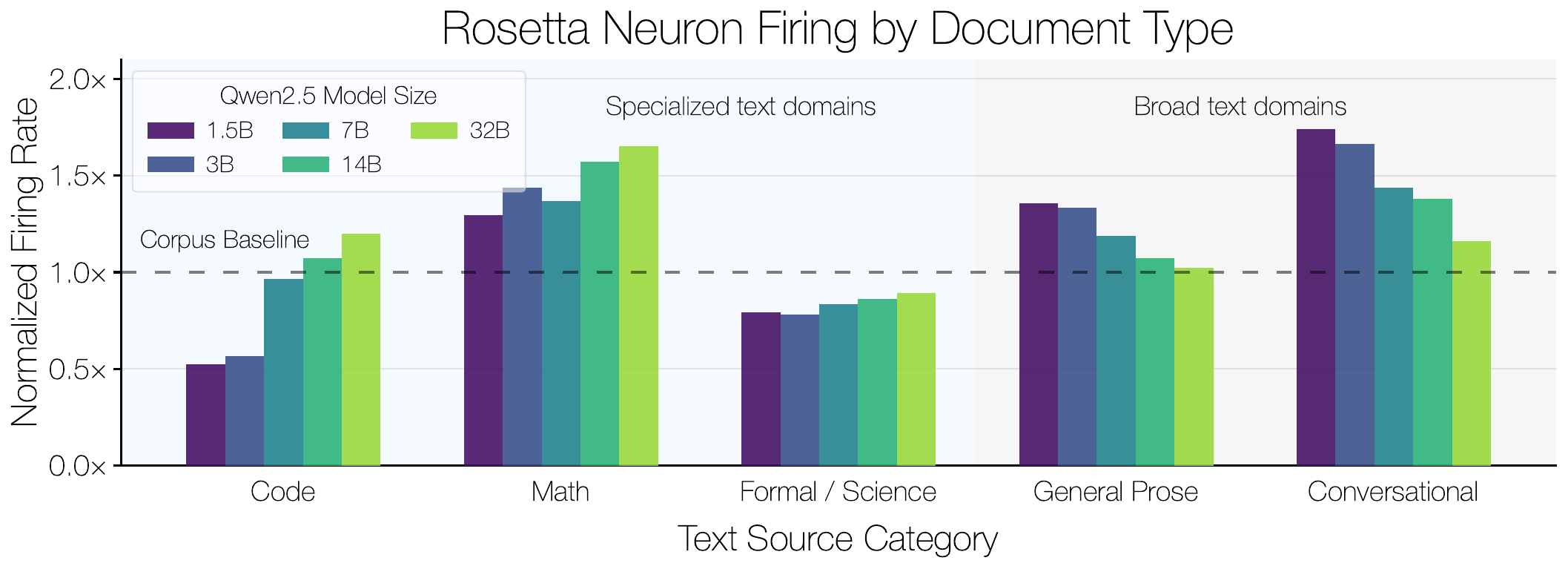}
        \caption{Qwen2.5 Rosetta Neurons.}
        \label{fig:qwen_rosetta_firing}
    \end{subfigure}

    \vspace{0.75em}

    \begin{subfigure}{0.95\textwidth}
        \centering
        \includegraphics[width=\textwidth]{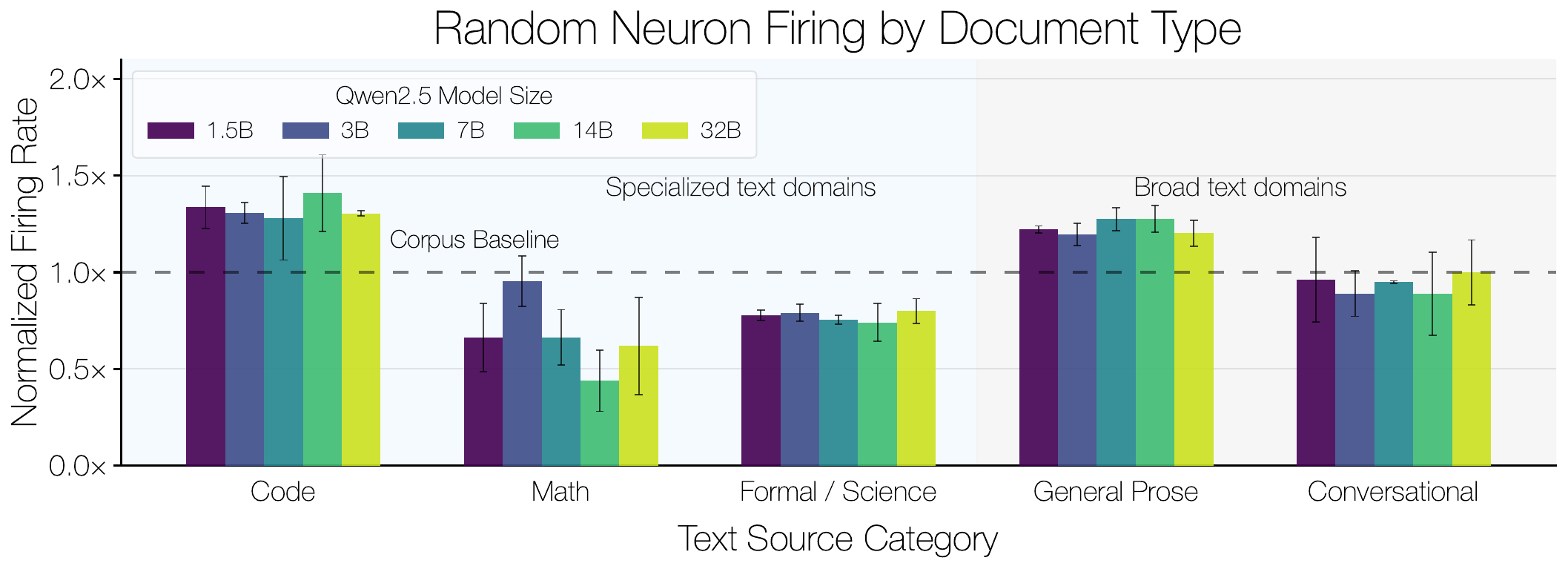}
        \caption{Qwen2.5 random non-Rosetta neurons.}
        \label{fig:qwen_random_firing}
    \end{subfigure}

    \caption{\textbf{Document-type firing in Qwen2.5.}
    We use the same normalized document-type firing statistic as in~\Cref{fig:pythia_document_firing_supp}. Rosetta Neurons show an increasing shift toward specialized categories such as code and math with scale. Random non-Rosetta neurons may exhibit category-specific biases, but do not exhibit the same consistent scale-dependent shift toward specialized domains.}
    \label{fig:qwen_document_firing_supp}
\end{figure}

As a control, for each model family and scale, we randomly sample 1000 non-Rosetta neurons from the corresponding model and compute the same normalized document-type firing statistic. We repeat this procedure three times and report the mean and 95\% confidence interval across random non-Rosetta neuron samples. This baseline tests whether the observed document-type trends reflect a general property of neurons at a given scale, rather than a property specific to the Rosetta population. We apply this analysis on Pythia and Qwen-2.5 and report the results in~\Cref{fig:pythia_document_firing_supp,fig:qwen_document_firing_supp}. Rosetta Neuron firing becomes increasingly concentrated on specialized categories such as code and math as model size increases for both model families. In contrast, random neurons may exhibit category-specific biases, but they do not show the same systematic shift in firing preference; their category-level firing patterns remain comparatively stable across scale. Since the models within each family are trained on the same data mixture, this shift is unlikely to be explained solely by changing exposure to specialized documents. It suggests that scale may affect which domain-specific features are represented among Rosetta Neurons.

\begin{table}[t]
    \centering
    \small
    \setlength{\tabcolsep}{4pt}
    \begin{tabularx}{\textwidth}{@{}l c X@{}}
        \toprule
        \textbf{Category} & \textbf{Corpus Share} & \textbf{Pile Subsets} \\
        \midrule
        Code & 0.1195 & GitHub \\
        Math & 0.0317 & DM Mathematics \\
        Formal/scientific & 0.4644 & ArXiv, PubMed Central, PubMed Abstracts, FreeLaw, USPTO Backgrounds, NIH ExPorter, StackExchange, PhilPapers \\
        General prose & 0.3430 & Pile-CC, OpenWebText2, Wikipedia (en), Books3, Gutenberg (PG-19), EuroParl, BookCorpus2 \\
        Conversational & 0.0415 & OpenSubtitles, Ubuntu IRC, HackerNews, Enron Emails, YouTubeSubtitles \\
        \bottomrule
    \end{tabularx}
    \vspace{0.2em}
    \caption{\textbf{Pile source categories used for document-type firing analysis.}
    We aggregate the 22 Pile subsets into five coarse source categories. The corpus share denotes each category's token frequency in the validation cache and is used to normalize top-activation frequencies.}
    \label{tab:pile_category_mapping}
\end{table}

\subsection{Depth-Wise Distribution of Rosetta Neurons}
\label{app:depthwise_rosetta}

We analyze where Rosetta Neurons appear across network depth as models scale. For language models, we use Rosetta Neurons discovered from the Pythia--OPT matching runs. For vision models, we use Rosetta Neurons discovered from the Diffusion--OpenCLIP matching runs. Because models at different scales have different numbers of layers, we compare layers using normalized depth. For each model, we map layer index $\ell$ to normalized depth $\ell/(L-1)$, where $L$ is the total number of transformer blocks, and divide the interval $[0,1]$ into 12 equally spaced bins. We then compute the fraction of discovered Rosetta Neurons that fall into each depth bin.

\begin{figure}
    \centering
    \includegraphics[width=0.9\linewidth]{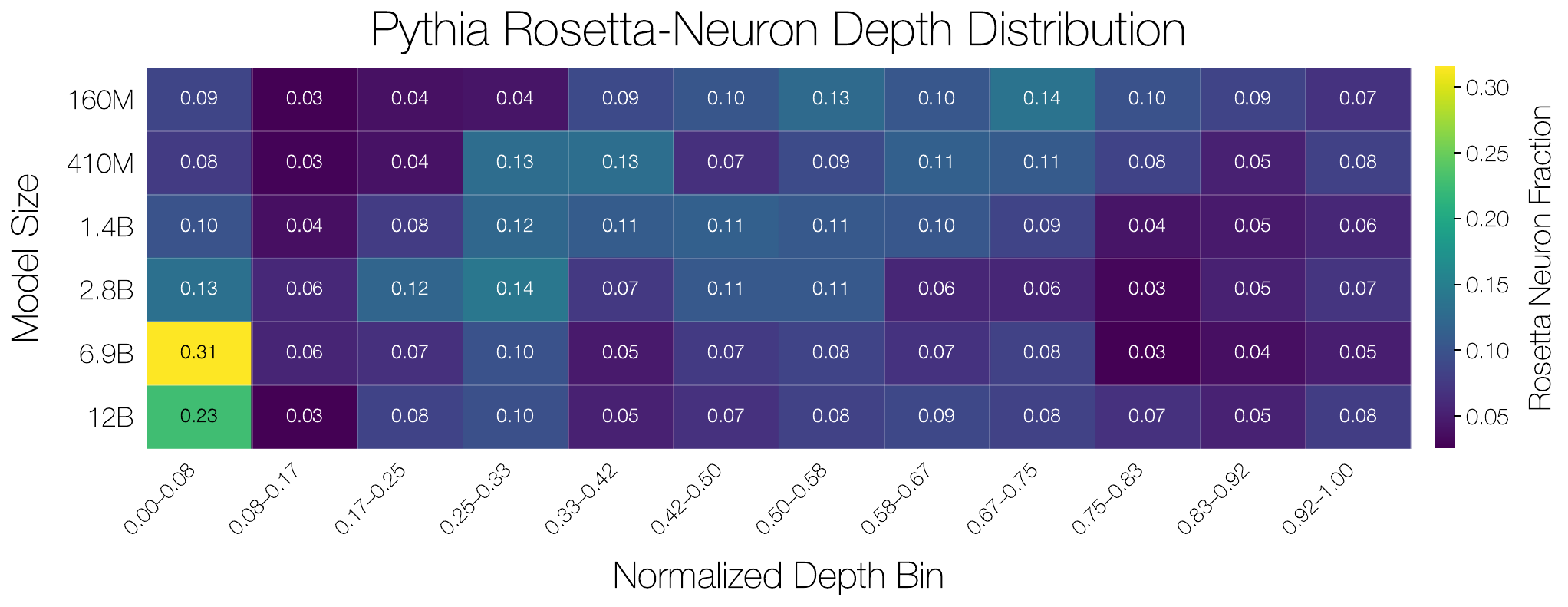}
    \caption{\textbf{Depth-wise distribution of Rosetta Neurons in Pythia across scale.} Rosetta Neurons discovered from the Pythia--OPT matching runs.}
    \label{fig:pythia_depth_distribution}
\end{figure}

\begin{figure}
    \centering
    \includegraphics[width=0.9\linewidth]{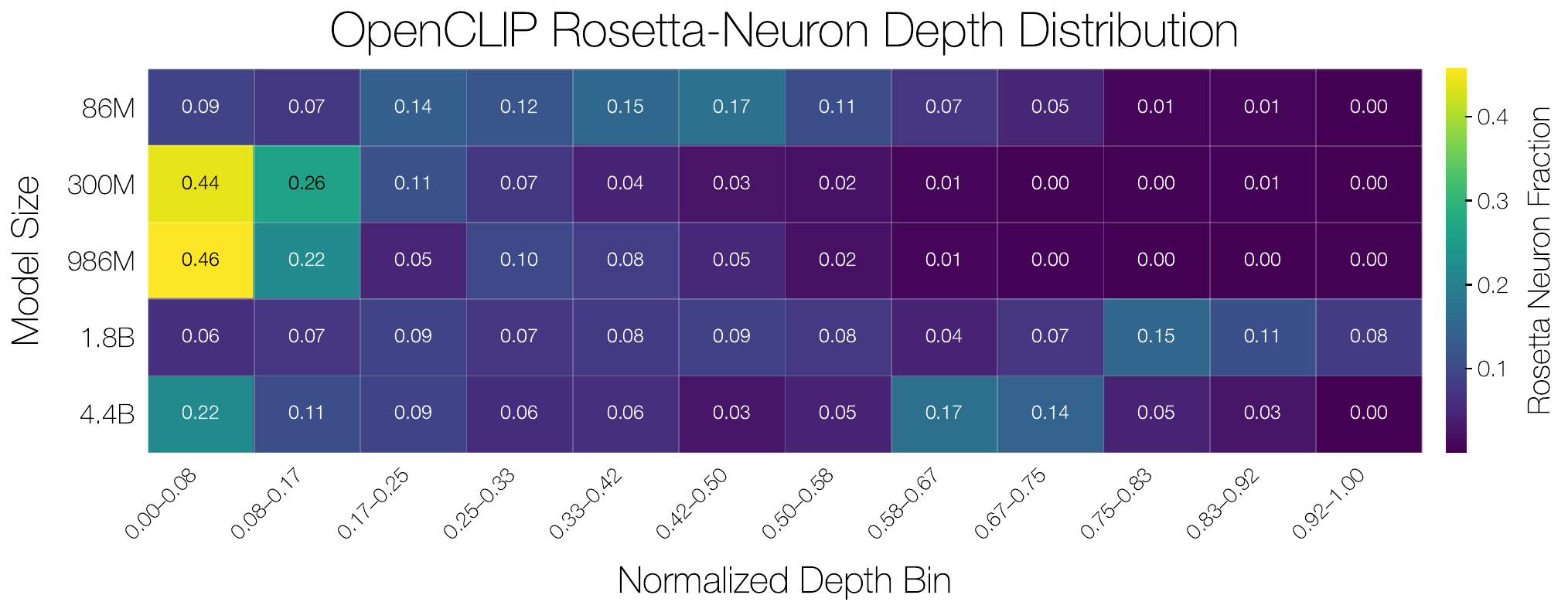}
    \caption{\textbf{Depth-wise distribution of Rosetta Neurons in OpenCLIP across scale.} Rosetta Neurons discovered from the Diffusion--OpenCLIP matching runs.}
    \label{fig:clip_depth_distribution}
\end{figure}

As shown in~\Cref{fig:pythia_depth_distribution,fig:clip_depth_distribution}, Rosetta Neurons are distributed across multiple depth bins in both Pythia and OpenCLIP, rather than being confined to a single depth range across scale. The distribution is not uniform: some models show stronger concentration in early layers, such as Pythia-6.9B in language and OpenCLIP ViT-L/14 (300M) in vision. Overall, however, the depth-wise analysis does not reveal a single consistent depth profile shared across all scales and modalities. We therefore interpret this analysis primarily as a non-degeneracy check: the Rosetta Neuron population is not explained solely by matches from one fixed layer region, and the observed scaling trends are not driven by a single depth profile.

\newpage
\section{Data Filtering Experimental Details}
\label{sec:data_filter_supp}

\textbf{Data.} We use CodeSearchNet~\citep{husain2019codesearchnet}, a function-level code corpus extracted from publicly available GitHub repositories. This dataset spans six programming languages: Python, JavaScript, Java, Go, Ruby, PHP. Each example in the dataset is a single parsed function from a real-world repository. We use JavaScript as a representative target domain. Its training split contains $58{,}025$ functions, corresponding to roughly 16M tokens under the GPT-2 tokenizer, and accounts for approximately 6\% of the total tokens in the full multilingual training set. The corresponding JavaScript test split contains $3291$ functions and 1.11M tokens.

\textbf{Filtering.} Each filtering method selects functions from the full multilingual CodeSearchNet training set under a matched-token setup. The target budget is set to the size of the JavaScript training split, approximately 16M GPT-2 tokens. We compare four filtering methods: a Rosetta Neuron filter, a non-Rosetta neuron filter, an i.i.d. random sample from the full multilingual training set, and the oracle JavaScript training split. For the neuron-based filters, we score each function by the selected neuron's mean activation over tokens in the function, and select the highest-scoring functions until reaching the matched token budget.

The neurons used for filtering are selected prior to the CodeSearchNet experiment using a different data distribution. We identify candidate JavaScript-selective neurons in Pythia-6.9B using the LLM-based neuron labeling procedure from~\Cref{sec:qualitative_supp}, which infers neuron function from top-activating contexts in The Pile. Among Rosetta Neurons whose top-activating Pile contexts are annotated as JavaScript-related, we select the single neuron with the highest mean normalized activation over its top-20 Pile contexts. We apply the same rule to non-Rosetta neurons in the same layer whose top activations are annotated as JavaScript-related. This selection is performed using only The Pile activations and LLM annotations before any CodeSearchNet filtering or downstream evaluation. To evaluate domain recovery, we compute F1 between the selected training functions and the oracle JavaScript training split (\Cref{tab:js_filtering}). Since all methods use the same token budget, precision and recall are nearly identical, so F1 compactly summarizes ground truth recovery. 

We present qualitative examples of dataset recovery below. The Rosetta Neuron, Layer 16 Unit 11168 in Pythia-6.9B, retrieves JavaScript functions with high selectivity. In contrast, the non-Rosetta neuron, Layer 16 Unit 12873, shows weaker JavaScript selectivity and fires on a broader range of code, especially web-oriented languages such as PHP.

\textbf{Training details.} To evaluate the downstream utility of the filters, we conduct continued pretraining of GPT2-1.5B using rank-1 LoRA in the attention and MLP layers~\citep{hu2022lora}. All runs use AdamW~\citep{loshchilov2018decoupled} with peak learning rate $10^{-3}$ and a 3\% linear warmup followed by cosine decay to 0. We train for one epoch at sequence length 1024 using an effective batch size of 32 sequences. All filtering variants use the same model, hyperparameters, token budget, and number of optimizer steps; only the composition of the training data varies. We report perplexity in~\Cref{tab:js_filtering} with 95\% confidence intervals calculated over three runs.
\smallskip
\begin{tcolorbox}[
  title={\centering\textbf{Rosetta Neuron-Filtered Function \#1:} Label=JavaScript},
  colback=white, colframe=black, coltitle=black,
  colbacktitle=gray!15, fonttitle=\normalsize,
  boxrule=0.4pt, arc=0pt, left=4pt, right=4pt, top=2pt, bottom=2pt,
]
\begin{Verbatim}[fontsize=\scriptsize]
function (firstStart) {
    this._startTime = (firstStart) ? Aria._start : new Date();
    if (firstStart) {
        this._logs = [{
            classpath : "Aria",
            msg : "Framework initialization",
            start : Aria._start
        }, {
            classpath : "Aria",
            stop : (new Date()).getTime()
        }];
        this._nbLogs = 2;
    } else {
        this._logs = [];
        this._nbLogs = 0;
    }

    // map function on JsObject prototype
    ariaCoreJsObject.prototype.$logTimestamp = function (msg, classpath) {
        classpath = classpath ? classpath : this.$classpath;
        aria.utils.Profiling.logTimestamp(classpath, msg);
    };
    ariaCoreJsObject.prototype.$startMeasure = function (msg, classpath) {
        classpath = classpath ? classpath : this.$classpath;
        return aria.utils.Profiling.startMeasure(classpath, msg);
    };
    ariaCoreJsObject.prototype.$stopMeasure = function (id, classpath) {
        classpath = classpath ? classpath : this.$classpath;
        aria.utils.Profiling.stopMeasure(classpath, id);
    };
}
\end{Verbatim}
\end{tcolorbox}

\smallskip
\begin{tcolorbox}[
  title={\centering\textbf{Rosetta Neuron-Filtered Function \#2:} Label=JavaScript},
  colback=white, colframe=black, coltitle=black,
  colbacktitle=gray!15, fonttitle=\normalsize,
  boxrule=0.4pt, arc=0pt, left=4pt, right=4pt, top=2pt, bottom=2pt,
]
\begin{Verbatim}[fontsize=\scriptsize]
function(fn, mOptions) {
    // Functionality taken from lodash open source library and adapted as needed

    mOptions = Object.assign({
        wait: 0,
        leading: true
    }, mOptions);
    mOptions.maxWait = mOptions.wait;
    mOptions.trailing = true;
    mOptions.requestAnimationFrame = false;

    return TableUtils.debounce(fn, mOptions);
}
\end{Verbatim}
\end{tcolorbox}
\smallskip
\smallskip
\begin{tcolorbox}[
  title={\centering\textbf{Rosetta Neuron-Filtered Function \#3:} Label=JavaScript},
  colback=white, colframe=black, coltitle=black,
  colbacktitle=gray!15, fonttitle=\normalsize,
  boxrule=0.4pt, arc=0pt, left=4pt, right=4pt, top=2pt, bottom=2pt,
]
\begin{Verbatim}[fontsize=\scriptsize]
function forEach(f, arr) {
  if (NATIVE_ARRAY_FOREACH) {
    if (arr) {
      NATIVE_ARRAY_FOREACH.call(arr, f);
    }

    return;
  }

  for (var i = 0; i < arr.length; i++) {
    f(arr[i], i, arr);
  }
}
\end{Verbatim}
\end{tcolorbox}
\smallskip
\smallskip
\smallskip
\smallskip
\smallskip
\smallskip
\begin{tcolorbox}[
  title={\centering\textbf{Non-Rosetta Neuron-Filtered Function \#1:} Label=PHP},
  colback=white, colframe=black, coltitle=black,
  colbacktitle=gray!15, fonttitle=\normalsize,
  boxrule=0.4pt, arc=0pt, left=4pt, right=4pt, top=2pt, bottom=2pt,
]
\begin{Verbatim}[fontsize=\footnotesize]
protected static function parseSequence($sequence, &$i = 0)
  {
    $output = array();
    $len = strlen($sequence);
    $i += 1;

    // [foo, bar, ...]
    while ($i < $len) {
      switch ($sequence[$i]) {
        case '[':
          // nested sequence
          $output[] = self::parseSequence($sequence, $i);
          break;
        case '{':
          // nested mapping
          $output[] = self::parseMapping($sequence, $i);
          break;
        case ']':
          return $output;
        case ',':
        case ' ':
          break;
        default:
          $isQuoted = in_array($sequence[$i], array('"', "'"));
          $value = self::parseScalar($sequence, array(',', ']'), array('"', "'"), $i);

          if (!$isQuoted && false !== strpos($value, ': ')) {
            // embedded mapping?
            try {
              $value = self::parseMapping('{'.$value.'}');
            } catch (InvalidArgumentException $e) {
              // no, it's not
            }
          }

          $output[] = $value;

          --$i;
      }

      ++$i;
    }

    throw new InvalidArgumentException(sprintf('Malformed inline YAML string %s', $sequence));
  }
\end{Verbatim}
\end{tcolorbox}

\begin{tcolorbox}[
  title={\centering\textbf{Non-Rosetta Neuron-Filtered Function \#2:} Label=PHP},
  colback=white, colframe=black, coltitle=black,
  colbacktitle=gray!15, fonttitle=\normalfont,
  boxrule=0.4pt, arc=0pt, left=4pt, right=4pt, top=2pt, bottom=2pt,
]
\begin{Verbatim}[fontsize=\scriptsize]
private function parsePhpValue($key, $value, array &$result)
    {
        $node =& $result;
        $keyBuffer = '';

        for ($i = 0, $t = strlen($key); $i < $t; $i++) {
            switch ($key[$i]) {
                case '[':
                    if ($keyBuffer) {
                        $this->prepareNode($node, $keyBuffer);
                        $node =& $node[$keyBuffer];
                        $keyBuffer = '';
                    }
                    break;
                case ']':
                    $k = $this->cleanKey($node, $keyBuffer);
                    $this->prepareNode($node, $k);
                    $node =& $node[$k];
                    $keyBuffer = '';
                    break;
                default:
                    $keyBuffer .= $key[$i];
                    break;
            }
        }

        if (isset($node)) {
            $this->duplicates = true;
            $node[] = $value;
        } else {
            $node = $value;
        }
    }
\end{Verbatim}
\end{tcolorbox}
\smallskip
\smallskip
\begin{tcolorbox}[
  title={\centering\textbf{Non-Rosetta Neuron-Filtered Function \#3:} Label=JavaScript},
  colback=white, colframe=black, coltitle=black,
  colbacktitle=gray!15, fonttitle=\normalfont,
  boxrule=0.4pt, arc=0pt, left=4pt, right=4pt, top=2pt, bottom=2pt,
]
\begin{Verbatim}[fontsize=\scriptsize,breaklines,breakanywhere]
function generateUniqueKey(index, initialKey) {
  var currentCandidate = initialKey;

  var counter = 0;
  while (index[currentCandidate]) {
    var numberAtEndOfKeyMatches = currentCandidate.match(
      NUMBER_AT_END_OF_KEY_REGEX
    );
    if (numberAtEndOfKeyMatches !== null) {
      var nextNumber = parseInt(numberAtEndOfKeyMatches[1], 10) + 1;

      currentCandidate = currentCandidate.replace(
        NUMBER_AT_END_OF_KEY_REGEX,
        "(" + nextNumber + ")"
      );
    } else {
      currentCandidate += " (1)";
    }

    // This loop should always find something eventually, but because it's a bit dangerous looping endlessly...
    counter++;
    if (counter >= 100000) {
      throw new DeveloperError(
        "Was not able to find a unique key for " +
          initialKey +
          " after 100000 iterations." +
          " This is probably because the regex for matching keys was somehow unable to work for that key."
      );
    }
  }

  return currentCandidate;
}
\end{Verbatim}
\end{tcolorbox}

\newpage
\clearpage
\section{Dataset Ablations}
\label{sec:dataset_ablations}

\subsection{Ablation on the Number of Tokens Used for Language Model Matching}

\begin{figure}
    \centering
    \includegraphics[width=1.0\linewidth]{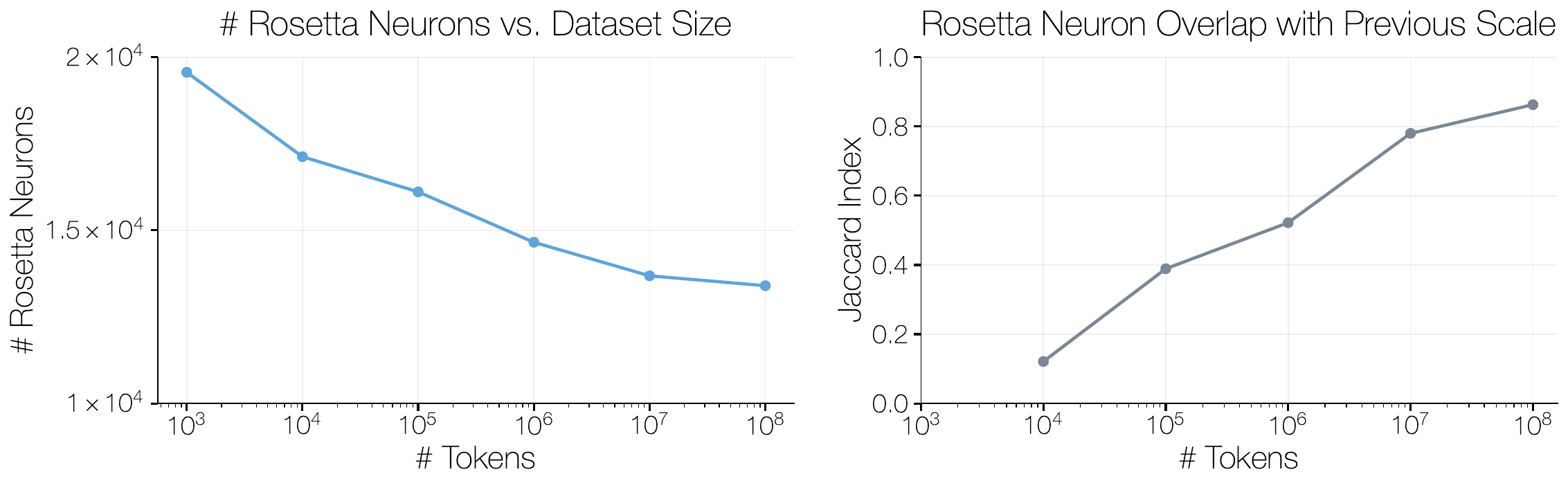}
    \caption{\textbf{Language matching stability as a function of dataset size.} We vary the number of tokens used to match neurons between Pythia-6.9B and OPT-6.7B. Left: number of Rosetta Neurons discovered at each data scale. Right: overlap with the Rosetta Neuron set from the previous data scale. The discovered Rosetta Neuron set becomes increasingly stable as the token budget grows.}
    \label{fig:lm_token_scaling}
\end{figure}

We ablate the number of tokens for matching neurons between language models, taking the Pythia-6.9B--OPT-6.7B model pair as a representative run. Specifically, we sample i.i.d. token sequences from the validation split of The Pile and apply the procedure described in~\Cref{sec:rosetta-methods} to identify Rosetta Neurons. We sweep the total number of tokens used for matching from $10^3$ to $10^8$. At each data scale, we identify the set of Rosetta Neurons and report the total count in~\Cref{fig:lm_token_scaling} (left). We also compute the overlap between the Rosetta Neuron set identified at each scale and the set identified at the immediately preceding scale, shown in~\Cref{fig:lm_token_scaling} (right). We find that the discovered Rosetta Neuron set becomes increasingly stable as the number of tokens used for matching increases. We use $10$M tokens for the main language-model scaling runs, as this provides a practical balance between stability and computational cost.

\subsection{Ablation on the Number of Images Used for Vision Model Matching}
\begin{figure}
    \centering
    \includegraphics[width=1.0\linewidth]{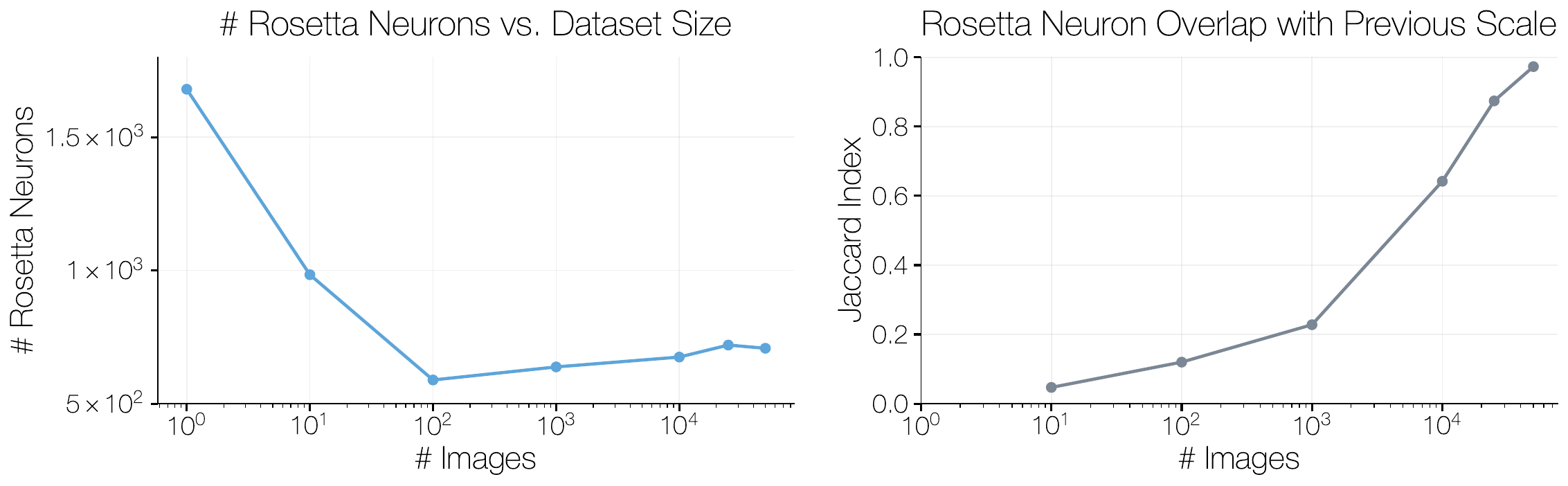}
    \caption{\textbf{Diffusion-to-discriminative matching stability as a function of dataset size.} We vary the number of generated images used to match neurons between pMF DiT-B/16 and OpenCLIP ViT-B/16. Left: number of Rosetta Neurons discovered at each data scale. Right: overlap with the Rosetta Neuron set from the previous data scale. Stability improves as the number of images used for neuron matching approaches 50,000.}
    \label{fig:diffusion_data_scaling}
\end{figure}

We ablate the number of images used to match neurons between a diffusion model and a discriminative model. Specifically, we take pMF DiT-B/16 and OpenCLIP ViT-B/16 as a representative pairing and follow the procedure described in~\Cref{sec:rosetta-methods} to identify Rosetta Neurons. We sweep the number of images generated by the diffusion model from 1 to 50{,}000. At each data scale, we identify the set of Rosetta Neurons and report the total count in~\Cref{fig:diffusion_data_scaling} (left). We also measure the overlap between the set of Rosetta Neurons identified at each data scale and the set identified at the immediately preceding data scale, shown in~\Cref{fig:diffusion_data_scaling} (right). We find that the discovered set of Rosetta Neurons stabilizes by 50,000 images: the total count plateaus, and the overlap between successive data scales approaches 1.

\subsection{Ablation on the Image Distribution Used for Vision Model Matching}

\begin{figure}
    \centering
    \includegraphics[width=1.0\linewidth]{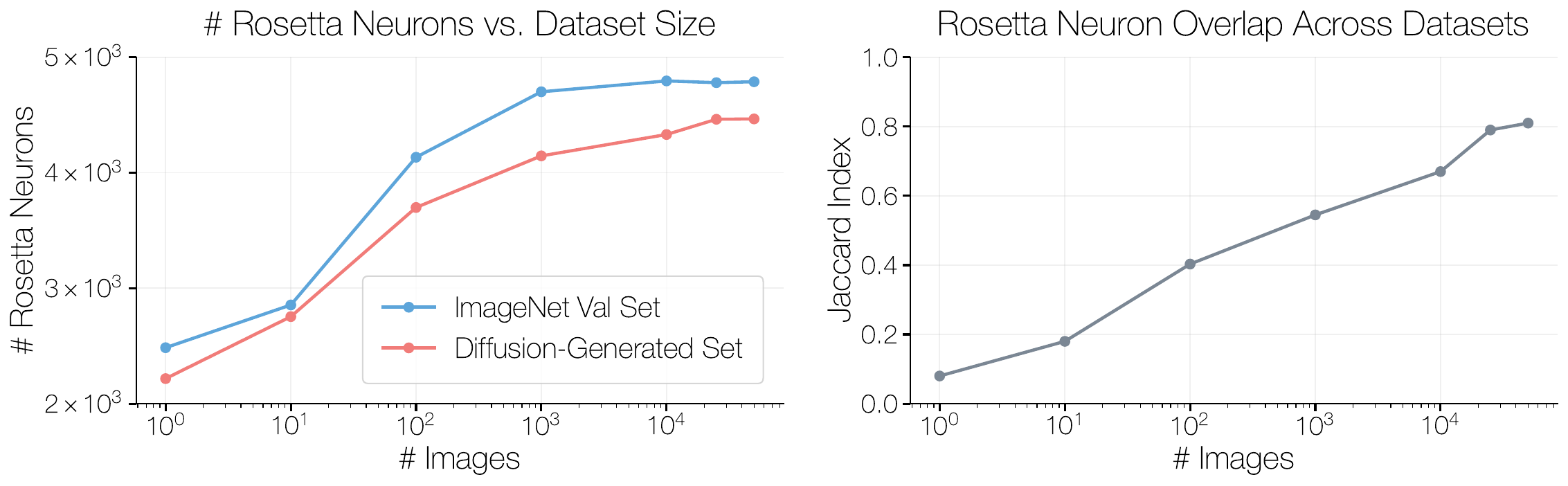}
    \caption{\textbf{Effect of image distribution on vision model matching.} We compare Rosetta Neuron matching between OpenCLIP ViT-B/16 and DINOv2 ViT-B/14 using real and diffusion-generated images. Left: number of Rosetta Neurons identified at each data scale. Right: Jaccard index between the Rosetta Neuron sets obtained from the two image distributions. At larger data scales, the two distributions yield similar numbers of matches and substantial overlap.
    }
    \label{fig:image_distribution_ablation}
\end{figure}

Our vision experiments match neurons between a generative model and a discriminative model, following the GAN-based setup of~\citep{dravid2023rosetta}. For modern diffusion-based generators, this requires generated images, since activations from the generative model are only available along the generation trajectory. A natural concern is that this introduces a distribution shift for the discriminative model, which may affect which Rosetta Neurons are identified. As a proxy for this distribution-shift concern, we compare matching under real and generated image distributions using two discriminative models, OpenCLIP ViT-B/16 and DINOv2 ViT-B/14. We run the Rosetta Neurons procedure at data scales from 1 to 50{,}000 images, once with real images and once with diffusion-generated images. As shown in~\Cref{fig:image_distribution_ablation}, at larger data scales of 25{,}000 to 50{,}000 images, generated images recover approximately $93\%$ of the Rosetta Neuron count found with real images, with an overlap of roughly 0.8. Matching on generated images recovers most of the same correspondences, suggesting that distribution shift has a modest effect. 

\newpage

\newpage
\clearpage
\section{Additional Details on VLM-as-a-Judge}
\label{sec:vlm_appendix}
This section provides additional details on our VLM-as-a-judge setup for measuring neuron selectivity in vision models. We examine sensitivity to the number of top-activating examples shown to the VLM, validate the approach with a held-out prediction task, and finally show that the same qualitative selectivity trends hold for diffusion models and DINOv2.

\subsection{Detailed Experimental Setup}
\label{sec:detailed_vlm_setup}
\textbf{Collecting activations and top-$k$ images for Rosetta neurons.}
Given a generative--discriminative model pair (e.g., DiT-B/16 and OpenCLIP ViT-B/16), we generate 50{,}000 images with the generative model and pass them through the discriminative model. We randomly sample 100 Rosetta Neuron pairs and evaluate each neuron in the pair independently. For the discriminative model, we record each sampled neuron’s patch-level activations over the 50{,}000 images. For each neuron and image, we compute a scalar activation score by averaging the neuron’s activation over all spatial patches. We rank images by this score and retrieve the top 20 activating images for VLM evaluation. We apply the same procedure to the corresponding generative-model neurons, using activations cached during image generation.

\textbf{Collecting activations and top-$k$ images for random neurons.}
For the non-Rosetta neuron baseline, we apply the same procedure independently within each model rather than using matched neuron pairs. In each model, we randomly sample 100 non-Rosetta neurons and record their patch-level activations over the 50{,}000 images. For each neuron and image, we compute a scalar activation score by averaging the neuron’s activation over all spatial patches. We rank images by this score and retrieve the top 20 activating images for VLM evaluation.

\textbf{VLM judging.} For each selected Rosetta Neuron, we form a grid of the top 20 activating images together with the activation heatmaps from one model. The composite image is provided to GPT-5.4 along with a prompt asking it to determine whether the neuron responds to a single coherent visual feature. We provide this prompt below. For each neuron, the VLM returns a binary monosemanticity judgment together with a short natural-language description of the inferred feature or set of features. We follow a similar procedure for the non-Rosetta neuron baseline. We report results for both Rosetta Neurons and the baseline in a model over five independent trials. For the Rosetta Neuron analysis, each trial consists of a disjoint random subset of 100 Rosetta Neurons. For the non-Rosetta baseline, each trial consists of disjoint random subsets of 100 neurons from the non-Rosetta set. We calculate 95\% confidence intervals for the percentage identified as monosemantic in each setting.

\smallskip
\begin{user_prompt}
You are evaluating whether a neuron in a vision model is monosemantic (responds to one visual feature) or polysemantic (responds to multiple unrelated visual features). Below are the top 20 images that most strongly activate this neuron. Each row shows: the original image, then the activation heatmap from a model and its heatmap overlay over the image. These maps show where the neuron fires (bright = strong activation). Focus on the heatmap regions — what visual feature (texture, shape, color, pattern, object part, object category, etc.) is consistently highlighted across images? Even if the objects in the images differ, the neuron may still be monosemantic if it consistently responds to the same low-level or mid-level visual feature (e.g., "striped patterns", "curved edges",  "glossy surfaces"). It could be an abstract, high level concept, or it could be a common visual feature such as texture, which may not be easily nameable but is still a single unifying pattern.
\\
\\
Is there a single visual feature or concept that explains why this neuron fires in all of these images?
Respond in JSON:
\\
\{ \\
  "is\_monosemantic": true/false, \\
  "description": "the shared visual feature if monosemantic, or why not if polysemantic"\\
\}
 
\end{user_prompt}

\newpage
\subsection{Sensitivity to the Number of Top-$k$ Activating Images}
\label{sec:topk_sensitivity}

\begin{wrapfigure}{r}{0.45\textwidth}
    \vspace{-12pt}
    \centering
    \includegraphics[width=0.45\textwidth]
    {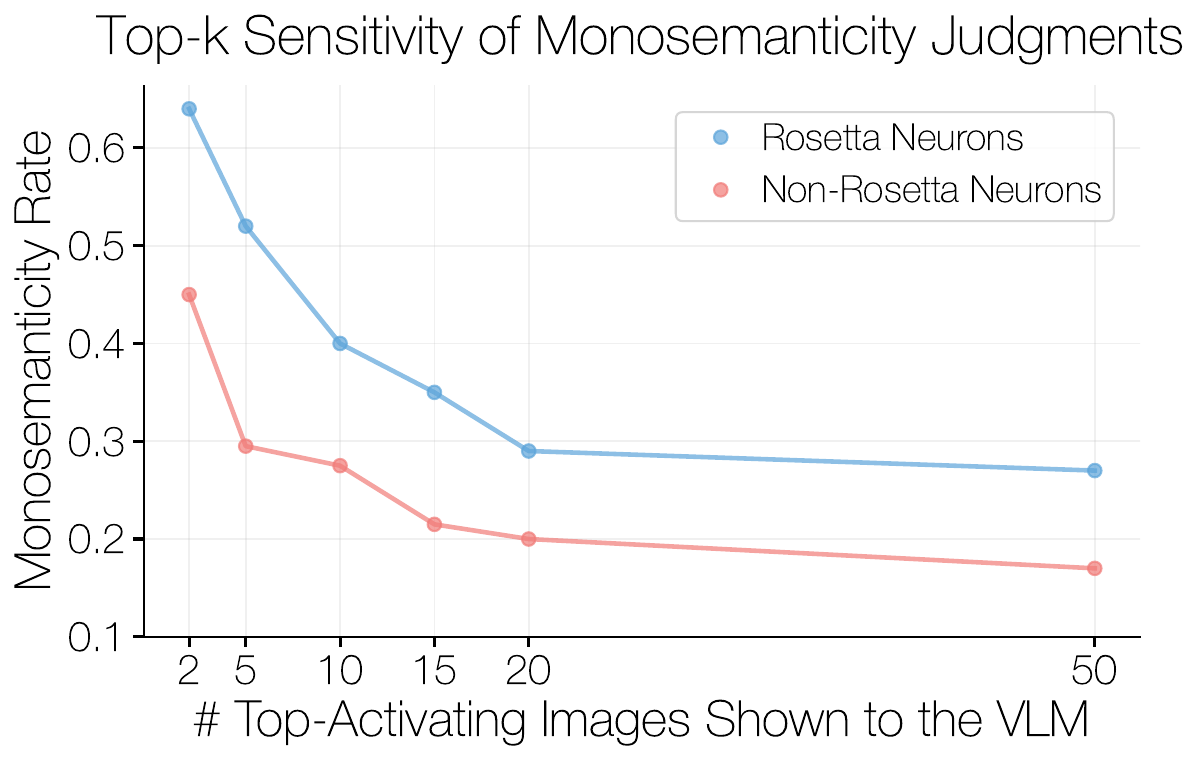}
    \vspace{-15pt}
    \caption{
\textbf{Effect of the number of top-activating images shown to the VLM judge.}
}
    \label{fig:monosemanticity_sensitivity}

\end{wrapfigure}

We ablate the number of top-activating images shown to the VLM judge. Specifically, we repeat the monosemanticity evaluation with $k \in \{2,5,10,15,20,50\}$, constructing each composite from the top-$k$ images and their corresponding activation maps and overlays. As shown in Figure~\ref{fig:monosemanticity_sensitivity}, the monosemanticity rates decrease as more examples are shown, but the gap between Rosetta and non-Rosetta neurons remains stable across all values of $k$. The estimates begin to stabilize around $k=20$, while using substantially fewer examples provides less evidence for judging whether a single coherent feature explains the neuron. We therefore use $k=20$ in the main experiments.

\subsection{Validation of VLM-as-a-Judge as a Predictive Metric}
\label{sec:vlm_validation}

\textbf{Experimental setup.} To validate our VLM-as-a-judge setup as a reliable metric, we design a prediction task that tests whether the VLM can recover and generalize the visual features that neurons fire for, rather than relying on spurious cues. We sample 100 random neurons from OpenCLIP ViT-B/16. For each neuron, the VLM is shown the top 20 most-activating images together with their activation maps in a composite grid. We first provide the system prompt described in~\Cref{sec:detailed_vlm_setup} to the model. We then construct a held-out test set consisting of the next 5 highest-activating images and 5 random images drawn from the bottom 50\% of activations. These images are shuffled and presented as a labeled grid containing only the raw images. The VLM must predict which images would activate the neuron based only on the training composite. To do this, we provide the prior conversation based on the 20-image training composite in context, and then issue a test prompt asking the model to predict on the 10 test images. We provide this prompt below.
\smallskip
\smallskip
\begin{user_prompt}
IMAGE 2 (TEST): The second composite image is a grid of 10 test images labeled A through J. Given the previous composite image you saw of top activating images and the corresponding activation maps, predict which of these ten test images will activate this neuron. 
\\
Respond in JSON:
\\
\{\\
  "neuron\_description": "brief description of what visual feature(s) the neuron responds to",\\
  "predicted\_activating": ["A", "C", ...],\\
  "reasoning": "brief explanation of why you chose these images and rejected the others"\\
\}
 
\end{user_prompt}
\smallskip
\smallskip
\textbf{VLM-as-a-Judge is predictive of neuron selectivity.} Performance is evaluated over 5 independent trials, each using a disjoint random subset of 100 neurons. We report 95\% confidence intervals for accuracy, precision, recall, and F1 in Table~\ref{tab:vlm_test}. VLM-as-a-judge performs meaningfully above a random baseline. The baseline chance is a random predictor that independently marks each test image as activating with probability $0.5$. In this 5-positive/5-negative setup, its expected accuracy and precision are $0.5$, while its expected Recall and F1 are $0.4995$ and $0.4865$, respectively.

\begin{table}[htbp]
\centering
\label{tab:appendix_metrics}
\begin{tabular}{lcccc}
\hline
Model & Accuracy & Precision & Recall & F1 \\
\hline
VLM-as-a-Judge & $0.792\pm{0.032}$ & $0.865\pm{0.042}$ & $0.698\pm{0.038}$ & $0.764\pm{0.037}$ \\
Chance (expected) & 0.500 & 0.500 & 0.500 & 0.487 \\
\hline
\end{tabular}
\vspace{0.5em}
\caption{\textbf{Validation of GPT-5.4 as a neuron selectivity judge.} For each sampled OpenCLIP ViT-B/16 neuron, GPT-5.4 is shown the top 20 activating images with activation maps. It is then asked to identify the activating images in a shuffled set of 10 unseen raw images containing 5 positives and 5 negatives. We report $95\%$ confidence intervals across 5 independent evaluation runs, with each run sampling 100 distinct neurons. We also report the expected metrics for a random predictor.}
\label{tab:vlm_test}
\end{table}

\newpage
\begin{figure}
    \centering
    \includegraphics[width=1.0\linewidth]{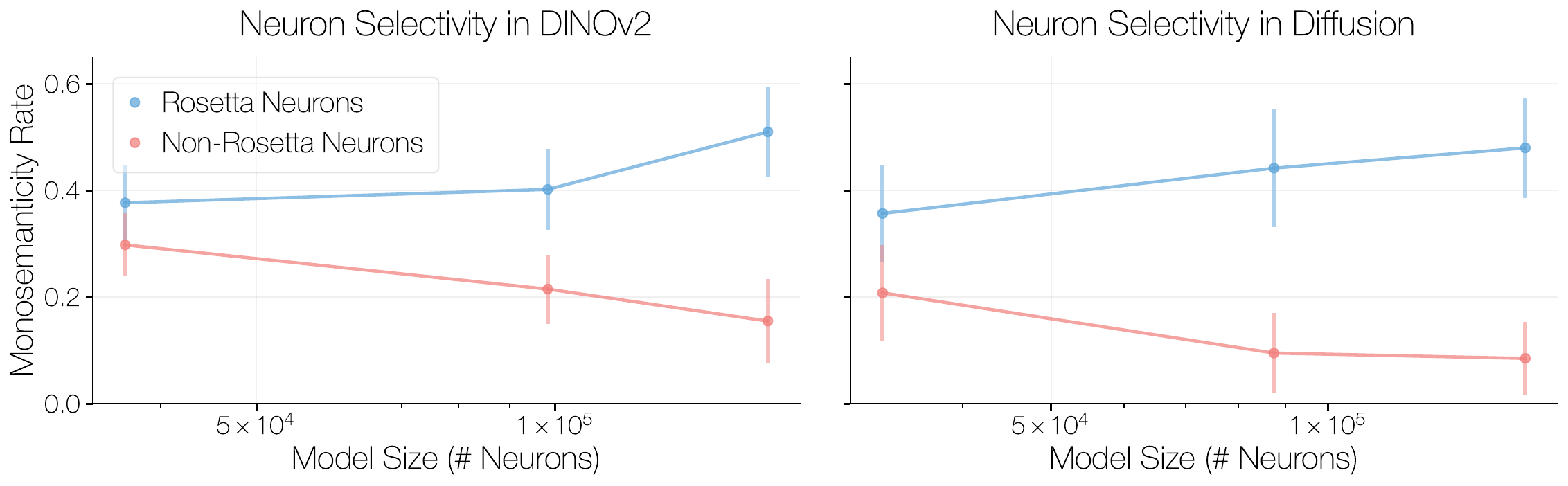}
    \caption{\textbf{VLM-judged monosemanticity rate in DINOv2 and diffusion models.} Rosetta Neurons discovered from DINOv2--Diffusion matching runs exhibit increasing monosemanticity with scale, while non-Rosetta neurons become polysemantic according to the metric. This provides additional evidence for the Neuron Polarization Effect across vision model families.}
    \label{fig:vlm_more_models}
\end{figure}
\subsection{Results for Neuron Selectivity in Other Vision Models}
\label{sec:more_vision_selectivity}
In~\Cref{polsemanticity_section}, we showed evidence for the Neuron Polarization Effect in OpenCLIP: Rosetta Neurons become more selective with scale, while the non-Rosetta population remains comparatively less selective. In this section, we apply the same VLM-as-a-judge setup to measure neuron monosemanticity in DINOv2 and diffusion models as they scale. We use the Rosetta Neurons discovered from the DINOv2--Diffusion matching runs. In~\Cref{fig:vlm_more_models}, we plot the fraction of neurons judged monosemantic as a function of model size. The monosemanticity fraction increases with model scale for Rosetta Neurons and decreases for non-Rosetta neurons in both families, consistent with the trend observed in OpenCLIP and language models. The monosemanticity rates for Rosetta Neurons in DINOv2 and diffusion are similar, as these neurons are identified from the same cross-family matching runs. In contrast, the non-Rosetta populations exhibit model-specific levels of polysemanticity.

\section{Further Discussion and Limitations}
\label{sec:further_limitations}
In this section, we further contextualize the Rosetta Neuron scaling picture. We begin with a DINOv3 failure case, which helps clarify the conditions under which Rosetta Neuron scaling arises. We then discuss the challenge of operationalizing monosemanticity and polysemanticity, and relate our selectivity measures to these concepts.

\begin{wrapfigure}{r}{0.45\textwidth}
    \vspace{-15pt}
    \centering
    \includegraphics[width=0.45\textwidth]
    {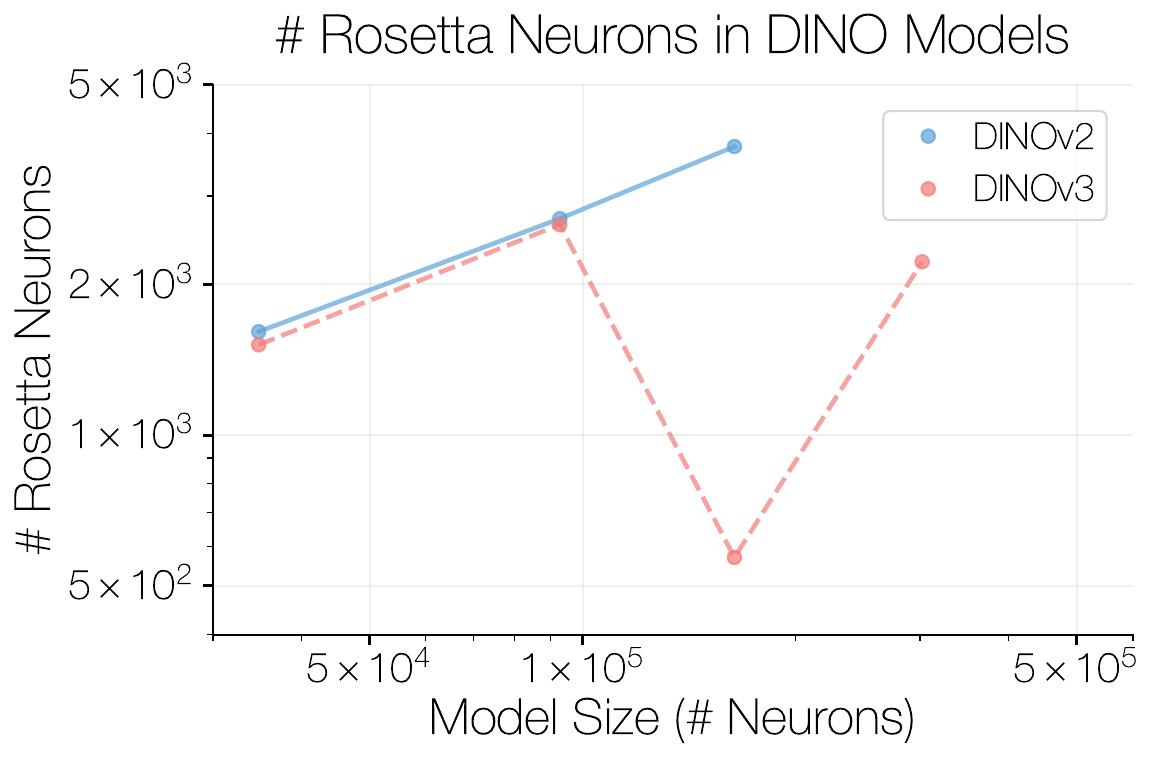}
    \vspace{-15pt}
    \caption{\textbf{DINOv3 does not exhibit a clear Rosetta Neuron scaling law.}}
    \label{fig:rosetta-dinov3}
\end{wrapfigure}
\subsection{DINOv3 as an Informative Failure Case}
\label{sec:dinov3_failure}
A notable exception to the scaling behavior observed in our main experiments from~\Cref{scaling_law_section} is DINOv3~\citep{simeoni2025dinov3}, which does not exhibit a clear Rosetta Neuron scaling law. We repeat the discriminative-to-diffusion matching procedure from~\Cref{scaling_law_section} with DINOv3, and report the resulting Rosetta Neuron counts alongside DINOv2 results in~\Cref{fig:rosetta-dinov3}. In contrast to DINOv2, which follows the trend observed in other vision model families, DINOv3 Rosetta Neuron counts do not follow a monotonic scaling trend. This deviation is consistent with the fact that DINOv3 modifies the DINOv2 training setup with additional constraints on intermediate representations, encouraging them to match statistics from earlier in training. We hypothesize that these constraints alter how neuron-level features are organized. More broadly, the absence of a clean trend in DINOv3 suggests that Rosetta Neuron scaling depends on the interaction between data, architecture, optimization, and training objective. 

\subsection{Operationalizing Monosemanticity and Polysemanticity}
\label{sec:mono_vs_poly}
To interpret neuron selectivity, we need a working definition of monosemanticity and polysemanticity~\citep{bricken2023monosemanticity}. These are general semantic descriptions of representations: a unit is considered monosemantic if it responds to one coherent feature, and polysemantic if it responds to multiple unrelated features. These terms are related to, but distinct from the more formal notion of superposition~\citep{elhage2022toy}, which describes feature interference in activation space. Since the underlying feature directions are not directly available in real models, prior work has relied on proxies such as top-activating examples, output-space projections~\citep{geva2021transformer}, activation maps~\citep{bau2017network}, and VLM-based evaluations~\citep{shaham2024multimodal}. These judgments therefore depend on the input distribution, the evaluation procedure, and the level of semantic granularity. For example, a unit that responds to both cats and dogs may be viewed as monosemantic at the level of mammals, but polysemantic at the level of animal species.

Given this ambiguity, we do not treat monosemanticity and polysemanticity as fixed formal definitions. Our goal is instead to understand population-level trends for selectivity as models scale. To do this, we use modality-specific proxies such as vocabulary-space excess kurtosis for language models and VLM judgments for vision models. Additionally, qualitative visualizations of top-activating examples and activation maps show patterns consistent with our measurements. This evaluation does not settle the general problem of defining monosemanticity in real models, but it is well suited to the relative prediction studied here: Rosetta Neurons should become increasingly selective with scale compared with the non-Rosetta background. Developing more principled operational definitions remains an important direction for future work.

\section{Compute Resources}
\label{sec:compute}

All experiments were run on NVIDIA A100 80GB GPUs. Neuron matching experiments and related ablations used 8 GPUs per model pair. Matching between two models in the ${\sim}{100}\text{M}$ parameter family takes around 30 minutes, while matching models in the ${\sim}{30}\text{B}$ parameter family takes about 24 hours. This corresponds to roughly 4--192 A100 GPU-hours per model-pair matching run depending on the model scale. The continued pretraining of GPT2-1.5B takes around 30 minutes on 8 GPUs, corresponding to roughly 4 A100 GPU-hours per run. The rest of the experiments were run using a single GPU. We did not train any of the pretrained language or vision models used in the scaling-law experiments. These experiments use existing checkpoints.


\end{document}